\documentclass[11pt]{article}

\usepackage[T1]{fontenc}
\usepackage[utf8]{inputenc}
\usepackage[margin=1in]{geometry}
\usepackage{microtype}
\usepackage{graphicx}
\usepackage{amsmath,amssymb}
\usepackage{multirow}
\usepackage{booktabs}
\usepackage{makecell}
\usepackage{relsize}
\usepackage{caption}
\usepackage{tabularx}
\usepackage{colortbl,xcolor}
\usepackage{longtable}
\usepackage{adjustbox}
\usepackage{pdflscape}
\usepackage{subcaption}
\usepackage{ragged2e}
\usepackage[numbers,sort&compress]{natbib}
\usepackage{xurl}
\usepackage[hidelinks]{hyperref}

\hypersetup{
  pdftitle={Learning Interpretable Differentiable Logic Networks for Time-Series Classification},
  pdfauthor={Chang Yue and Niraj K. Jha},
  pdfsubject={Time-series classification and differentiable logic networks},
  pdfkeywords={classification, edge computing, interpretability, logic rules, machine learning, neural networks, time series}
}

\newcommand{\Description}[1]{}

\newcolumntype{Y}{>{\centering\arraybackslash}X}
\newcolumntype{C}[1]{>{\centering\arraybackslash}p{#1}}
\newcolumntype{L}{>{\raggedright\arraybackslash}X}
\definecolor{lightgray}{gray}{0.75}
\newcommand{\meanstd}[2]{%
  \makecell[c]{#1 \\[-3pt] {\scriptsize$\pm$ #2}}%
}

\newcolumntype{Z}[1]{>{\RaggedRight\arraybackslash}p{#1}}

\title{Learning Interpretable Differentiable Logic Networks for Time-Series Classification}
\author{Chang Yue \quad Niraj K. Jha\\
\small Department of Electrical and Computer Engineering, Princeton University\\
\small Princeton, New Jersey, USA\\
\small \texttt{\{cyue,jha\}@princeton.edu}}
\date{}

\begin{document}
\maketitle

\begin{abstract}
Differentiable logic networks (DLNs) have shown promising results in tabular domains by 
combining accuracy, interpretability, and computational efficiency. In this work, we 
apply DLNs to the domain of TSC for the first time, focusing on univariate 
datasets. To enable DLN application in this context, we adopt feature-based 
representations relying on Catch22 and TSFresh, converting sequential time series into 
vectorized forms suitable for DLN classification. Unlike prior DLN studies that fix
the training configuration and vary various settings in isolation via ablation, we integrate
all such configurations into the hyperparameter search space, enabling the search 
process to select jointly optimal settings. We then analyze the distribution of 
selected configurations to better understand DLN training dynamics. We evaluate our approach 
on 51 publicly available univariate TSC benchmarks. 
The results confirm that classification DLNs maintain their core strengths in this new 
domain: they deliver competitive accuracy, retain low inference cost, and provide 
transparent, interpretable decision logic, thus aligning well with previous DLN findings in 
the realm of tabular classification and regression tasks.
\end{abstract}

\vspace{0.75em}

\section{Introduction}
The synthesis of symbolic logic and deep learning has given rise to a new 
class of interpretable models that stand as a powerful alternative to opaque, black-box 
architectures. The pioneering work in this domain includes deep differentiable logic 
gate networks (LGNs) and their successor, convolutional LGNs, both proposed by 
Petersen \emph{et al.}~\cite{petersen2022deep, petersen2024convolutional}. These models 
establish a method for training networks of Boolean logic gates via gradient descent 
by creating differentiable relaxations of the logic operations. This core strategy 
yields transparent models from which logical rules can be extracted, while extensions 
like logical convolutions and pooling enable these architectures to scale effectively 
to complex classification tasks.

The LGN paradigm was significantly generalized with the development of differentiable logic 
networks (DLNs) \cite{10681646}. By removing prior constraints, such as 
fixed network topologies and binary-only inputs, DLNs broadened the framework's 
applicability to general tabular classification, resulting in highly efficient models 
that match or exceed the performance of traditional neural networks~(NNs) on a wide 
range of benchmarks. The scope of these logic-based systems was expanded once more 
to handle continuous targets through the regression DLN~\cite{yue2025rdln}. This 
extension integrates a weighted sum of the final layer's logical rule outputs to 
produce a numerical prediction, successfully adapting the architecture to regression 
while maintaining its inherent interpretability.

\begin{figure}[!t]
\centering
\includegraphics[width=\linewidth]{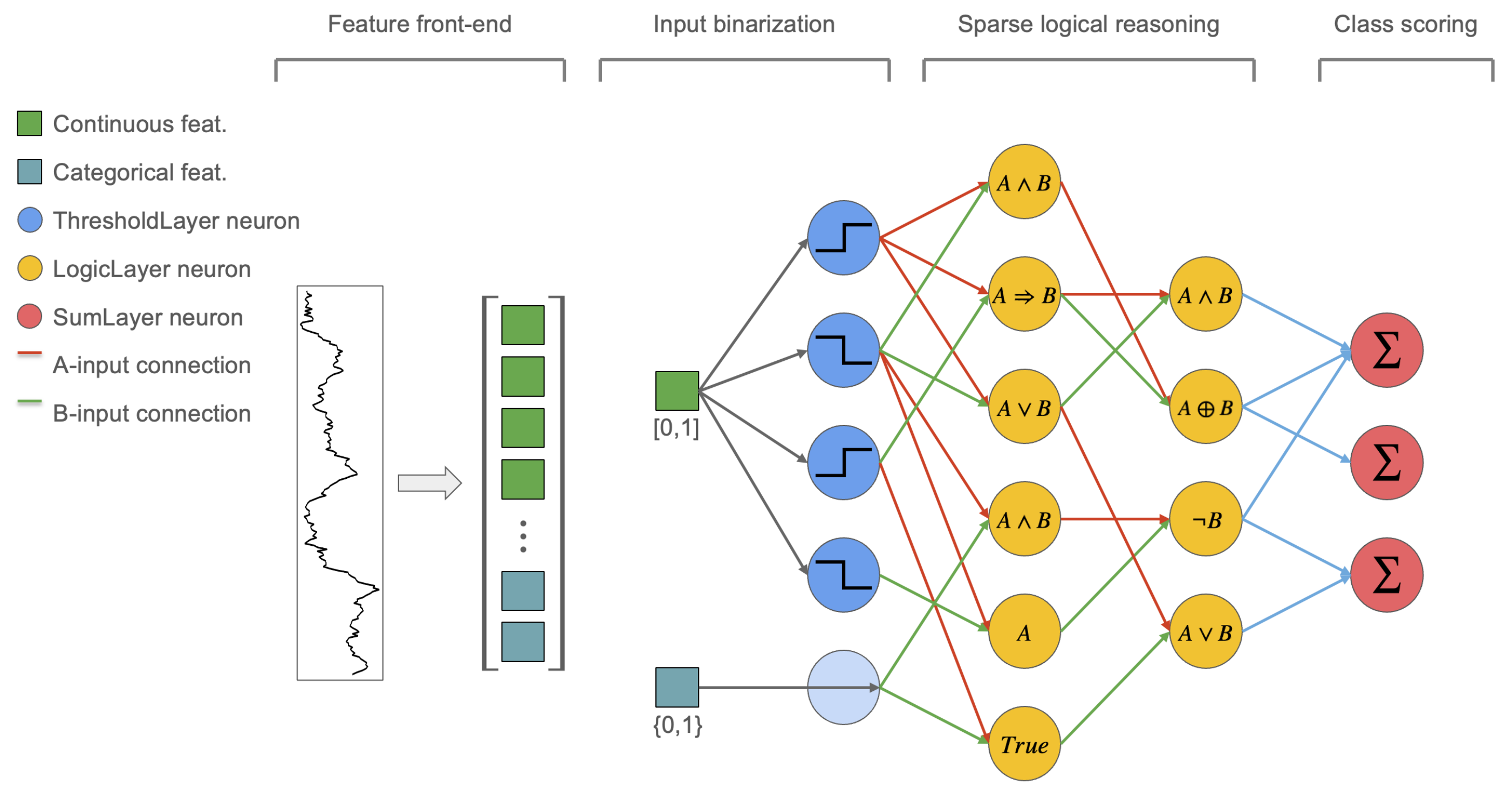}
\caption{Overview of the TSC-DLN architecture. The feature front-end transforms each raw time series into a fixed-dimensional vector, with continuous features scaled to $[0,1]$ and categorical features represented as binary values. The ThresholdLayer converts continuous features into learned binary predicates, while categorical features enter the logical network directly. Each LogicLayer neuron selects two incoming signals, denoted by $A$ and $B$ (red and green connections, respectively), and implements
one of the 16 two-input Boolean operations. The SumLayer aggregates selected logic outputs into one score per class, and the class with the highest score is predicted.}
\Description{Diagram of the TSC-DLN pipeline. On the left, time series inputs are 
converted by candidate transformations into fixed-length real-valued feature vectors. 
On the right, these features pass through a ThresholdLayer that binarizes them, then 
through stacked LogicLayers composed of binary logic operations, and finally into a 
SumLayer that aggregates class scores to produce the predicted label.}
\label{example}
\end{figure}

Despite the success of DLNs in tabular classification and their extension to regression, 
their application to time series data, particularly for classification tasks, remains 
largely unexplored. This article addresses this gap by applying the classification DLN 
paradigm to univariate time-series classification (TSC). 
DLNs bypass traditional multiply-accumulate (MAC) operations, 
outputting pure Boolean logic networks that can be mapped directly to look-up tables 
in field-programmable gate arrays (FPGAs) or standard cells in application-specific integrated circuits (ASICs). This 
provides a pathway to ultra-low-power, interpretable edge-device deployment.
A simplified time-series 
classification DLN (TSC-DLN) is shown in Fig.~\ref{example}. Time-series 
data present unique challenges due to their sequential nature and varying lengths. This is
not directly compatible with DLNs that typically require fixed-size input vectors. 
To overcome this obstacle, we leverage feature-based transformations that convert raw 
univariate time-series sequences into fixed-length feature vectors, making them 
suitable for DLN processing. 
This transformation-based paradigm is highly effective and widely adopted; 
in fact, many state-of-the-art (SOTA) time-series models across different algorithmic families 
are fundamentally feature-based at their core. For instance, convolution-based methods 
like MiniRocket~\cite{dempster2021minirocket} and MultiRocket~\cite{tan2022multirocket} operate by extracting roughly 10,000 to 50,000 features 
via convolutional kernels, while modern shapelet-based~\cite{hills2014classification} and dictionary-based~\cite{schafer2015boss} methods 
similarly convert sequence shapes and repeating temporal patterns into fixed-length 
tabular vectors before applying a classifier. By showing that DLNs can optimize standard 
feature sets like TSFresh, we demonstrate that DLNs can seamlessly adapt to this broader 
paradigm of converting time-series to tabular formats.

Specifically, we employ two distinct feature extraction methodologies: Catch22 
\cite{lubba2019catch22} and TSFresh~\cite{christ2018time}. \emph{Catch22} (CAnonical 
Time-series CHaracteristics) includes distillation of approximately 
7700 candidate statistics extracted from the extensive \emph{hctsa}~\cite{fulcher2017hctsa} 
repository into a fixed compact set of 22 carefully chosen, non-redundant features. These 
features are not only interpretable and fast to compute but also collectively cover 
a wide range of time-series concepts, offering strong classification accuracy across 
diverse benchmarks. The second method, \emph{TSFresh} (Time Series FeatuRe Extraction 
on basis of Scalable Hypothesis tests), is a comprehensive library capable of 
automatically extracting over 750 time-series features using 63 different 
characterization methods. In our experiments, to manage the high dimensionality of 
TSFresh, we employ cross-validated random forests to select smaller feature subsets 
of sizes 10, 20, or 40 for each dataset. This strategy serves two primary purposes.
First, it significantly reduces the input dimension and lowers computational costs
and, second, provides a more direct comparison with the 22 features extracted using Catch22. 
Crucially, both Catch22 and TSFresh extract a fixed number of 
features, independent of the original time-series sequence length. This is essential 
for the application of classification DLNs and other methods that require a predefined 
input size.

Another significant contribution of this work is an integrated study of DLN training 
configurations. Prior DLN studies typically adopted a fixed training setting and 
conducted ablation studies by altering one flag at a time. For instance, the original 
DLN work found that learning neuron operations and connections separately yielded 
better average balanced accuracy across 20 classification datasets, whereas the study 
on the regression DLN found that learning them simultaneously resulted in a better 
average R$^2$ score. In contrast, our approach integrates various DLN training settings, such 
as the use of straight-through estimators (STEs), the number of candidate logic gate 
and input link options for each logic operator, and whether to train neuron 
functionalities and connections simultaneously, into a comprehensive hyperparameter 
search space. We then allow a search algorithm to determine the optimal configuration, 
followed by a detailed statistical analysis of the settings chosen by the 
hyperparameter optimization (HPO) process. 

To rigorously evaluate our proposed methodology and the impact of training configurations, we 
conduct extensive DLN experiments on 51 univariate TSC datasets from the UCR 
benchmark suite~\cite{dau2019ucr}, testing a total of four transformation methods (Catch22; 
TSFresh with its three feature subsets). We further compare the performance of 
our DLN approach against both traditional machine learning methods and SOTA 
neural networks, demonstrating its efficacy and interpretability in the time-series domain.

The remainder of this article is organized as follows. Section~\ref{sec-related} provides 
an overview of related work on interpretable logic networks and time-series tasks. 
Section~\ref{sec-method} details the classification DLN architecture and the feature 
transformation pipelines. Section~\ref{sec-experiments} presents the experimental setup, 
results, and a comprehensive analysis of the DLN training configurations. Finally, 
Section~\ref{sec-conclusion} concludes the article with a summary of findings and 
outlines directions for future research.

\section{Related Work}\label{sec-related}
This research builds upon the previously developed DLN 
framework and enhances it to the domain of TSC. We first summarize 
the core principles of DLNs and then review related work on time-series analysis.

\subsection{The Differentiable Logic Network Framework}
DLN is a neuro-symbolic architecture introduced for tabular 
classification~\cite{10681646} and later extended to regression~\cite{yue2025rdln}. 
DLN design is guided by three fundamental principles:

\begin{itemize}
    \item \textbf{Differentiability:} The differentiability of the DLN framework is 
    founded on the combination of real-valued logic and softmax function. Real-valued 
    (or fuzzy) logic~\cite{zadeh1978fuzzy, menger2003statistical} enables use of the continuous 
    domain for the operations themselves. The softmax function then serves as a 
    differentiable proxy for `choose-one-from-$N$' structural decisions, such as 
    selecting a specific logic gate or input connection.

    \item \textbf{Interpretability:} DLNs are inherently interpretable, producing a final 
    model that is a clear, human-readable logical formula. This intrinsic transparency 
    removes the need for potentially unreliable post-hoc explainers, such as LIME (Local 
    Interpretable Model-Agnostic Explanations) or SHAP (Shapley Additive Explanations) 
    that are applied to black-box models~\cite{rudin2019stop, ribeiro2016should, lundberg2017unified}. 
    The goal is to create a model that is transparent by design, aligning with other 
    interpretable systems like decision trees and rule-based 
    models~\cite{xin2022exploring, wang2015falling, rudin2022interpretable}.

    \item \textbf{Efficiency:} The architecture is inherently efficient. The learning 
    process promotes sparsity, resulting in a lightweight final model. Because the model 
    is composed of simple operations, it is computationally fast at inference 
    time and well-suited for deployment on resource-constrained hardware or accelerators like 
    FPGAs, which excel at running binarized and logic-based 
    networks~\cite{umuroglu2020logicnets, petersen2024convolutional}.
\end{itemize}

In this article, we demonstrate the versatility of an enhanced DLN framework by applying it to 
TSC, achieved by integrating a time-series feature transformation 
front-end.

\subsection{Time-series Classification}
Time-series data, i.e., sequences of observations ordered chronologically, are ubiquitous 
across many domains, including finance, healthcare, and industrial monitoring. Analysis of
such data encompasses a variety of tasks, such as forecasting future values, clustering 
similar series, detecting anomalies, and classification. These tasks may involve either 
univariate time series, which consist of a single sequence of observations, or 
multivariate time series, which involve multiple time-dependent variables. To 
standardize the evaluation of algorithms for these tasks, several benchmark archives have 
been developed, with the UCR/UEA Time-Series Classification Archive being one of the most 
widely used and recognized~\cite{dau2019ucr, bagnall2018uea}.

The field of TSC has seen a significant expansion with diverse 
methodologies. A comprehensive review and experimental evaluation by Bagnall 
\emph{et al.}~\cite{middlehurst2024bake} organizes the landscape of TSC algorithms into 
eight distinct families. These families include: 1) Distance-based methods that
classify series based on their proximity under a specific distance measure, with dynamic 
time warping being a classic example~\cite{rakthanmanon2013addressing}; 
2) Feature-based methods that transform the series into a vector of summary 
statistics~\cite{lubba2019catch22, christ2018time, middlehurst2022freshprince}; 
3) Interval-based approaches that extract ensembles of features from various 
intervals, such as QUANT~\cite{dempster2024quant}; 4) Shapelet-based methods that 
identify discriminative sub-sequences (shapelets) to distinguish between 
classes~\cite{hills2014classification}; 5) Dictionary-based methods, where histograms of 
counts of repeating patterns serve as features for a classifier, exemplified by BOSS 
\cite{schafer2015boss} and symbolic Fourier approximation~\cite{schafer2012sfa}; 
6) Convolution-based approaches that use convolutional kernels to detect 
patterns~\cite{dempster2021minirocket}; 7) Deep learning models that leverage neural 
networks~\cite{wang2017time, ismail2023lite}; and 8) Hybrid approaches that combine two 
or more of these families~\cite{middlehurst2021hive}.

Our work focuses on feature-based methods that transform a time series of arbitrary 
length into a fixed-length feature vector, enabling the use of standard machine learning 
classifiers. This approach offers an interpretable way to handle time-series data. 
Prominent examples of feature-based toolkits include \textit{Catch22}~\cite{lubba2019catch22}
that provides a curated set of 22 highly informative and minimally redundant features, 
and \textit{TSFresh}~\cite{christ2018time} that extracts a comprehensive set of over 
750 features. While our primary focus is on these feature extraction libraries, it is 
worth noting that other transformation techniques also convert sequences into a vector 
format. For instance, the random dilated shapelet transform~\cite{guillaume2022random} 
uses an efficient transformation based on randomly generated, dilated subsequences 
(shapelets). Similarly, the dictionary-based WEASEL 2.0~\cite{schafer2023weasel} and the 
convolution-based ROCKET~\cite{dempster2020rocket} frameworks also produce feature vectors from the 
original time series, demonstrating the convergence of different algorithmic families 
towards a transformation-based paradigm.

Recent deep sequence models have further advanced TSC through learned temporal 
representations. Zerveas \emph{et al.}~\cite{zerveas2021transformer} introduced a 
Transformer encoder with supervised and self-supervised objectives for multivariate 
time-series classification and regression. PatchTST~\cite{nie2023patchtst} divides each 
channel into temporal patches and employs a channel-independent Transformer, reducing 
attention cost and enabling longer input contexts; although developed primarily for 
forecasting, it is representative of recent patch-based time-series representation 
learning. TimesNet~\cite{wu2023timesnet} instead identifies dominant periods, reshapes 
one-dimensional sequences into multiple two-dimensional tensors, and models intra- and 
inter-period variations, supporting classification in addition to forecasting, imputation, 
and anomaly detection. In contrast to these end-to-end representation-learning approaches, 
our framework uses explicit statistical descriptors and focuses on an interpretable and 
efficient downstream logical classifier.

Many of the aforementioned techniques were initially developed for univariate time series. 
Extending them to the multivariate domain is an active area of research. Broadly, these 
extensions can be categorized into channel-independent and cross-channel approaches. 
Channel-independent methods, such as the collective of transformation ensembles 
\cite{bagnall2015time}, apply univariate classifiers to each channel separately and 
combine their predictions. In contrast, cross-channel methods aim to capture the 
dependencies and interactions between different channels. For example, the multivariate 
extensions of ROCKET~\cite{tan2022multirocket} have demonstrated strong performance by 
explicitly considering cross-channel information during the transformation process.

\section{Methodology}\label{sec-method}
\begin{figure}[!htbp]
\centering
\includegraphics[width=0.33\columnwidth]{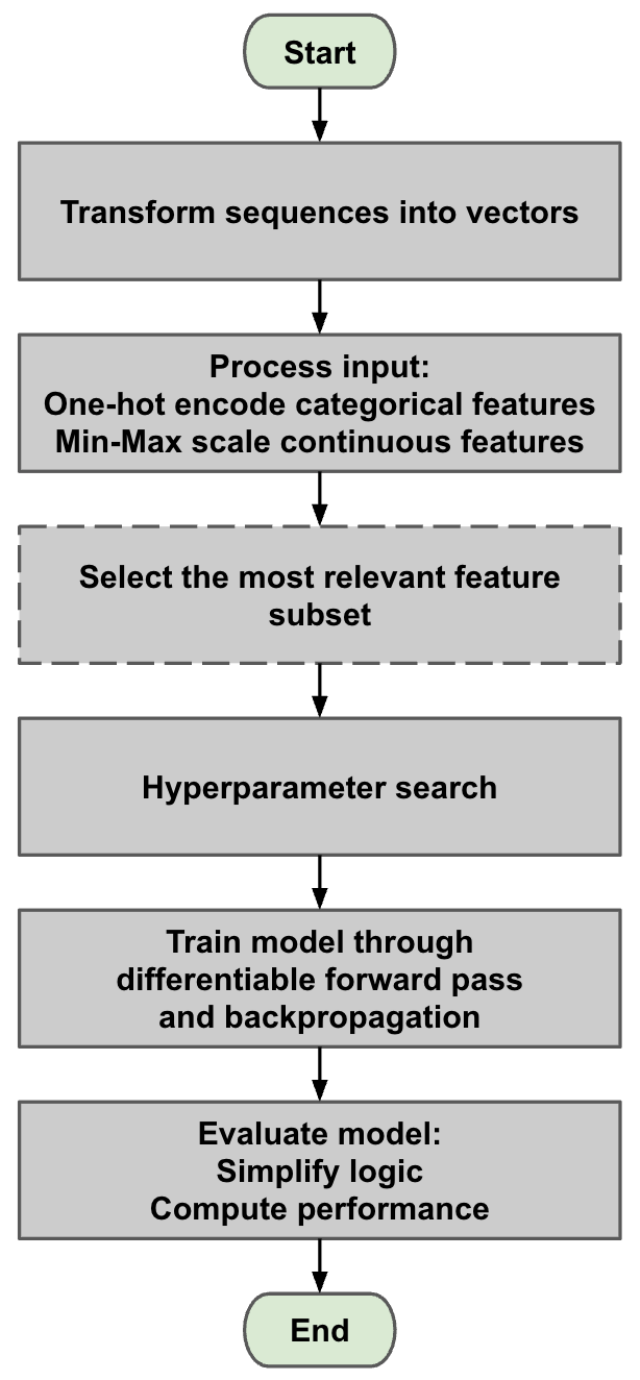}
\caption{Experimental workflow for TSC-DLN.}
\label{flowchart}
\Description{Flowchart showing the experimental workflow for TSC-DLN.}
\end{figure}

This work enhances the DLN framework, originally proposed for tabular 
classification~\cite{10681646} and later adapted to regression~\cite{yue2025rdln}, and applies it to the 
TSC domain. The complete workflow, called TSC-DLN and shown in 
Fig.~\ref{flowchart}, begins with data transformation, followed by a hyperparameter search, 
and concludes with model training and evaluation. Compared to our earlier DLN variants, 
two modifications are required:
\begin{enumerate}
    \item Feature extraction for time series: Raw, variable-length sequences are converted 
    to fixed-length tabular vectors.
    \item Automated architectural search: Several design choices (two-phase vs. unified 
    training, search-space pruning, use of straight-through estimators, and input 
    concatenation) are no longer hard-coded but are selected automatically through
    hyperparameter optimization.
\end{enumerate}

\subsection{Time-series Feature-based Transformation}
To apply our tabular DLN model, we first transform each variable-length time series into a 
fixed-dimension feature vector. We explore the following two feature extraction methods:
\begin{itemize}
    \item \textbf{Catch22}~\cite{lubba2019catch22}: This method transforms any time series 
    into a canonical and efficient set of 22 features, capturing a wide range of sequence 
    properties.
    \item \textbf{TSFresh}~\cite{christ2018time}: This method generates a comprehensive 
    set of features. To ensure relevance and reduce dimensionality, we then employ cross-validated 
    random forests to select the most salient features, creating feature vectors of size 
    10, 20, or 40.
\end{itemize}
The resulting feature vector is then scaled to have values within $[0,1]$ and serves as the 
DLN input.

\definecolor{func}{RGB}{31,119,180}
\definecolor{conn}{RGB}{214,39,40}
\newcommand{\funcparam}[1]{\textcolor{func}{#1}}
\newcommand{\connparam}[1]{\textcolor{conn}{#1}}
\newcommand{\cellcontent}[1]{\begin{minipage}{\linewidth}\raggedright #1\end{minipage}}
\newcommand{\phspace}{\hspace{0.8em}}

\subsection{Differentiable Logic Network}
The DLN follows the three-layer template, as illustrated in Fig.~\ref{example}:
\begin{enumerate}
    \item \textbf{ThresholdLayer}: Differentiable binarization of continuous inputs. Each 
    input is linked to a predefined number of threshold neurons (10 or 14 in our 
    experiments), with the layer learning the optimal cutoff value for each one.
    \item \textbf{LogicLayer(s)}: Sparse, two-input Boolean neurons that enable nonlinearity 
    and interpretability. For each neuron, the layer learns its logic operation and which 
    two outputs from the previous layer serve as its inputs.
    \item \textbf{SumLayer}: Class-wise aggregation of the final LogicLayer's activations. 
    This layer learns which of these activations (i.e., logic rules) to sum up for each class to 
    produce the final scores.
\end{enumerate}

To enable gradient-based training, we employ the same differentiable relaxation techniques 
as in our prior work. Discrete choices, such as selecting a logic function for a neuron or 
its input connections, are relaxed into continuous, probabilistic mixtures using softmax 
or sigmoid functions. Table~\ref{layer-params} summarizes the layer-wise parameters and 
their forward-pass computations. Trainable parameters are color-coded: \funcparam{blue} 
for neuron function weights and \connparam{red} for connection weights. We now detail the 
forward equations of each DLN layer, both during training (differentiable relaxations) and 
during inference (discrete execution).

\begin{table}[t]
\centering
\caption{Trainable parameters and forward equations of the proposed TSC-DLN.
Blue entries (\funcparam{$\bullet$}) are \emph{neuron‑function} weights; 
red entries (\connparam{$\bullet$}) are \emph{connection} weights. 
$\mathbf{x}$ / $\mathbf{y}$ denote layer input / output, 
$\tau$ is the shared temperature during training, and $\theta_{\text{sum}}\!=\!0.8$.}
\label{layer-params}
\small{
\setlength{\tabcolsep}{4pt}
\begin{adjustbox}{max width=\linewidth}
\begin{tabular}{>{\centering\arraybackslash}p{1.3cm}
                >{\raggedright\arraybackslash}p{3.9cm}
                >{\raggedright\arraybackslash}p{3.8cm}
                >{\raggedright\arraybackslash}p{3.7cm}}
\toprule
 & Trainable parameters & Training forward (differentiable) & Inference forward (discrete) \\ \midrule
\cellcontent{Threshold\\Layer} &
\cellcontent{\funcparam{$\bullet$} bias $\funcparam{\mathbf{b}}\!\in\!\mathbb R^{\text{in}}$\\
             \funcparam{$\bullet$} slope $\funcparam{\mathbf{s}}\!\in\!\mathbb R^{\text{in}}$} &
\cellcontent{$\displaystyle
 \mathbf{y}_i=\mathrm{Sigmoid} \Bigl(\funcparam{\mathbf{s}_i}\,(\mathbf{x}_i-\funcparam{\mathbf{b}_i}) /\tau \Bigr)$} &
\cellcontent{$\displaystyle
 \mathbf{y}_i=\mathrm{Heaviside} \bigl(\funcparam{\mathbf{s}_i}\,(\mathbf{x}_i-\funcparam{\mathbf{b}_i})\bigr)$}\\
\midrule
\cellcontent{Logic\\Layer} &
\cellcontent{\funcparam{$\bullet$} logic‑fn weights $\funcparam{\mathbf{W}}\!\in\!\mathbb R^{\text{out}\times16}$\\
             \connparam{$\bullet$} link‑$a$ weights $\connparam{\mathbf{U}}\!\in\!\mathbb R^{\text{out}\times\text{in}}$\\
             \connparam{$\bullet$} link‑$b$ weights $\connparam{\mathbf{V}}\!\in\!\mathbb R^{\text{out}\times\text{in}}$} &
\cellcontent{$\displaystyle
 P=\mathrm{Softmax}\bigl(\funcparam{\mathbf{W}_{i,:}}/\tau\bigr)$\\
 $a=\sum\nolimits_{j}\bigl[\mathrm{Softmax}\bigl(\connparam{\mathbf{U}_{i,:}}/\tau \bigr)\bigr]_j\,\mathbf{x}_j$\\
 $b=\sum\nolimits_{j}\bigl[\mathrm{Softmax}\bigl(\connparam{\mathbf{V}_{i,:}}/\tau \bigr)\bigr]_j\,\mathbf{x}_j$\\
 $\mathbf{y}_i=\sum\nolimits_{k=0}^{15}P_k\,\mathrm{SoftLogic}_k(a,b)$} &
\cellcontent{$\displaystyle
\begin{aligned}
k   &= \arg\max\nolimits_{j}\funcparam{\mathbf{W}_{i,j}}\\
a_i &= \arg\max\nolimits_{j}\connparam{\mathbf{U}_{i,j}}\\
b_i &= \arg\max\nolimits_{j}\connparam{\mathbf{V}_{i,j}}\\
\mathbf{y}_i &= \mathrm{HardLogic}_k\bigl(\mathbf{x}_{a_i},\mathbf{x}_{b_i}\bigr)
\end{aligned}$} \\
\midrule
\cellcontent{Sum\\Layer} &
\cellcontent{\connparam{$\bullet$} link weights $\connparam{\mathbf{S}}\!\in\!\mathbb R^{\text{in}\times C}$} &
\cellcontent{$\displaystyle
\mathbf{y}_c=\sum\nolimits_{j}\mathrm{Sigmoid}\bigl(\connparam{\mathbf{S}_{j,c}}/\tau\bigr)\,\mathbf{x}_j$} &
\cellcontent{$\displaystyle
 \mathbf{y}_c=\sum\nolimits_{j}1_{\{\mathrm{Sigmoid}(\connparam{\mathbf{S}_{j,c}}/\tau)\ge\theta_{\text{sum}}\}}\,\mathbf{x}_j$}\\
\bottomrule
\end{tabular}
\end{adjustbox}
}
\end{table}

\textit{Notation.} Let $\mathbf{x}\!\in\![0,1]^{\text{in}}$ be a layer input and $\mathbf{y}$ its output, $\tau$ the 
common decaying temperature during training, and $\theta_{\text{sum}}\!=\!0.8$ the fixed connectivity threshold used 
by the SumLayer at inference.

\textit{ThresholdLayer.}  
During training, each neuron applies a temperature-scaled sigmoid,
\[
\mathbf{y}_i \;=\;
\mathrm{Sigmoid}\bigl(\funcparam{\mathbf{s}_i}\,(\mathbf{x}_i-\funcparam{\mathbf{b}_i})/\tau\bigr),
\]
where \funcparam{$\mathbf{b}_i$} and \funcparam{$\mathbf{s}_i$} are its learnable bias and slope, respectively.
At inference time, the layer is discretised to a hard threshold,
\[
\mathbf{y}_i \;=\;
\mathrm{Heaviside}\bigl(\funcparam{\mathbf{s}_i}\,(\mathbf{x}_i-\funcparam{\mathbf{b}_i})\bigr).
\]
Each continuous input is connected to a group of threshold neurons. The biases for these 
neurons are initialized using the bin edges from a decision tree trained on that input 
feature and their slopes are initialized to a value of $2$. After training, any neurons 
with bias values pushed beyond the input range of $[0, 1]$ are treated as constant 
\texttt{True} or \texttt{False} values. Experimental results show that this design enables 
the DLN to perform efficient feature selection.

\begin{table}[t]
    \centering
    \small
    \caption{List of all real-valued binary logic operations. Adapted from LGN~\cite{petersen2022deep}.}  
    \label{operator}
    \begin{tabular}{rclccccccccc}
        \toprule
        ID    & Operator                    & Real-valued   & 00 & 01 & 10 & 11 \\
        \midrule
           0  & False                       & $0$                       & 0     & 0     & 0     & 0     \\
           1  & $A\land B$                  & $A\cdot B$                & 0     & 0     & 0     & 1     \\
           2  & $\neg(A \Rightarrow B)$     & $A-AB$                    & 0     & 0     & 1     & 0     \\
           3  & $A$                         & $A$                       & 0     & 0     & 1     & 1     \\
           4  & $\neg(A \Leftarrow B)$      & $B-AB$                    & 0     & 1     & 0     & 0     \\
           5  & $B$                         & $B$                       & 0     & 1     & 0     & 1     \\
           6  & $A \oplus B$                & $A + B - 2AB$             & 0     & 1     & 1     & 0     \\
           7  & $A \lor B$                  & $A + B - AB$              & 0     & 1     & 1     & 1     \\
           8  & $\neg(A \lor B)$            & $1 - (A + B - AB)$        & 1     & 0     & 0     & 0     \\
           9  & $\neg(A \oplus B)$          & $1 - (A + B - 2AB)$       & 1     & 0     & 0     & 1     \\
           10 & $\neg B$                    & $1 - B$                   & 1     & 0     & 1     & 0     \\
           11 & $A \Leftarrow B$            & $1-B+AB$                  & 1     & 0     & 1     & 1     \\
           12 & $\neg A$                    & $1-A$                     & 1     & 1     & 0     & 0     \\
           13 & $A \Rightarrow B$           & $1-A+AB$                  & 1     & 1     & 0     & 1     \\
           14 & $\neg(A \land B)$           & $1 - AB$                  & 1     & 1     & 1     & 0     \\
           15 & True                        & $1$                       & 1     & 1     & 1     & 1     \\
        \bottomrule
    \end{tabular}
\end{table}

\textit{LogicLayer.} Each LogicLayer neuron realizes one of the $16$ possible two-input 
Boolean functions. A complete lookup table, adapted from the LGN paper, is provided in 
Table \ref{operator}. Training employs the differentiable real-valued logic 
$\mathrm{SoftLogic}_k$, while inference switches to the binary logic of $\mathrm{HardLogic}_k$. 
For example,
\[
\begin{aligned}
  \mathrm{SoftLogic}_{\textsc{and}}(a,b) &= a\,b,\\
  \mathrm{HardLogic}_{\textsc{and}}(a,b) &= 1_{\{a=1\}}\,1_{\{b=1\}}.
\end{aligned}
\]

Let
\[
\begin{aligned}
P_k      &= \operatorname{Softmax}\bigl(\funcparam{\mathbf{W}_{i,:}}/\tau\bigr)_k,\\
\alpha_j &= \operatorname{Softmax}\bigl(\connparam{\mathbf{U}_{i,:}}/\tau\bigr)_j,\\
\beta_j  &= \operatorname{Softmax}\bigl(\connparam{\mathbf{V}_{i,:}}/\tau\bigr)_j,
\end{aligned}
\]
select, respectively, the Boolean operator and the two incoming signals.
Training employs the differentiable relaxation
\[
\begin{aligned}
a          &= \sum_{j}\alpha_j\,\mathbf{x}_j,\\
b          &= \sum_{j}\beta_j\,\mathbf{x}_j,\\
\mathbf{y}_i &= \sum_{k=0}^{15} P_k\,\operatorname{SoftLogic}_k(a,b).
\end{aligned}
\]
Inference quantizes all three categorical choices,
\[
\begin{aligned}
k^{\star} &= \arg\max_k \funcparam{\mathbf{W}_{i,k}},\\
a^{\star} &= \arg\max_j \connparam{\mathbf{U}_{i,j}},\\
b^{\star} &= \arg\max_j \connparam{\mathbf{V}_{i,j}},
\end{aligned}
\]
and evaluates the corresponding hard gate,
\[
\mathbf{y}_i \;=\;
\operatorname{HardLogic}_{k^{\star}}
  \bigl(\mathbf{x}_{a^{\star}},\mathbf{x}_{b^{\star}}\bigr).
\]

\textit{SumLayer.}  
For each class $c\in\{1,\dots,C\}$, the output logit during training is
\[
\mathbf{y}_c
  \;=\;
  \sum_{j}
  \mathrm{Sigmoid}\bigl(\connparam{\mathbf{S}_{j,c}}/\tau\bigr)\,\mathbf{x}_j,
\]
where \connparam{$\mathbf{S}$} contains the (logit) connection strengths.
At inference time, the connectivity pattern is binarized using
$\theta_{\text{sum}} = 0.8$:
\[
\mathbf{y}_c = \sum_{j}
     \mathbf{1}_{\{\mathrm{Sigmoid}(\connparam{\mathbf{S}_{j,c}}/\tau)\ge\theta_{\text{sum}}\}}
     \,\mathbf{x}_j.
\]
The final class prediction is:
\[
\hat{y} = \arg\max_c \mathbf{y}_c.
\]

After the model is trained, we extract the learned logic rules and simplify 
them using SymPy~\cite{10.7717/peerj-cs.103} to reduce model complexity and enhance 
interpretability.

\subsection{Hyperparameter-driven Methodological Choices}
A primary contribution of this work is to delegate several key methodological decisions 
to the HPO process. While previous work fixes these settings and evaluates them in 
ablation studies, we incorporate them into a unified search space. This empowers the HPO 
algorithm to discover the optimal configuration for each dataset with a 128-trial 
search. The choices automated by HPO include:

\begin{itemize}
    \item \textbf{Training Strategy:} Optimization can proceed either via the two-phase 
    approach from our original work (learning neuron functions and connections alternately) 
    or the unified end-to-end approach from our regression DLN (learning both 
    simultaneously). This is controlled by a Boolean flag.
    
    \item \textbf{Search Subspacing:} To manage complexity and improve gradient flow, we can 
    constrain the search for logic functions and input connections. The HPO process 
    selects the subspace size for logic gate functions from $\{16, 8, 4\}$ candidates and 
    for neuron links from $\{16, 8, 4, 2, 1\}$ candidates.
    
    \item \textbf{Straight-through Estimators (STEs):} To mitigate the impact of vanishing gradients and 
    concentrated activations from cascaded relaxation functions while maintaining a discrete 
    forward pass, we treat the use of STEs~\cite{bengio2013estimating} for each of the three 
    layer types (Threshold, Logic, and Sum) as a Boolean hyperparameter.

    \item \textbf{Input Concatenation:} Inspired by the Wide \& Deep 
    model~\cite{cheng2016wide}, we test the efficacy of creating shortcut connections by 
    concatenating the ThresholdLayer's output to the input of each subsequent LogicLayer. 
    This is also controlled by a Boolean HPO flag.
\end{itemize}

\section{Experiments}\label{sec-experiments}
\begin{figure}[!htbp]
\centering
\includegraphics[width=0.8\columnwidth]{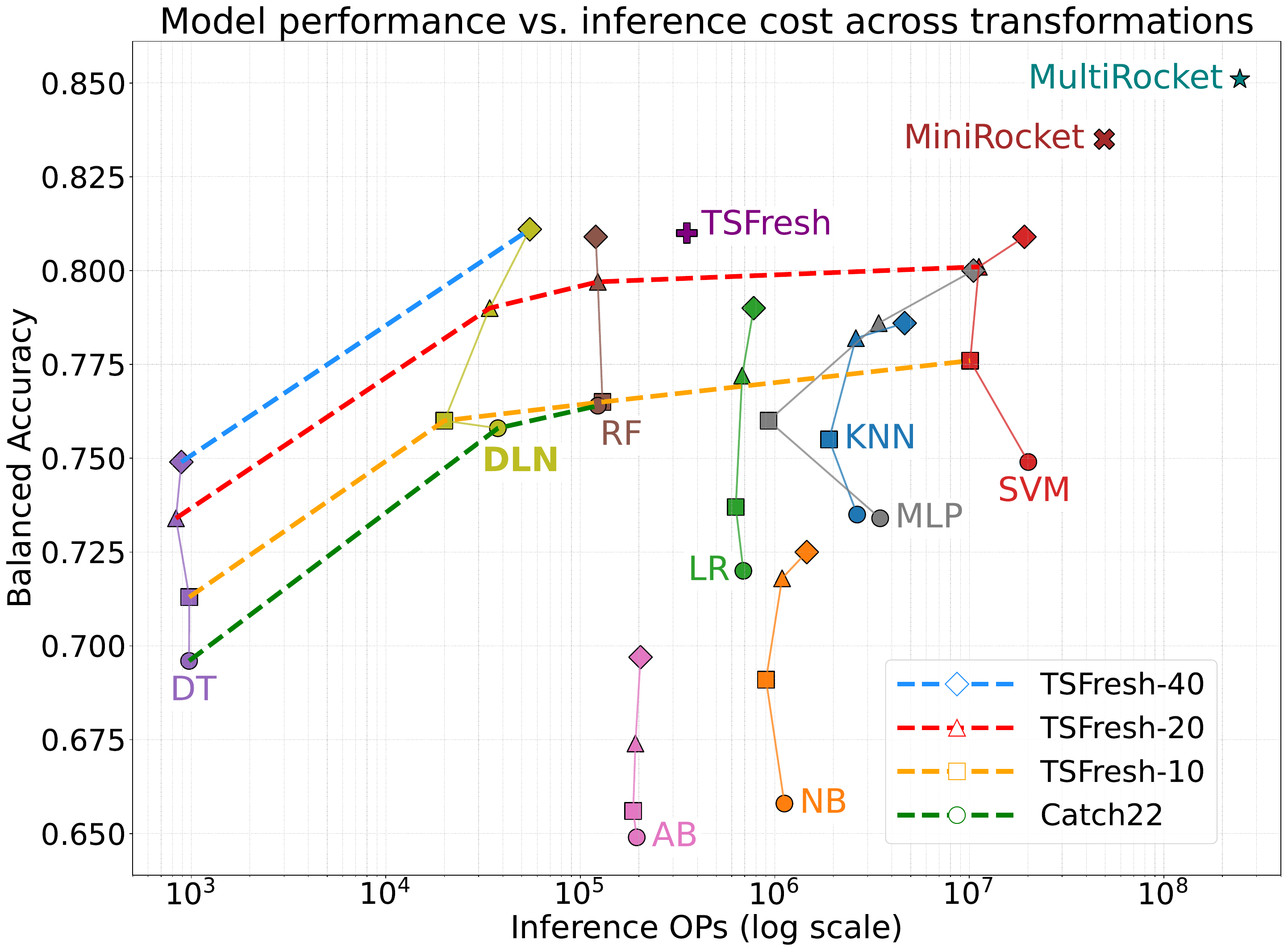}
\caption{Average balanced accuracy (best of 10 runs) versus inference operation cost for all 
models across 51 datasets. Results are shown for the Catch22, TSFresh-10, TSFresh-20, and 
TSFresh-40 transformations, with the Pareto frontier drawn for each case. 
SOTA time-series specific methods (the full TSFresh 
implementation, MiniRocket, and MultiRocket) are also included to contextualize the 
broader accuracy-cost landscape. Notably, DLN with TSFresh-40 outperforms the full TSFresh 
implementation in both accuracy and computational efficiency. The DLN is 
Pareto-optimal in every instance, and is the top-performing model with the TSFresh-40 
transformation among the general classifiers}
\label{pareto}
\Description{Scatter plot of models where the x-axis is inference operation cost and the 
y-axis is average balanced accuracy (best of 10 runs). Points are grouped by 
transformation method (Catch22 and three TSFresh variants), with a Pareto frontier 
curve drawn for each group. DLN points lie on the Pareto frontier for all 
transformations and reach the highest accuracy region for TSFresh-40.}
\end{figure}

We evaluate our TSC-DLN framework against eight baseline methods on 51 TSC
datasets from the UCR Time-Series Classification Archive~\cite{dau2019ucr}. The baselines 
are $k$-nearest neighbors~(KNN), Gaussian Naive Bayes~(NB), logistic regression~(LR), 
support vector machine~(SVM), decision tree~(DT), AdaBoost~(AB), random forest~(RF), and 
a multilayer perceptron~(MLP). 
In addition, we contrast DLNs against SOTA time-series models, 
including the full TSFresh implementation, MiniRocket, and MultiRocket. 
We also conduct a targeted feasibility study on three datasets from the UEA
multivariate archive~\cite{bagnall2018uea} to examine whether the same feature-based DLN framework extends to
multiple channels.
In Appendix~\ref{appendix-datasets}, Table~\ref{datasets-stats} 
provides the characteristics (training set size, test set size, number of classes, and number of continuous and categorical
features for Catch22 and TSFresh variants) of each dataset after transformation and processing. Sample sizes 
range from hundreds to thousands and input dimensions range from 10 to 40. We evaluate 
models based on balanced accuracy and inference cost and illustrate the DLN's 
interpretability by providing examples of its decision-making processes. As summarized in 
Fig.~\ref{pareto}, the DLN lies on the Pareto frontier under every transformation setting.
Inference OPs refers to the number of logic-gate operations in the model 
(training is a one-time offline cost, whereas inference OP cost dictates 
edge deployment feasibility).

\subsection{Experimental Setup}
We select 51 datasets from the 128 available in the UCR Archive based on the 
following criteria:
\begin{itemize}
    \item The dataset has no empty dimensions.
    \item No missing values are present.
    \item All sequences in both the training and test sets have the same length.
    \item The training set contains between 100 and 10,000 samples.
    \item The test set contains at least 100 samples.
    \item Every class has at least 5 samples in both the training and test sets.
\end{itemize}
Because the UCR Archive already provides train-test splits, we use those splits directly.

Each dataset is then preprocessed as follows:
\begin{enumerate}
    \item Convert every time-series sequence into a feature vector with the chosen 
    transformation method.
    \item Remove columns that contain NaNs (not a number) after transformation, as some features may not be 
    applicable to every dataset.
    \item Remove duplicate rows and constant features.
    \item When feature selection is required (e.g., for TSFresh), rank features in the 
    training set with cross-validated random forests and keep the most informative subset.
    \item One-hot encode categorical variables; we treat both originally categorical 
    variables and continuous variables with too few unique values as categorical.
    \item Clip outliers and apply min-max scaling to continuous features so they fall 
    within the \([0,1]\) range.
\end{enumerate}
Detailed dataset characteristics after selection, transformation, and preprocessing are 
provided in Table~\ref{datasets-stats}.

For every dataset-transformation-model combination, we run 10 independent experiments, each 
with a distinct random seed. Each run begins with a hyperparameter search in which 
Optuna~\cite{optuna_2019} samples configurations from a predefined grid; the configuration 
that yields the highest cross-validated balanced accuracy is retained. For each model type, 
we use a fixed model-specific HPO search space across
all dataset-transformation pairs. We employ two to 
four cross-validation folds, allocating more folds to smaller datasets. The search space 
includes parameters for both model structure (e.g., DLN's hidden sizes and DT's 
maximum height) and learning optimizations (e.g., DLN's learning rate and SVM's 
kernel type). After selecting the optimal hyperparameters, we train the model and then 
evaluate it on the held-out test split. We use scikit-learn~\cite{scikit-learn} for 
traditional models and PyTorch~\cite{NEURIPS2019_9015} for neural network models. 
In addition, aeon~\cite{aeon24jmlr} handles dataset loading and basic time-series 
processing, scikit-learn manages tabular preprocessing, and Ray~\cite{moritz2018ray, 
liaw2018tune} orchestrates parallel HPO and training jobs.

\subsection{Accuracy}
For each dataset-transformation-model combination, we record the test balanced accuracies 
of 10 independent runs. We are interested in both the numerical values and the \emph{ease} 
of finding a well-performing model, which is reflected, for example, in the best 
accuracy achieved across 10 runs. We report the best-out-of-10 and mean accuracies 
and corresponding ranks in Tables~\ref{results-best} and~\ref{results-mean}. 
For each transformation type, we present the average across all datasets.

Overall, the DLN performs well. In inference OPs, only DT outranks it. In best-out-of-10 balanced test accuracy, it ranks
first, second or third among general classifiers. It improves considerably with more independent HPO-and-training runs. 
It is particularly competitive 
among non-ensemble methods, outperforming the MLP in three cases and tying it once. Among the transformations, 
TSFresh yields better results than Catch22 on an average, and model performance generally improves as more 
features are used. 

When compared to SOTA time-series-specific methods, DLN remains 
highly competitive. The full-size TSFresh implementation (using over 750 features) 
achieves an average accuracy of 0.810 but requires 353K inference OPs. Our DLN with TSFresh-40 
slightly outperforms this accuracy (0.811) while requiring only 55K OPs, demonstrating DLN's 
superior feature selection and sparsity. While MiniRocket and MultiRocket achieve higher absolute 
accuracies (0.835 and 0.851, respectively), they require 49.4M and 246M operations, respectively.
DLN+TSFresh-40 thus secures a critical point on the Pareto frontier, trading a few percentage 
points of accuracy for nearly a 1000x to 4000x reduction in computational cost.

Traditional methods like RF and SVM achieve 
strong accuracy. The DLN paired with the TSFresh-40 transformation emerges as the 
highest-performing combination when evaluated over multiple runs.
Detailed results for each dataset-model pair for the Catch22, TSFresh-10, TSFresh-20, 
and TSFresh-40 transformations are reported in Appendices~\ref{appendix-catch22}, 
\ref{appendix-tsfresh10}, \ref{appendix-tsfresh20}, and~\ref{appendix-tsfresh40}, 
respectively.

We plot the \textit{Best@k} curves for the five strongest models, defining
\textit{Best@k} as the expected maximum test accuracy from $k$ draws (without replacement)
out of the 10 independent runs. Each run comprises a complete HPO-and-training experiment;
thus, $k$ counts independent runs, not individual Optuna trials or models jointly evaluated
at inference. This metric complements the mean and maximum by showing how the expected
best result changes as more independent runs are available.
The closed-form expression is
\[
\mathbb{E}\left[\text{Best@k}\right] \;=\;
\sum_{i=k}^{n} \frac{\binom{i-1}{k-1}}{\binom{n}{k}}\,x_{(i)},
\]
where $n=10$ is the total number of runs and $x_{(i)}$ denotes the $i$-th order statistic 
(accuracies sorted in ascending order). The \textit{Best@k} is the mean when $k = 1$ and 
the best-out-of-10 when $k = 10$. We present the \textit{Best@k} curves after the 
TSFresh-40 transformation in Fig.~\ref{best-k-tsfresh40} and provide the curves for the 
remaining transformations in the Appendix. 
As can be observed from Fig.~\ref{best-k-tsfresh40}, DLN accuracy is initially 
lower than SVM and RF when $k < 4$. This occurs because DLN possesses a highly complex, 
discrete search space consisting of structural hyperparameters and logic relaxations, 
making it more susceptible to settling in sub-optimal local minima during independent HPO-and-training runs. 
Conversely, traditional models like SVM and RF have smoother, more continuous search 
spaces that are easier to optimize quickly. However,
DLN exhibits the fastest-growing curve, 
indicating that running more independent HPO-and-training runs is generally beneficial for finding a high-performing 
DLN. It also achieves the best overall performance when $k \ge 7$.
Increasing $k$ affects only the offline search budget: each additional value of $k$
adds another complete HPO-and-training run; hence, offline computation grows approximately linearly.
This analysis does not alter deployment: it is not an ensemble, and inference uses a single
selected model. Therefore, $k$ does not change the deployed model size, inference OP count, or
inference latency. The curve should consequently be interpreted as an optimization-sensitivity
analysis rather than an inference procedure.

\begin{table}[ht]
\centering
\caption{\textbf{Best-out-of-10} test balanced accuracy and corresponding inference operations (OPs), 
with rank (R: lower is better) computed for both. All values are averaged across datasets: accuracy and rank
use an arithmetic mean, while OPs use a geometric mean. OPs are calculated assuming FP16 
for floating-point and INT16 for integer arithmetic.}
\label{results-best}
\small{
\setlength{\tabcolsep}{6pt}                
\renewcommand{\arraystretch}{1.25}         
\begin{adjustbox}{max width=\linewidth}
\begin{tabularx}{\linewidth}{ll*{9}{Y}}
\toprule
 &  & \multicolumn{9}{c}{\textbf{Model}}\\
\cmidrule(lr){3-11}
Transform. & Metric & KNN & NB & LR & SVM & DT & RF & AB & MLP & DLN\\
\midrule
\multirow{4}{*}{Catch22}
 & Acc $\uparrow$      & 0.735 & 0.658 & 0.720 & 0.749 & 0.696 & 0.764 & 0.649 & 0.734 & 0.758\\
 & Acc R $\downarrow$   & 4.75  & 7.83  & 5.52  & 3.54  & 6.76  & 2.50  & 6.94  & 4.65  & 2.50\\
\arrayrulecolor{lightgray}\cmidrule(lr){3-11}\arrayrulecolor{black}
 & Cost $\downarrow$     & 2.65M & 1.12M & 689K  & 20.1M & 974 & 123K  & 195K  & 3.48M & 37.7K\\
 & Cost R $\downarrow$  & 6.92  & 6.41  & 4.96  & 8.76  & 1.00  & 3.22  & 3.90  & 7.69  & 2.14\\
\midrule
\multirow{4}{*}{TSFresh-10}
 & Acc $\uparrow$      & 0.755 & 0.691 & 0.737 & 0.776 & 0.713 & 0.765 & 0.656 & 0.760 & 0.760\\
 & Acc R $\downarrow$   & 4.55  & 7.23  & 5.12  & 2.51  & 6.93  & 3.41  & 7.09  & 4.14  & 4.03\\
\arrayrulecolor{lightgray}\cmidrule(lr){3-11}\arrayrulecolor{black}
 & Cost $\downarrow$     & 1.90M & 901K  & 630K  & 10.1M & 976 & 130K  & 187K  & 929K  & 20.0K\\
 & Cost R $\downarrow$  & 7.25  & 6.96  & 5.41  & 8.47  & 1.00  & 3.37  & 4.02  & 6.51  & 2.00\\
\midrule
\multirow{4}{*}{TSFresh-20}
 & Acc $\uparrow$      & 0.782 & 0.718 & 0.772 & 0.801 & 0.734 & 0.797 & 0.674 & 0.786 & 0.790\\
 & Acc R $\downarrow$   & 4.56  & 7.35  & 4.99  & 2.88  & 6.96  & 3.33  & 7.10  & 4.32  & 3.50\\
\arrayrulecolor{lightgray}\cmidrule(lr){3-11}\arrayrulecolor{black}
 & Cost $\downarrow$     & 2.61M & 1.09M & 679K  & 11.2M & 834 & 123K  & 192K  & 3.42M & 34.2K\\
 & Cost R $\downarrow$  & 7.14  & 6.39  & 5.04  & 8.51  & 1.00  & 3.24  & 3.86  & 7.75  & 2.08\\
\midrule
\multirow{4}{*}{TSFresh-40}
 & Acc $\uparrow$      & 0.786 & 0.725 & 0.790 & 0.809 & 0.749 & 0.809 & 0.697 & 0.800 & 0.811 \\
 & Acc R $\downarrow$   & 5.10 & 7.66 & 5.14 & 3.05 & 6.92 & 3.62 & 6.66 & 3.96 & 2.90 \\
\arrayrulecolor{lightgray}\cmidrule(lr){3-11}\arrayrulecolor{black}
 & Cost $\downarrow$     & 4.65M & 1.46M & 779K & 19.2M & 889 & 120K & 204K & 10.5M & 55.0K \\
 & Cost R $\downarrow$  & 7.14 & 6.20 & 4.86 & 8.47 & 1.00 & 3.12 & 3.90 & 8.12 & 2.20 \\
\bottomrule
\end{tabularx}
\end{adjustbox}
}
\end{table}

\begin{table}[ht]
\centering
\caption{\textbf{Mean} test balanced accuracy and inference operations (OPs) averaged over 10 random 
seeds, with ranks computed for both metrics. Accuracy and rank are arithmetically averaged 
across datasets, while OPs are geometrically averaged. The OP calculation assumes FP16 for 
floating-point and INT16 for integer arithmetic.}
\label{results-mean}
\small{
\begin{adjustbox}{max width=\linewidth}
\setlength{\tabcolsep}{6pt}
\renewcommand{\arraystretch}{1.25}
\begin{tabularx}{\linewidth}{ll*{9}{Y}}
\toprule
 &  & \multicolumn{9}{c}{\textbf{Model}}\\
\cmidrule(lr){3-11}
Transform. & Metric & KNN & NB & LR & SVM & DT & RF & AB & MLP & DLN\\
\midrule
\multirow{4}{*}{Catch22}
 & Acc $\uparrow$      & 0.721 & 0.657 & 0.705 & 0.731 & 0.672 & 0.743 & 0.616 & 0.707 & 0.731\\
 & Acc R $\downarrow$   & 4.31  & 7.22  & 5.03  & 3.10  & 6.99  & 2.78  & 6.91  & 5.54  & 3.12\\
\arrayrulecolor{lightgray}\cmidrule(lr){3-11}\arrayrulecolor{black}
 & Cost $\downarrow$     & 2.88M & 1.12M & 689K  & 20.9M & 1.00K    & 121K  & 212K  & 4.37M & 38.6K\\
 & Cost R $\downarrow$  & 6.92  & 6.31  & 4.88  & 8.82  & 1.00  & 3.00  & 4.24  & 7.76  & 2.06\\
\midrule
\multirow{4}{*}{TSFresh-10}
 & Acc $\uparrow$      & 0.743 & 0.691 & 0.726 & 0.759 & 0.690 & 0.745 & 0.626 & 0.732 & 0.735\\
 & Acc R $\downarrow$   & 3.97  & 6.21  & 4.65  & 2.41  & 7.09  & 3.66  & 7.29  & 4.98  & 4.75\\
\arrayrulecolor{lightgray}\cmidrule(lr){3-11}\arrayrulecolor{black}
 & Cost $\downarrow$     & 2.53M & 901K  & 630K  & 16.4M & 942 & 113K  & 205K  & 1.19M & 18.6K\\
 & Cost R $\downarrow$  & 7.33  & 6.65  & 5.18  & 8.86  & 1.00  & 3.08  & 4.18  & 6.73  & 2.00\\
\midrule
\multirow{4}{*}{TSFresh-20}
 & Acc $\uparrow$      & 0.768 & 0.718 & 0.759 & 0.786 & 0.711 & 0.777 & 0.647 & 0.762 & 0.767\\
 & Acc R $\downarrow$   & 4.14  & 6.54  & 4.64  & 2.72  & 7.24  & 3.56  & 7.02  & 4.92  & 4.24\\
\arrayrulecolor{lightgray}\cmidrule(lr){3-11}\arrayrulecolor{black}
 & Cost $\downarrow$     & 3.36M & 1.09M & 679K  & 18.3M & 901 & 115K  & 210K  & 3.90M & 35.5K\\
 & Cost R $\downarrow$  & 7.08  & 6.27  & 4.96  & 8.78  & 1.00  & 2.98  & 4.18  & 7.69  & 2.06\\
\midrule
\multirow{4}{*}{TSFresh-40}
 & Acc $\uparrow$      & 0.775 & 0.725 & 0.773 & 0.795 & 0.723 & 0.792 & 0.666 & 0.779 & 0.787 \\
 & Acc R $\downarrow$   & 4.74 & 6.84 & 4.86 & 2.77 & 7.26 & 3.61 & 6.84 & 4.31 & 3.75 \\
\arrayrulecolor{lightgray}\cmidrule(lr){3-11}\arrayrulecolor{black}
 & Cost $\downarrow$     & 5.73M & 1.46M & 779K & 24.3M & 901 & 117K & 208K & 13.4M & 63.7K \\
 & Cost R $\downarrow$  & 7.18 & 6.10 & 4.94 & 8.63 & 1.00 & 2.94 & 4.06 & 8.04 & 2.12 \\
\bottomrule
\end{tabularx}
\end{adjustbox}
}
\end{table}

\begin{figure}[!htbp]
\centering
\includegraphics[width=0.6\columnwidth]{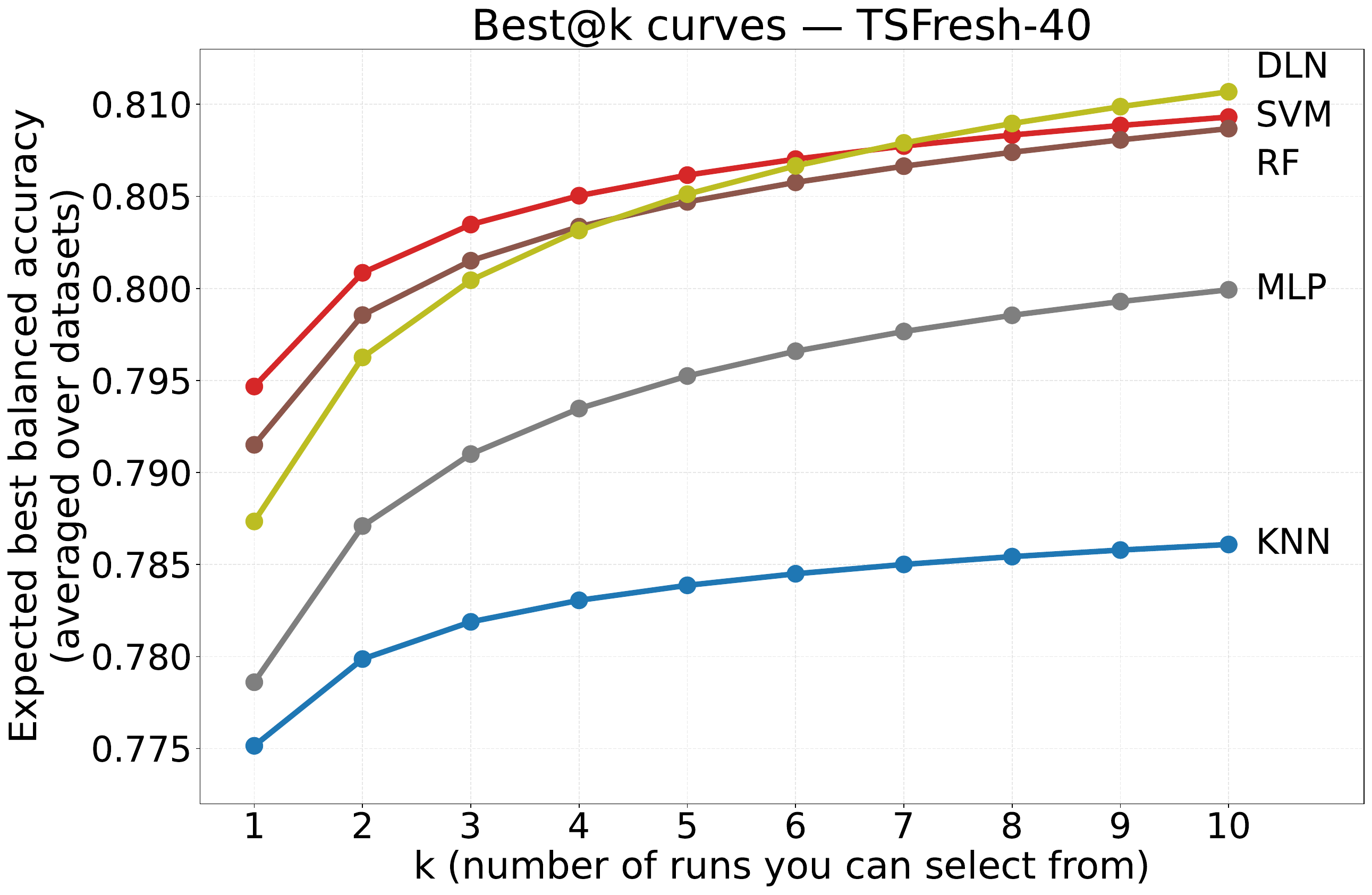}
\caption{\textit{Best@k} curves for the five strongest models using the TSFresh-40 
transformation. The DLN curve exhibits the fastest growth and achieves the best overall 
performance, given a sufficient number of independent HPO-and-training runs ($k \ge 7$).}
\label{best-k-tsfresh40}
\Description{Line plot of Best@k performance for the five strongest models using 
TSFresh-40 features. The x-axis is the number of independent HPO-and-training runs $k$ and the y-axis is the best 
balanced accuracy achieved among the first $k$ runs. The DLN curve rises more quickly 
than the other models and attains the highest accuracy once $k$ is at least 7.}
\end{figure}

The final accuracy of this two-stage pipeline reflects both the time-series
representation and the downstream classifier. Across the evaluated general classifiers, moving from
Catch22 to TSFresh-40 increases mean balanced accuracy by 0.049-0.072; for DLN, the increase is
0.056 (from 0.731 to 0.787). Thus, the richer TSFresh representation materially contributes to the
observed performance. Conversely, within each transformation, the general classifiers are evaluated
on the same selected and preprocessed feature matrix; hence, their differences reflect the downstream
classifier conditional on that representation. The results should, therefore, be interpreted as showing
that the observed performance jointly arises from the compact statistical descriptors and the
downstream logical classifier. TSFresh-40 is the strongest feature budget among
those evaluated, but is not necessarily universally optimal. More broadly, many TSC methods follow
a transformation-plus-classifier paradigm, although convolutional, shapelet, and dictionary
transformations preserve different forms of temporal information. A sequence-native DLN would
require a new temporal architecture rather than simply flattening raw samples, and is left to future
work.

The dataset-level results reveal three distinct failure modes. A persistent gap to
another classifier using the same features, as on ChlorineConcentration under TSFresh-40
(DLN: 0.646; SVM: 0.776), indicates a downstream classifier or optimization limitation. When all
same-feature models perform poorly, as on Haptics (all mean accuracies below 0.47), the limitation
is shared by the representation or task. A large mean-to-best gap, as on Earthquakes
(0.643 versus 0.718), instead indicates optimization sensitivity. Exploratory correlations across
the four transformations further show that DLN's relative gap tends to widen as the number of
classes increases and narrow as the number of training
examples per class increases; smaller datasets also show
greater run-to-run variation. In contrast, sequence length does not significantly predict the
same-feature DLN gap. These are aggregate tendencies rather than deterministic rules.

\subsection{Efficiency}
We assess inference efficiency by quantifying the total number of basic hardware logic-gate 
operations (OPs) required for a single prediction, following the methodology established 
in the original DLN article~\cite{10681646}. It involves translating a model's 
high-level computational steps (e.g., floating-point arithmetic, comparisons) into their 
fundamental logic gate equivalents. For this conversion, we define the cost of each gate 
type: a two-input AND, OR, NAND, or NOR gate counts as one OP, whereas a two-input XOR or 
XNOR gate counts as three OPs. NOT gates are considered to have zero cost. To ensure a 
fair comparison across all models, our analysis assumes a standardized precision of 16 bits 
for both floating-point and integer data types. While models are trained using larger 
default precisions (e.g., float32 or float64), they typically maintain performance when 
downscaled. This assumption creates a consistent baseline but also conservatively estimates 
the advantage of logic-heavy models like the DLN, whose efficiency benefits are more 
pronounced when using larger, more costly data types.

The operational costs are also summarized in Tables~\ref{results-best} and~\ref{results-mean}. 
DLNs are highly efficient, second only to single decision trees in all cases, and lie on 
the Pareto frontier in the accuracy-versus-inference-cost plot shown in Fig.~\ref{pareto}. 
On an average, they are several times more computationally efficient than strong-performing 
models like random forests and orders of magnitude more efficient than SVMs. The sizes of 
NN-based models (i.e., DLN and MLP) are more impacted by the input size because their HPO 
search spaces are defined based on the input dimension.

In addition to theoretical OP counts, we measured single-threaded CPU 
inference latency to provide a concrete baseline for edge deployment. 
Table~\ref{results-latency} compares the latency of DLN and MLP. We evaluated both models 
on the test splits across all dataset-seed pairs, forcing single-threaded execution. 
For DLN, we used a compiled C backend utilizing bit-parallel logic evaluation, 
while MLP was evaluated using PyTorch. We omit traditional baselines (e.g., scikit-learn 
and aeon models) from this latency comparison because their standard software 
implementations are not highly optimized for inference speed and exhibit higher latency. 
The DLN exhibits significantly lower latency, 
particularly at a batch size of 1, which is the most common scenario for real-time 
edge devices. While the gap narrows at a batch size of 32, DLN still maintains an edge. 
It is important to emphasize that CPUs are heavily optimized for floating-point MAC 
operations rather than the bitwise logic that DLNs utilize. 
These wall-clock measurements are implementation- and platform-dependent
and do not establish FPGA- or ASIC-level performance; hardware synthesis
and accelerator evaluation remain future work.

\begin{table}[ht]
\centering
\caption{Classifier-stage inference latency ($\mu$s/sample) on CPU. Values are the geometric mean 
across all dataset-seed pairs, measured using single-threaded execution.}
\label{results-latency}
\small{
\begin{tabularx}{0.4\columnwidth}{l Y Y}
\toprule
Batch Size & MLP & DLN \\
\midrule
Batch = 1  & 72.0 \(\mu\text{s}\) & 23.8 \(\mu\text{s}\) \\
Batch = 32 & 3.15 \(\mu\text{s}\) & 2.78 \(\mu\text{s}\) \\
\bottomrule
\end{tabularx}
}
\end{table}

We separately measured the deployment-time feature transformation. Catch22 required approximately
0.817~ms/sample, whereas computing the selected TSFresh-40 calculators required approximately
11.5~ms/sample, with both values reported as geometric means. Thus, the current TSFresh-40
pipeline is feature-extraction-bound: transformation takes milliseconds, compared with tens of
microseconds for the classifier. This bottleneck belongs to the modular feature 
front-end rather than the DLN classifier. Feature
extraction can be parallelized across samples and channels, and repeated or overlapping windows may
permit caching or incremental reuse of intermediate statistics. Highly vectorized convolutional
transformations such as ROCKET and MiniRocket also provide alternative fixed-dimensional
representations, although their outputs require dimensionality control before DLN training. Hence,
front-end improvements can be adopted without changing the discrete logical classifier.

\subsection{Hyperparameter Analysis}
Unlike previous tabular DLN studies, which fix many model-training configurations before 
HPO and ablate them afterward, we include these configurations directly in the HPO search 
space. They are:
\begin{itemize}
    \item \texttt{phase\_unified}: whether to learn functionality and connections 
    simultaneously or alternately; a value of~1 denotes simultaneous learning.
    \item \texttt{ste\_threshold\_layer}: whether to apply STE when training the 
    \textit{ThresholdLayer}.
    \item \texttt{ste\_logic\_layer}: whether to apply STE when training the 
    \textit{LogicLayer}.
    \item \texttt{ste\_sum\_layer}: whether to apply STE when training the \textit{SumLayer}.
    \item \texttt{subset\_gate\_num}: the number of candidate logic gates considered when learning a 
    logic neuron's functionality. Options are 16, 8, and~4, where 16 is the full set.
    \item \texttt{subset\_link\_num}: the number of candidate input links evaluated for each 
    of the two links ($a$ and $b$) in every logic neuron. Options are 16, 8, 4, 2, and 1; 
    a value of~1 initializes a link once and then keeps it fixed.
    \item \texttt{concat\_input}: whether to concatenate an input-\textit{ThresholdLayer} 
    pair to each hidden \textit{LogicLayer}.
\end{itemize}

\begin{figure}[!htbp]
\centering
\includegraphics[width=\linewidth]{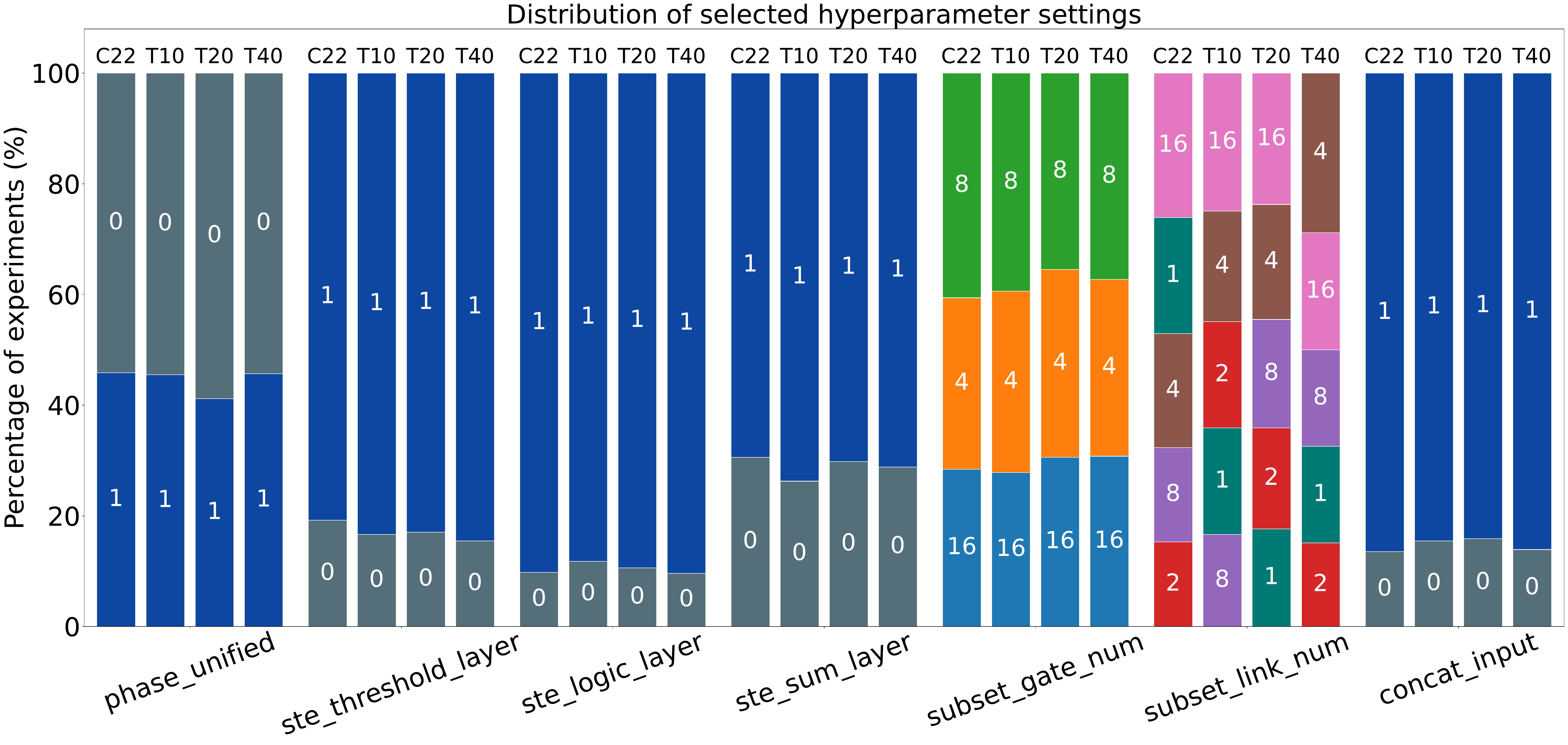}
\caption{Distribution of model training configurations selected by the HPO algorithm for 
each of the four transformation methods. The statistics are aggregated across all 
(dataset, random seed) pairs, with each bar representing results from $51 \times 10 = 510$ experiments.}
\Description{Grouped bar chart summarizing how often different training configurations 
are selected by the hyperparameter optimization algorithm for each of the four feature 
transformation methods. For each method, a bar (or group of bars) aggregates 510 
experiments, illustrating preferences for specific model hyperparameters across 
datasets and random seeds.}
\label{hpo-distr}
\end{figure}

We summarize the statistics and show the HPO algorithm’s final choices in 
Fig.~\ref{hpo-distr}. For each transformation type, 510 experiments are carried out in all. 
Our findings are consistent with those for tabular classification DLNs: (i) alternating 
training is slightly better than updating both parameter types jointly; (ii) applying STE 
to all three layer types is beneficial; and (iii) concatenating the ThresholdLayer's output 
is important. Furthermore, for the logic neurons, gate subsetting is usually selected, and 
it is preferable \emph{not} to use an initialize-then-fix strategy for the links. Based on 
the combination statistics, a configuration with 8 candidate gates and 16 candidate links 
is the most common, although no single setting is significantly more frequent than the others.
Crucially, these distributions reveal a strong domain-agnostic stability. 
The optimal training configurations for time-series data do not differ meaningfully 
from previous tabular findings.

\subsection{Interpretability}
\begin{figure}[!htbp]
    \centering
    \begin{subfigure}[b]{\linewidth}
        \centering
        \includegraphics[width=\linewidth]{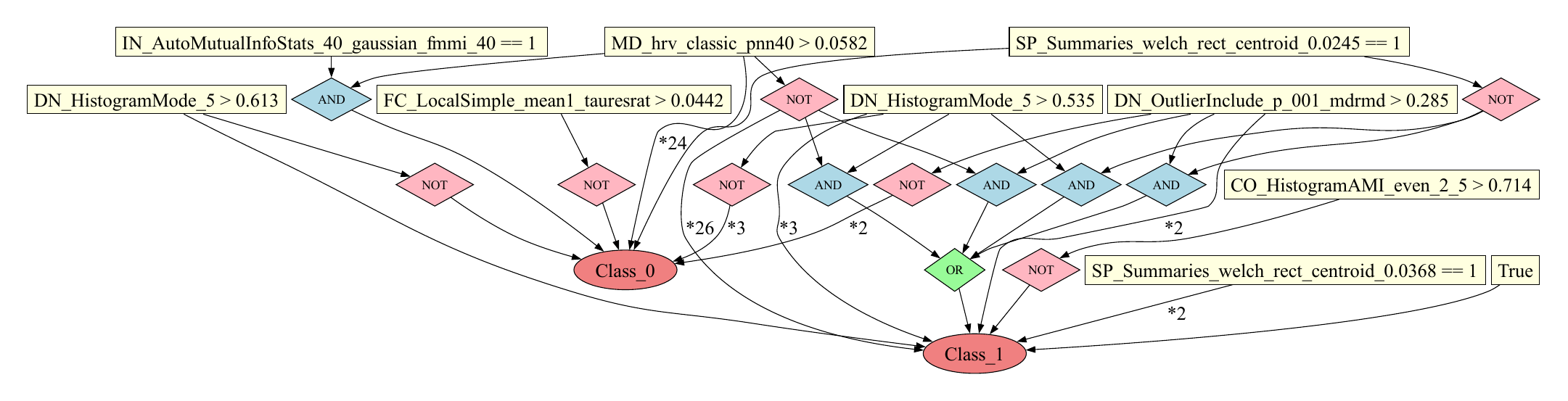}
        \caption{Decision process using the Catch22 transformation. The model achieves a 
        test balanced accuracy of 0.998 with 5 continuous and 3 categorical features from 
        an original set of 20 continuous and 5 one-hot features.}
        \label{viz-catch22}
    \end{subfigure}
    \hfill
    \begin{subfigure}[b]{\linewidth}
        \centering
        \includegraphics[width=\linewidth]{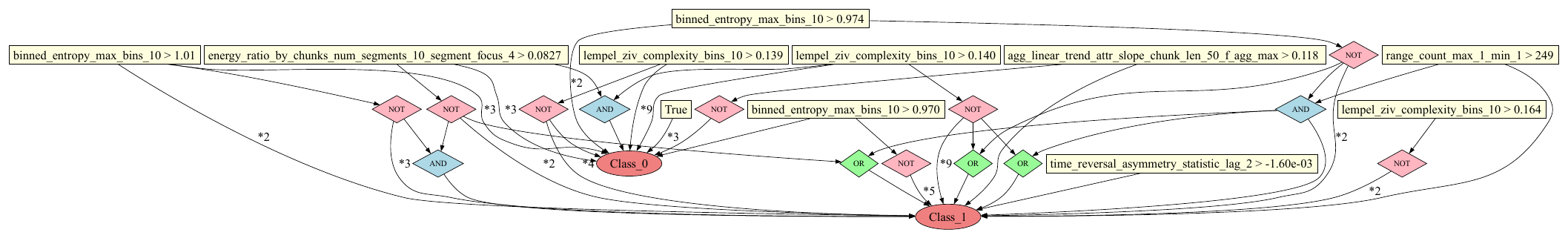}
        \caption{Decision process using the TSFresh-20 transformation on the same dataset 
        as in (a). The model achieves a test balanced accuracy of 0.944 with 6 continuous 
        features from an original set of 20 continuous features.}
        \label{viz-tsfresh20}
    \end{subfigure}
    \caption{Illustration of decision processes for DLN models trained on the 
    \textit{FreezerRegularTrain} dataset: (a) DLN resulting from the 
    Catch22 feature set, (b) DLN resulting from the TSFresh-20 feature set.}
    \Description{Two visualizations of DLN decision processes on the 
    \textit{FreezerRegularTrain} dataset. Subfigure (a) shows a compact network using 
    Catch22 features, highlighting 5 continuous and 3 categorical inputs contributing 
    to almost perfect test accuracy. Subfigure (b) shows a different but similarly 
    small network using 6 continuous TSFresh-20 features, with slightly lower accuracy. 
    In both diagrams, nodes represent threshold and logic operations applied to selected 
    features, and edges trace how feature conditions lead to the final class output.}
\end{figure}

A key strength of the DLN architecture is its inherent interpretability. Once trained, a 
DLN can be visualized directly as a transparent, rule-based system. We illustrate this 
with models learned on the \textit{FreezerRegularTrain} dataset using Catch22 and 
TSFresh-20 features in Figs.~\ref{viz-catch22} and~\ref{viz-tsfresh20}, respectively. In 
these diagrams, the decision-making path is explicit. The process begins with input 
features (yellow rectangles), which are binarized against learned thresholds. These binary 
signals then propagate through a network of logic operators (diamonds) that formulate 
logic rules. The final output is an aggregation of the outputs from these rules. These 
networks are not only interpretable but also achieve strong performance while using only a 
small fraction of the available features, mirroring the feature selection behavior observed 
for tabular DLNs~\cite{10681646, yue2025rdln}.

\subsection{Multivariate Feasibility Study}\label{sec-multivariate}
We conduct a targeted first-step extension on \textit{RacketSports}, \textit{NATOPS}, and
\textit{Libras} from the UEA multivariate archive. As summarized in
Table~\ref{multivariate-datasets}, they span 2-24 channels and 4-15 classes and represent
complementary settings: wearable inertial strokes, articulated-body gestures, and two-dimensional
hand trajectories. Their short, fixed-length sequences and official train-test splits also avoid
additional alignment or missing-value preprocessing, making them suitable for a controlled
feasibility study.

\begin{table}[ht]
\centering

\caption{Characteristics of the multivariate feasibility datasets.}
\label{multivariate-datasets}
\small{
\begin{adjustbox}{max width=\linewidth}
\begin{tabular}{lrrrrrr}
\toprule
Dataset & Train & Test & Length & Channels & Classes & C22 dim. \\
\midrule
RacketSports & 151 & 152 & 30 & 6 & 4 & 132 \\
NATOPS       & 180 & 180 & 51 & 24 & 6 & 528 \\
Libras       & 180 & 180 & 45 & 2 & 15 & 44 \\
\bottomrule
\end{tabular}
\end{adjustbox}
}
\end{table}

For each channel, we independently extract its 22 Catch22 features and concatenate the
channel-level vectors, yielding 44, 132, and 528 initial features for Libras, RacketSports,
and NATOPS, respectively. We then apply the same training-set-based, cross-validated RF
ranking used in the univariate pipeline and retain the top 40 features globally across
channels. The resulting vector is processed with the same cleaning, encoding, clipping, and scaling
pipeline and supplied to the unchanged classifiers. DLN can, therefore, combine summary features originating from
different channels, although this front-end does not explicitly preserve fine-grained
cross-channel temporal alignment.

This experiment is intended as a feasibility study rather than a comprehensive UEA benchmark.
For each dataset-model pair, we use eight HPO trials from reduced search spaces and
repeat the complete HPO-and-training procedure with five random seeds. Table~\ref{multivariate-results}
reports the mean and best observed test balanced accuracies. 
Across the three multivariate datasets, DLN remains competitive with the strongest baselines. Although the datasets and feature front-ends differ, these results are consistent with its competitive performance in the univariate experiments and support the applicability of the channel-wise extension to multivariate TSC.

\begin{table}[ht]
\centering

\caption{Mean and best observed test balanced accuracy over five random seeds in the multivariate
feasibility study. All models use channel-wise Catch22 followed by global top-40 feature selection.}
\label{multivariate-results}
\small{
\setlength{\tabcolsep}{5pt}
\renewcommand{\arraystretch}{1.15}
\begin{adjustbox}{max width=\textwidth}
\begin{tabular}{llcccccccc}
\toprule
Dataset & Metric & KNN & NB & LR & SVM & DT & RF & MLP & DLN \\
\midrule
\multirow{2}{*}{RacketSports}
 & Mean & 0.788 & 0.736 & 0.750 & 0.755 & 0.622 & \textbf{0.814} & 0.769 & 0.743 \\
 & Best & 0.798 & 0.736 & 0.779 & 0.770 & 0.672 & \textbf{0.834} & 0.790 & 0.816 \\
\midrule
\multirow{2}{*}{NATOPS}
 & Mean & 0.781 & 0.787 & 0.798 & 0.828 & 0.792 & 0.814 & 0.812 & \textbf{0.840} \\
 & Best & 0.806 & 0.794 & 0.817 & 0.850 & 0.811 & 0.828 & 0.828 & \textbf{0.861} \\
\midrule
\multirow{2}{*}{Libras}
 & Mean & 0.836 & 0.777 & 0.848 & \textbf{0.850} & 0.576 & 0.824 & 0.819 & 0.847 \\
 & Best & 0.844 & 0.778 & 0.861 & 0.867 & 0.689 & 0.839 & 0.850 & \textbf{0.878} \\
\midrule
\multirow{2}{*}{Average}
 & Mean & 0.801 & 0.766 & 0.799 & 0.811 & 0.663 & \textbf{0.818} & 0.800 & 0.810 \\
 & Best & 0.816 & 0.769 & 0.819 & 0.829 & 0.724 & 0.834 & 0.823 & \textbf{0.852} \\
\bottomrule
\end{tabular}
\end{adjustbox}
}
\end{table}

\subsection{Limitations}
Although exceptionally fast at inference (achieving microsecond-level latency), 
DLN training is computationally demanding. This is because the summation of real-valued 
logic operations in each logic neuron is not specifically accelerated by CUDA; 
this calculation also relies on the output of three preceding softmax functions. 
This makes directly applying DLNs to massively high-dimensional problems 
(such as the 10,000+ features generated by ROCKET) challenging without prior 
dimensionality reduction. 

In our experiments, due to these computational limits, we used only up to 40 
continuous features selected from the TSFresh transformation, which natively 
extracts over 750 features. However, we found that this aggressive feature 
selection actually matched or exceeded the accuracy of the full TSFresh 
feature set, highlighting DLN's strong synergy with sparse inputs. 

Furthermore, in the HPO stage, DLN has a large discrete search space due to its 
many structural configurations. HPO is more efficient on a refined search space 
when the number of trials is limited. While this currently requires a high upfront 
training cost, our discovery of domain-agnostic training stability provides a 
strong foundation for defining a refined, universally optimal search space in 
future work to alleviate this bottleneck.

The current end-to-end pipeline can also be feature-extraction-bound,
particularly with TSFresh. Parallel extraction and caching or incremental reuse can improve
throughput in applicable workloads, but low-latency single-sample transformation remains an
open problem.

The current feature-based design also cannot recover local ordering or temporal
alignment that is not retained by the transformation, and a sequence-native DLN is not evaluated.

Finally, the multivariate study covers only three short, fixed-length datasets
and uses channel-wise summary features. It validates basic applicability but does not establish
archive-wide performance or explicitly model fine-grained cross-channel temporal interactions.

\section{Conclusion and Future Directions}\label{sec-conclusion}
We tested the applicability of DLNs to univariate TSC by using 
feature-based transformations to convert sequence data into vectors. Our experiments on 51 
datasets show that the DLN is interpretable and efficient while maintaining high 
performance, similar to its performance on tabular tasks.

One promising research direction is extending DLN training to high-dimensional datasets. 
The DLN shows strong performance on transformations that output the highest number of 
features. Hence, developing efficient DLN training software for such inputs is an important 
next step. 
Because DLN demonstrates success on feature extractors like TSFresh, 
it is well-positioned for adaptation to features extracted by SOTA convolutional methods 
like ROCKET, as well as shapelet- or dictionary-based transformations, which similarly 
convert time-series into tabular formats but produce high-dimensional vectors.
A complementary direction is to optimize the complete raw-sequence-to-prediction
pipeline through vectorized and parallel feature extraction, caching or incremental updates for
overlapping windows, and joint selection of compact front-ends and DLN architectures.
In addition, the rapid growth in accuracy shown by DLN's \textit{Best@k} 
curves suggests that a well-performing configuration exists for many datasets. Developing 
efficient search methods for these optimal configurations is a key area for future work. 
This observation also suggests that a trained DLN's performance is sensitive to its 
initialization. Therefore, investigating how structural initialization, weight 
initialization, and multiple runs can boost DLN performance is another interesting 
research avenue.

A sequence-native DLN is another important direction. Rather than flattening
raw samples, such an architecture would require temporal weight sharing, local logical receptive
fields, and pooling or variable-length aggregation. For multivariate data, broader UEA evaluation
and cross-channel logical transformations that retain temporal interactions would extend the
channel-wise feasibility study presented here.

\section*{Acknowledgments}
This work was supported by NSF under Grant No. CCF-2416541.

\bibliographystyle{unsrtnat}
\bibliography{references}

\appendix
\vspace{1.6cm}
\noindent\large\textbf{Appendix}

\section{Dataset Characteristics after Transformation and Processing}\label{appendix-datasets}
Table~\ref{datasets-stats} summarizes the characteristics of each dataset after each
transformation. The majority of the features extracted by Catch22 and TSFresh are 
continuous. For each dataset, we remove features that are not applicable or whose values 
are constant across all samples. Features with only a few unique values are treated as categorical.

\begin{table}
\centering
\caption{Dataset characteristics after transformation and preprocessing.}
\label{datasets-stats}
\footnotesize{
\begin{adjustbox}{max width=\linewidth}
\begin{tabular}{lrrrrrrrrrrr}
\toprule
 & & & & \multicolumn{2}{c}{Catch22} & \multicolumn{2}{c}{TSFresh-10} & \multicolumn{2}{c}{TSFresh-20} & \multicolumn{2}{c}{TSFresh-40} \\
dataset & \# train & \# test & \# class & con. & cat. & con. & cat. & con. & cat. & con. & cat. \\
\midrule
ACSF1 & 100 & 100 & 10 & 15 & 14 & 10 & 0 & 20 & 0 & 40 & 0 \\
ChlorineConcentration & 467 & 3840 & 3 & 22 & 0 & 10 & 0 & 20 & 0 & 40 & 0 \\
Computers & 250 & 250 & 2 & 21 & 0 & 10 & 0 & 20 & 0 & 40 & 0 \\
CricketX & 390 & 390 & 12 & 22 & 0 & 10 & 0 & 20 & 0 & 40 & 0 \\
CricketY & 390 & 390 & 12 & 22 & 0 & 10 & 0 & 20 & 0 & 40 & 0 \\
CricketZ & 390 & 390 & 12 & 22 & 0 & 10 & 0 & 20 & 0 & 40 & 0 \\
Earthquakes & 322 & 139 & 2 & 22 & 0 & 6 & 4 & 14 & 4 & 34 & 4 \\
EOGHorizontalSignal & 362 & 362 & 12 & 22 & 0 & 10 & 0 & 20 & 0 & 40 & 0 \\
EOGVerticalSignal & 362 & 362 & 12 & 22 & 0 & 10 & 0 & 20 & 0 & 40 & 0 \\
EthanolLevel & 504 & 500 & 4 & 19 & 0 & 10 & 0 & 20 & 0 & 40 & 0 \\
FaceAll & 560 & 1688 & 14 & 22 & 0 & 10 & 0 & 20 & 0 & 40 & 0 \\
Fish & 175 & 175 & 7 & 19 & 0 & 10 & 0 & 20 & 0 & 39 & 3 \\
FordA & 3601 & 1320 & 2 & 22 & 0 & 10 & 0 & 20 & 0 & 40 & 0 \\
FordB & 3636 & 810 & 2 & 22 & 0 & 9 & 0 & 19 & 0 & 39 & 0 \\
FreezerRegularTrain & 150 & 2846 & 2 & 20 & 5 & 10 & 0 & 20 & 0 & 40 & 0 \\
GunPointAgeSpan & 135 & 316 & 2 & 21 & 3 & 10 & 0 & 19 & 2 & 39 & 2 \\
GunPointMaleVersusFemale & 135 & 316 & 2 & 21 & 3 & 9 & 2 & 19 & 2 & 39 & 2 \\
GunPointOldVersusYoung & 136 & 315 & 2 & 21 & 2 & 8 & 4 & 18 & 4 & 38 & 4 \\
Ham & 109 & 105 & 2 & 22 & 0 & 10 & 0 & 20 & 0 & 40 & 0 \\
HandOutlines & 1000 & 370 & 2 & 19 & 0 & 10 & 0 & 20 & 0 & 39 & 0 \\
Haptics & 155 & 308 & 5 & 22 & 0 & 10 & 0 & 20 & 0 & 40 & 0 \\
InlineSkate & 99 & 521 & 7 & 20 & 3 & 10 & 0 & 20 & 0 & 40 & 0 \\
InsectWingbeatSound & 220 & 1980 & 11 & 22 & 0 & 10 & 0 & 20 & 0 & 40 & 0 \\
LargeKitchenAppliances & 375 & 375 & 3 & 22 & 0 & 10 & 0 & 20 & 0 & 40 & 0 \\
MixedShapesRegularTrain & 500 & 2420 & 5 & 21 & 0 & 10 & 0 & 20 & 0 & 40 & 0 \\
MixedShapesSmallTrain & 100 & 2420 & 5 & 21 & 0 & 10 & 0 & 20 & 0 & 40 & 0 \\
NonInvasiveFetalECGThorax1 & 1800 & 1965 & 42 & 22 & 0 & 10 & 0 & 20 & 0 & 40 & 0 \\
NonInvasiveFetalECGThorax2 & 1800 & 1965 & 42 & 22 & 0 & 10 & 0 & 20 & 0 & 40 & 0 \\
OSULeaf & 200 & 242 & 6 & 22 & 0 & 10 & 0 & 20 & 0 & 40 & 0 \\
Plane & 105 & 105 & 7 & 22 & 0 & 10 & 0 & 20 & 0 & 40 & 0 \\
PowerCons & 145 & 145 & 2 & 22 & 0 & 9 & 2 & 19 & 2 & 39 & 2 \\
RefrigerationDevices & 375 & 375 & 3 & 22 & 0 & 10 & 0 & 20 & 0 & 39 & 2 \\
ScreenType & 375 & 375 & 3 & 21 & 0 & 10 & 0 & 20 & 0 & 40 & 0 \\
SemgHandGenderCh2 & 300 & 600 & 2 & 22 & 0 & 10 & 0 & 20 & 0 & 40 & 0 \\
SemgHandMovementCh2 & 450 & 450 & 6 & 22 & 0 & 10 & 0 & 20 & 0 & 40 & 0 \\
SemgHandSubjectCh2 & 450 & 450 & 5 & 22 & 0 & 10 & 0 & 20 & 0 & 40 & 0 \\
ShapesAll & 600 & 600 & 60 & 22 & 0 & 10 & 0 & 19 & 2 & 39 & 2 \\
SmallKitchenAppliances & 375 & 375 & 3 & 22 & 0 & 10 & 0 & 20 & 0 & 40 & 0 \\
StarLightCurves & 1000 & 8236 & 3 & 21 & 0 & 10 & 0 & 20 & 0 & 40 & 0 \\
Strawberry & 613 & 370 & 2 & 19 & 6 & 10 & 0 & 20 & 0 & 40 & 0 \\
SwedishLeaf & 500 & 624 & 15 & 22 & 0 & 10 & 0 & 20 & 0 & 40 & 0 \\
Trace & 100 & 100 & 4 & 20 & 3 & 7 & 2 & 15 & 2 & 34 & 4 \\
TwoPatterns & 1000 & 4000 & 4 & 22 & 0 & 10 & 0 & 20 & 0 & 40 & 0 \\
UWaveGestureLibraryAll & 896 & 3582 & 8 & 22 & 0 & 10 & 0 & 20 & 0 & 40 & 0 \\
UWaveGestureLibraryX & 896 & 3582 & 8 & 22 & 0 & 10 & 0 & 20 & 0 & 40 & 0 \\
UWaveGestureLibraryY & 896 & 3582 & 8 & 22 & 0 & 10 & 0 & 20 & 0 & 40 & 0 \\
UWaveGestureLibraryZ & 896 & 3582 & 8 & 18 & 0 & 10 & 0 & 20 & 0 & 40 & 0 \\
Wafer & 1000 & 6164 & 2 & 22 & 0 & 10 & 0 & 20 & 0 & 40 & 0 \\
Worms & 158 & 77 & 5 & 22 & 0 & 10 & 0 & 20 & 0 & 40 & 0 \\
WormsTwoClass & 158 & 77 & 2 & 22 & 0 & 9 & 0 & 19 & 0 & 39 & 0 \\
Yoga & 300 & 3000 & 2 & 21 & 3 & 9 & 2 & 19 & 2 & 39 & 2 \\
\bottomrule
\end{tabular}
\end{adjustbox}
}
\end{table}

\section{Detailed Results for Catch22}\label{appendix-catch22}
\begin{figure}[!htbp]
\centering
\includegraphics[width=0.6\columnwidth]{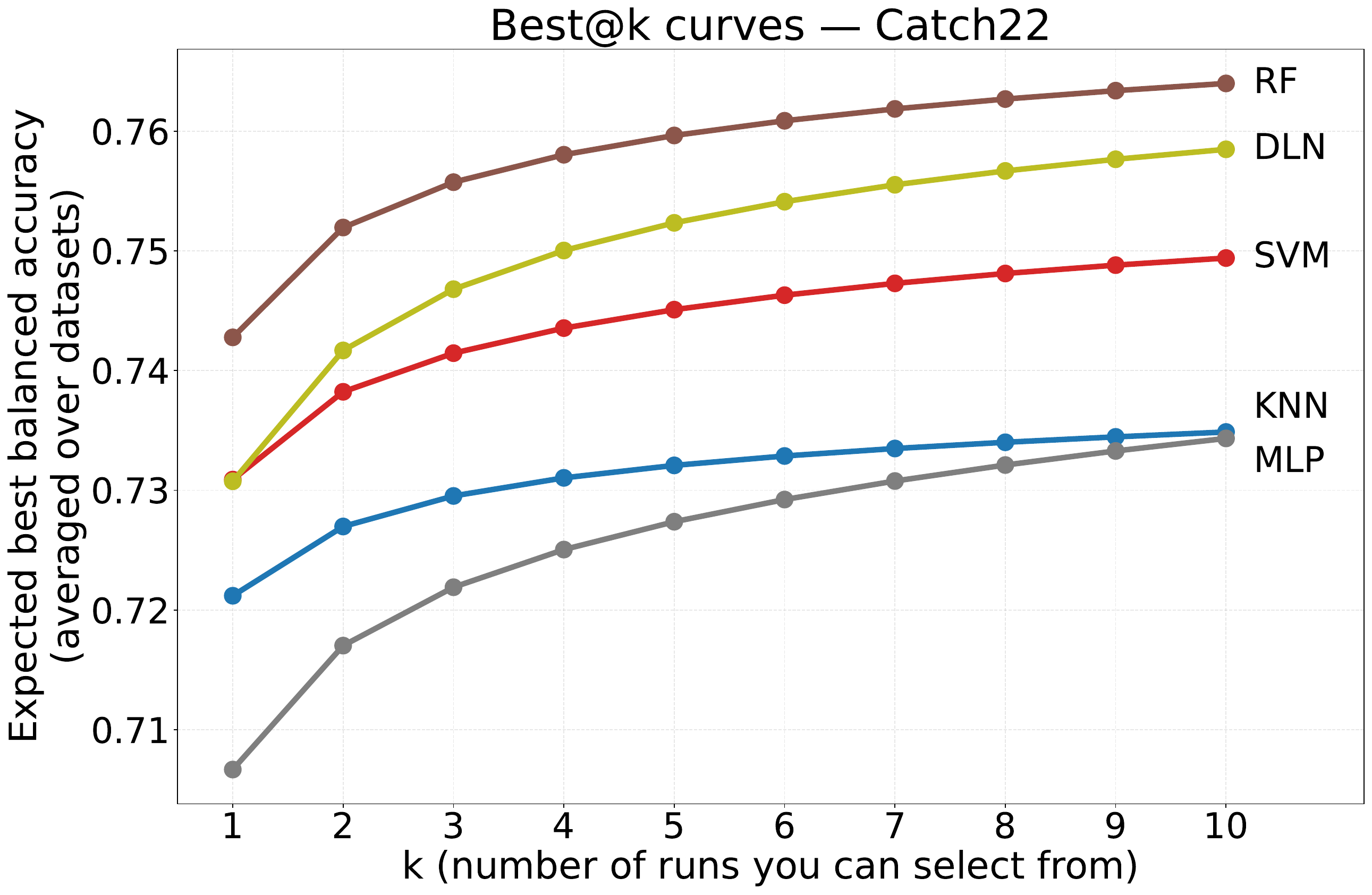}
\caption{The \textit{Best@k} curves for the five strongest models on Catch22. DLN’s curve grows the fastest.}
\label{best-k-catch22}
\Description{Line plot of Best@k performance for the five strongest models using 
Catch22 features. The DLN curve increases more rapidly than the other curves as $k$ 
grows, indicating that DLN reaches higher best balanced accuracy with fewer independent HPO-and-training runs.}
\end{figure}

In this section, we present the accuracy and inference-cost results for each dataset 
obtained with Catch22. Fig.~\ref{best-k-catch22} depicts the \textit{Best@k} curves for 
the five strongest models, acting as the Catch22 counterpart to 
Fig.~\ref{best-k-tsfresh40}. Table~\ref{results-best-acc-catch22} reports the best-of-10 
test balanced accuracy and serves as a detailed reference for Table~\ref{results-best}. 
Table~\ref{results-mean-acc-catch22} provides the dataset-wise mean test balanced accuracy 
averaged over 10 runs, corresponding to Table~\ref{results-mean}. 
Table~\ref{results-ops-catch22} lists the geometric mean number of OPs required during 
inference. A concise summary of these figures appears in Table~\ref{results-mean}.

\section{Detailed Results for TSFresh-10}\label{appendix-tsfresh10}
\begin{figure}[!htbp]
\centering
\includegraphics[width=0.6\columnwidth]{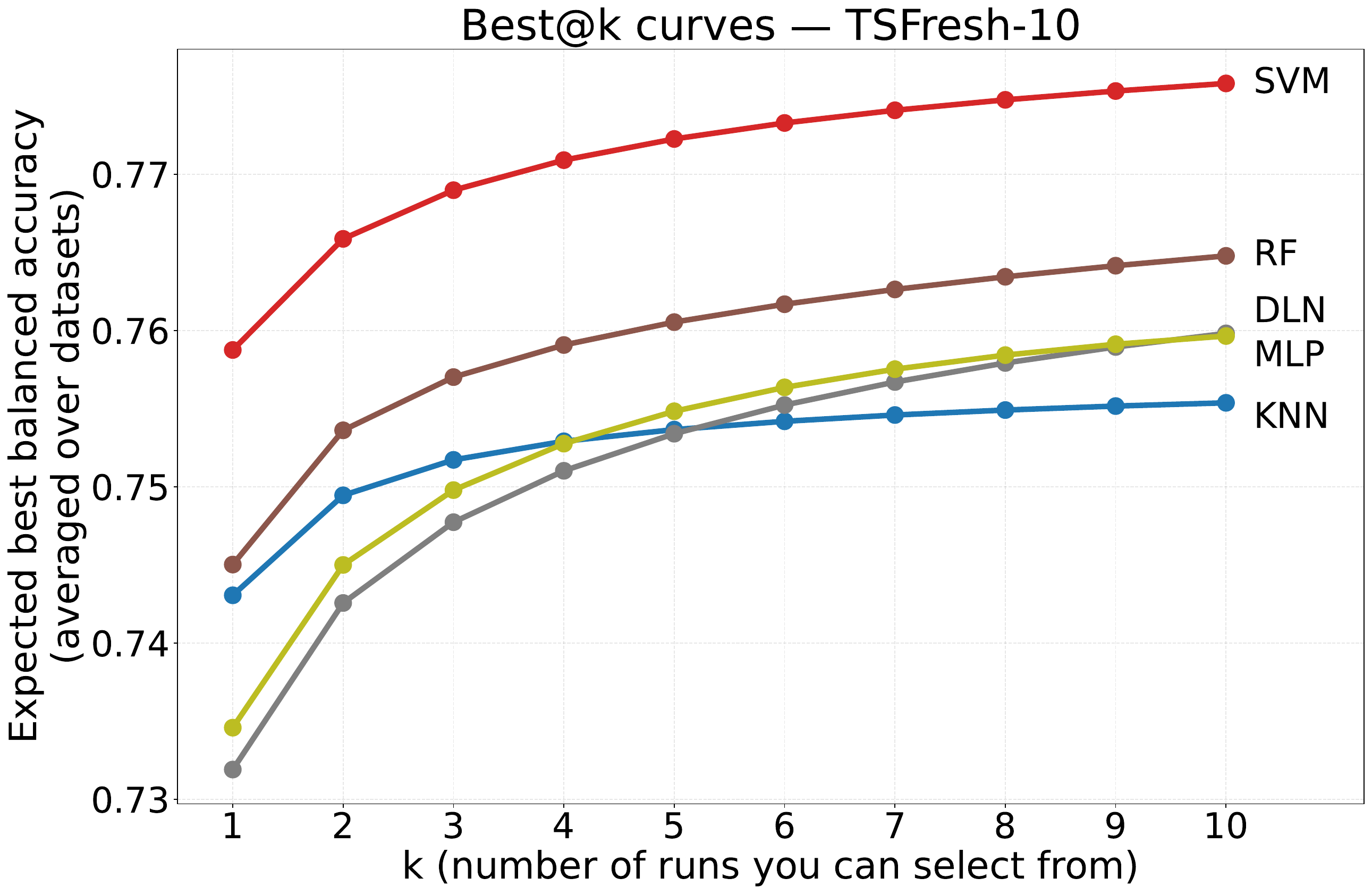}
\caption{The \textit{Best@k} curves for the five strongest models on TSFresh-10.}
\Description{Line plot showing Best@k balanced accuracy for the five strongest models 
with TSFresh-10 features. The x-axis is the number of independent HPO-and-training runs $k$ and the y-axis is the 
best balanced accuracy observed up to that run.}
\label{best-k-tsfresh10}
\end{figure}

This section reports the accuracy and inference-cost metrics obtained with the TSFresh-10 
feature transformation. Fig.~\ref{best-k-tsfresh10} illustrates the \textit{Best@k} curves 
for the five strongest models, serving as the TSFresh-10 counterpart to 
Fig.~\ref{best-k-tsfresh40}. Table~\ref{results-best-acc-tsfresh10} lists the best-of-10 
test balanced accuracies and serves as a detailed counterpart to Table~\ref{results-best}. 
Table~\ref{results-mean-acc-tsfresh10} shows the mean test balanced accuracy across 10 
runs for every dataset, corresponding to Table~\ref{results-mean}. Finally, 
Table~\ref{results-ops-tsfresh10} provides the geometric mean number of OPs required during 
inference; a concise summary of these values is included in Table~\ref{results-mean}.

\section{Detailed Results for TSFresh-20}\label{appendix-tsfresh20}
\begin{figure}[!htbp]
\centering
\includegraphics[width=0.6\columnwidth]{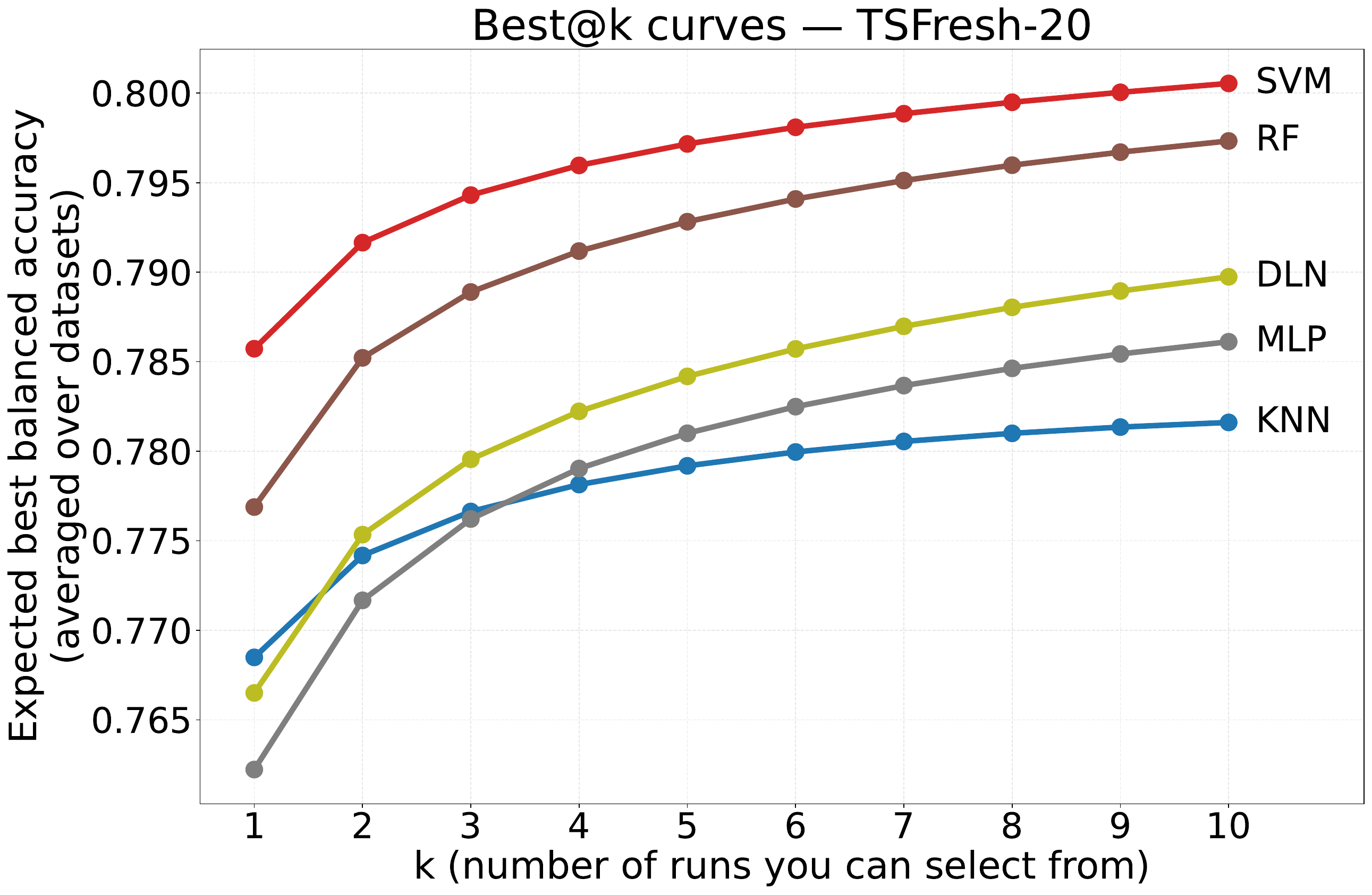}
\caption{The \textit{Best@k} curves for the five strongest models on TSFresh-20.}
\label{best-k-tsfresh20}
\Description{Line plot of Best@k results for the five strongest models using TSFresh-20 
features. Curves show, for each model, how the best balanced accuracy achieved grows as 
more independent HPO-and-training runs $k$ are performed, enabling comparison of convergence speed and final accuracy.}
\end{figure}

Analogous results for the TSFresh-20 feature transformation are presented next. 
Fig.~\ref{best-k-tsfresh20} shows the \textit{Best@k} curves for the five strongest models, 
constituting the TSFresh-20 counterpart to Fig.~\ref{best-k-tsfresh40}. 
Table~\ref{results-best-acc-tsfresh20} lists the best-of-10 test balanced accuracies, 
complementing Table~\ref{results-best}. Dataset-wise mean test balanced accuracies 
averaged over 10 runs appear in Table~\ref{results-mean-acc-tsfresh20}, providing the 
detailed values summarized in Table~\ref{results-mean}. Table~\ref{results-ops-tsfresh20} 
presents the geometric mean OP count during inference; its condensed summary is found in 
Table~\ref{results-mean}.

\section{Detailed Results for TSFresh-40}\label{appendix-tsfresh40}
Finally, we report the metrics for the TSFresh-40 feature transformation. 
Table~\ref{results-best-acc-tsfresh40} details the best-of-10 test balanced accuracies, 
corresponding to Table~\ref{results-best}. Table~\ref{results-mean-acc-tsfresh40} lists 
the mean test balanced accuracies across 10 runs for each dataset, providing the 
dataset-level figures summarized in Table~\ref{results-mean}. 
Table~\ref{results-ops-tsfresh40} reports the geometric mean number of OPs required at 
inference time; a summary of these results is included in Table~\ref{results-mean}.

\clearpage
\begin{table}[t]
\centering
\caption{Best-of-10 test balanced accuracy for Catch22.}
\label{results-best-acc-catch22}
\footnotesize{
\begin{adjustbox}{max width=\linewidth}
\begin{tabularx}{\linewidth}{l*{9}{Y}}
\toprule
& KNN & NB & LR & SVM & DT & RF & AB & MLP & DLN \\ 
\midrule
ACSF1 & 0.790 & 0.710 & 0.740 & 0.760 & 0.690 & 0.850 & 0.580 & 0.750 & 0.770 \\
\addlinespace[0pt]
ChlorineConcentration & 0.499 & 0.394 & 0.388 & 0.496 & 0.423 & 0.457 & 0.402 & 0.477 & 0.465 \\
\addlinespace[0pt]
Computers & 0.712 & 0.684 & 0.720 & 0.752 & 0.708 & 0.732 & 0.740 & 0.728 & 0.736 \\
\addlinespace[0pt]
CricketX & 0.597 & 0.451 & 0.501 & 0.579 & 0.457 & 0.573 & 0.413 & 0.527 & 0.574 \\
\addlinespace[0pt]
CricketY & 0.550 & 0.411 & 0.481 & 0.551 & 0.383 & 0.533 & 0.397 & 0.498 & 0.538 \\
\addlinespace[0pt]
CricketZ & 0.599 & 0.443 & 0.520 & 0.573 & 0.432 & 0.616 & 0.371 & 0.541 & 0.566 \\
\addlinespace[0pt]
Earthquakes & 0.580 & 0.684 & 0.693 & 0.674 & 0.641 & 0.703 & 0.552 & 0.609 & 0.689 \\
\addlinespace[0pt]
EOGHorizontalSignal & 0.495 & 0.445 & 0.521 & 0.491 & 0.417 & 0.580 & 0.376 & 0.501 & 0.525 \\
\addlinespace[0pt]
EOGVerticalSignal & 0.454 & 0.373 & 0.421 & 0.432 & 0.403 & 0.499 & 0.353 & 0.414 & 0.456 \\
\addlinespace[0pt]
EthanolLevel & 0.328 & 0.312 & 0.374 & 0.388 & 0.339 & 0.352 & 0.368 & 0.400 & 0.396 \\
\addlinespace[0pt]
FaceAll & 0.643 & 0.570 & 0.689 & 0.684 & 0.569 & 0.741 & 0.501 & 0.668 & 0.730 \\
\addlinespace[0pt]
Fish & 0.769 & 0.614 & 0.793 & 0.775 & 0.640 & 0.769 & 0.623 & 0.777 & 0.796 \\
\addlinespace[0pt]
FordA & 0.875 & 0.699 & 0.860 & 0.901 & 0.909 & 0.917 & 0.913 & 0.913 & 0.917 \\
\addlinespace[0pt]
FordB & 0.738 & 0.586 & 0.665 & 0.751 & 0.731 & 0.730 & 0.745 & 0.762 & 0.774 \\
\addlinespace[0pt]
FreezerRegularTrain & 0.961 & 0.867 & 0.996 & 0.996 & 0.997 & 0.998 & 0.997 & 0.995 & 0.998 \\
\addlinespace[0pt]
GunPointAgeSpan & 0.936 & 0.805 & 0.915 & 0.953 & 0.946 & 0.978 & 0.965 & 0.940 & 0.959 \\
\addlinespace[0pt]
GunPointMaleVersusFemale & 1.00 & 0.984 & 0.997 & 0.997 & 0.978 & 0.991 & 0.978 & 0.997 & 1.00 \\
\addlinespace[0pt]
GunPointOldVersusYoung & 1.00 & 0.988 & 1.00 & 1.00 & 1.00 & 1.00 & 1.00 & 1.00 & 1.00 \\
\addlinespace[0pt]
Ham & 0.678 & 0.511 & 0.725 & 0.743 & 0.628 & 0.641 & 0.629 & 0.704 & 0.697 \\
\addlinespace[0pt]
HandOutlines & 0.797 & 0.622 & 0.816 & 0.845 & 0.809 & 0.851 & 0.813 & 0.847 & 0.836 \\
\addlinespace[0pt]
Haptics & 0.424 & 0.374 & 0.430 & 0.430 & 0.397 & 0.502 & 0.422 & 0.394 & 0.456 \\
\addlinespace[0pt]
InlineSkate & 0.420 & 0.341 & 0.376 & 0.388 & 0.369 & 0.473 & 0.322 & 0.392 & 0.382 \\
\addlinespace[0pt]
InsectWingbeatSound & 0.461 & 0.507 & 0.466 & 0.483 & 0.465 & 0.573 & 0.427 & 0.444 & 0.502 \\
\addlinespace[0pt]
LargeKitchenAppliances & 0.723 & 0.800 & 0.864 & 0.869 & 0.805 & 0.853 & 0.797 & 0.859 & 0.827 \\
\addlinespace[0pt]
MixedShapesRegularTrain & 0.901 & 0.839 & 0.884 & 0.895 & 0.858 & 0.912 & 0.868 & 0.886 & 0.906 \\
\addlinespace[0pt]
MixedShapesSmallTrain & 0.847 & 0.832 & 0.847 & 0.844 & 0.829 & 0.862 & 0.842 & 0.847 & 0.860 \\
\addlinespace[0pt]
NonInvasiveFetalECGThorax1 & 0.765 & 0.766 & 0.816 & 0.849 & 0.712 & 0.822 & 0.208 & 0.801 & 0.820 \\
\addlinespace[0pt]
NonInvasiveFetalECGThorax2 & 0.830 & 0.798 & 0.849 & 0.857 & 0.744 & 0.852 & 0.212 & 0.840 & 0.857 \\
\addlinespace[0pt]
OSULeaf & 0.661 & 0.618 & 0.700 & 0.734 & 0.575 & 0.684 & 0.581 & 0.712 & 0.762 \\
\addlinespace[0pt]
Plane & 0.991 & 0.912 & 0.938 & 0.959 & 0.944 & 1.00 & 0.949 & 0.954 & 0.991 \\
\addlinespace[0pt]
PowerCons & 0.917 & 0.897 & 0.924 & 0.931 & 0.883 & 0.931 & 0.952 & 0.931 & 0.945 \\
\addlinespace[0pt]
RefrigerationDevices & 0.539 & 0.525 & 0.533 & 0.587 & 0.536 & 0.536 & 0.504 & 0.597 & 0.536 \\
\addlinespace[0pt]
ScreenType & 0.456 & 0.389 & 0.477 & 0.493 & 0.507 & 0.523 & 0.531 & 0.488 & 0.515 \\
\addlinespace[0pt]
SemgHandGenderCh2 & 0.915 & 0.742 & 0.822 & 0.899 & 0.876 & 0.903 & 0.872 & 0.904 & 0.917 \\
\addlinespace[0pt]
SemgHandMovementCh2 & 0.718 & 0.427 & 0.556 & 0.682 & 0.649 & 0.778 & 0.578 & 0.698 & 0.760 \\
\addlinespace[0pt]
SemgHandSubjectCh2 & 0.842 & 0.618 & 0.722 & 0.847 & 0.773 & 0.836 & 0.693 & 0.829 & 0.858 \\
\addlinespace[0pt]
ShapesAll & 0.718 & 0.688 & 0.632 & 0.677 & 0.575 & 0.727 & 0.123 & 0.600 & 0.717 \\
\addlinespace[0pt]
SmallKitchenAppliances & 0.755 & 0.741 & 0.811 & 0.795 & 0.763 & 0.819 & 0.797 & 0.803 & 0.816 \\
\addlinespace[0pt]
StarLightCurves & 0.931 & 0.885 & 0.937 & 0.943 & 0.946 & 0.947 & 0.943 & 0.939 & 0.946 \\
\addlinespace[0pt]
Strawberry & 0.913 & 0.705 & 0.865 & 0.929 & 0.891 & 0.926 & 0.880 & 0.918 & 0.917 \\
\addlinespace[0pt]
SwedishLeaf & 0.850 & 0.809 & 0.849 & 0.857 & 0.802 & 0.881 & 0.657 & 0.836 & 0.884 \\
\addlinespace[0pt]
Trace & 1.00 & 0.913 & 1.00 & 1.00 & 1.00 & 1.00 & 0.952 & 1.00 & 1.00 \\
\addlinespace[0pt]
TwoPatterns & 0.768 & 0.787 & 0.808 & 0.847 & 0.840 & 0.844 & 0.755 & 0.838 & 0.865 \\
\addlinespace[0pt]
UWaveGestureLibraryAll & 0.745 & 0.667 & 0.722 & 0.772 & 0.694 & 0.809 & 0.697 & 0.731 & 0.800 \\
\addlinespace[0pt]
UWaveGestureLibraryX & 0.712 & 0.664 & 0.692 & 0.732 & 0.674 & 0.747 & 0.628 & 0.698 & 0.736 \\
\addlinespace[0pt]
UWaveGestureLibraryY & 0.654 & 0.541 & 0.606 & 0.658 & 0.601 & 0.691 & 0.540 & 0.613 & 0.674 \\
\addlinespace[0pt]
UWaveGestureLibraryZ & 0.668 & 0.617 & 0.644 & 0.680 & 0.613 & 0.682 & 0.608 & 0.667 & 0.681 \\
\addlinespace[0pt]
Wafer & 0.962 & 0.969 & 0.979 & 0.980 & 0.983 & 0.995 & 0.992 & 0.981 & 0.992 \\
\addlinespace[0pt]
Worms & 0.765 & 0.635 & 0.675 & 0.703 & 0.574 & 0.705 & 0.536 & 0.669 & 0.709 \\
\addlinespace[0pt]
WormsTwoClass & 0.886 & 0.818 & 0.784 & 0.818 & 0.761 & 0.879 & 0.773 & 0.822 & 0.879 \\
\addlinespace[0pt]
Yoga & 0.740 & 0.582 & 0.678 & 0.739 & 0.737 & 0.762 & 0.716 & 0.712 & 0.753 \\
\bottomrule
\end{tabularx}
\end{adjustbox}
}
\end{table}

\clearpage
\begingroup
\setlength{\tabcolsep}{3pt}
\footnotesize{
\begin{longtable}[c]{@{}Z{0.22\textwidth}*{9}{c}@{}}
\caption{Mean and standard deviation of balanced accuracy for Catch22 across 10 random seeds.\label{results-mean-acc-catch22}}\\
\toprule
& KNN & NB & LR & SVM & DT & RF & AB & MLP & DLN \\ \midrule
\endfirsthead
\multicolumn{10}{c}{\tablename\ \thetable\ (continued)}\\
\toprule
& KNN & NB & LR & SVM & DT & RF & AB & MLP & DLN \\ \midrule
\endhead
\midrule
\multicolumn{10}{r}{continued on next page}\\
\endfoot
\bottomrule
\endlastfoot
ACSF1 & \meanstd{0.758}{0.030} & \meanstd{0.698}{0.018} & \meanstd{0.687}{0.052} & \meanstd{0.697}{0.028} & \meanstd{0.655}{0.028} & \meanstd{0.807}{0.033} & \meanstd{0.441}{0.081} & \meanstd{0.715}{0.028} & \meanstd{0.728}{0.026} \\
\addlinespace[0pt]
ChlorineConcentration & \meanstd{0.488}{0.010} & \meanstd{0.394}{0.0} & \meanstd{0.373}{8.5e-03} & \meanstd{0.475}{0.031} & \meanstd{0.411}{5.9e-03} & \meanstd{0.435}{0.013} & \meanstd{0.396}{5.0e-03} & \meanstd{0.464}{0.011} & \meanstd{0.440}{0.018} \\
\addlinespace[0pt]
Computers & \meanstd{0.683}{0.024} & \meanstd{0.684}{0.0} & \meanstd{0.704}{0.015} & \meanstd{0.723}{0.020} & \meanstd{0.682}{0.029} & \meanstd{0.711}{0.011} & \meanstd{0.710}{0.017} & \meanstd{0.682}{0.028} & \meanstd{0.690}{0.024} \\
\addlinespace[0pt]
CricketX & \meanstd{0.585}{0.011} & \meanstd{0.451}{0.0} & \meanstd{0.477}{0.017} & \meanstd{0.561}{0.016} & \meanstd{0.427}{0.034} & \meanstd{0.550}{0.017} & \meanstd{0.384}{0.017} & \meanstd{0.507}{0.023} & \meanstd{0.554}{0.017} \\
\addlinespace[0pt]
CricketY & \meanstd{0.537}{7.2e-03} & \meanstd{0.411}{0.0} & \meanstd{0.472}{6.1e-03} & \meanstd{0.534}{0.011} & \meanstd{0.362}{0.013} & \meanstd{0.515}{0.012} & \meanstd{0.360}{0.019} & \meanstd{0.476}{0.018} & \meanstd{0.509}{0.026} \\
\addlinespace[0pt]
CricketZ & \meanstd{0.591}{7.0e-03} & \meanstd{0.443}{0.0} & \meanstd{0.509}{6.4e-03} & \meanstd{0.540}{0.016} & \meanstd{0.417}{0.011} & \meanstd{0.590}{0.015} & \meanstd{0.344}{0.013} & \meanstd{0.492}{0.031} & \meanstd{0.530}{0.022} \\
\addlinespace[0pt]
EOGHorizontalSignal & \meanstd{0.470}{0.013} & \meanstd{0.444}{3.5e-03} & \meanstd{0.488}{0.036} & \meanstd{0.449}{0.029} & \meanstd{0.383}{0.029} & \meanstd{0.556}{0.013} & \meanstd{0.287}{0.040} & \meanstd{0.460}{0.034} & \meanstd{0.489}{0.027} \\
\addlinespace[0pt]
EOGVerticalSignal & \meanstd{0.414}{0.014} & \meanstd{0.369}{5.4e-03} & \meanstd{0.357}{0.027} & \meanstd{0.394}{0.025} & \meanstd{0.370}{0.017} & \meanstd{0.480}{0.012} & \meanstd{0.290}{0.040} & \meanstd{0.376}{0.020} & \meanstd{0.415}{0.027} \\
\addlinespace[0pt]
Earthquakes & \meanstd{0.573}{0.012} & \meanstd{0.684}{0.0} & \meanstd{0.678}{0.021} & \meanstd{0.638}{0.029} & \meanstd{0.602}{0.022} & \meanstd{0.674}{0.016} & \meanstd{0.537}{0.010} & \meanstd{0.559}{0.034} & \meanstd{0.600}{0.052} \\
\addlinespace[0pt]
EthanolLevel & \meanstd{0.313}{0.013} & \meanstd{0.312}{0.0} & \meanstd{0.366}{4.3e-03} & \meanstd{0.370}{0.011} & \meanstd{0.328}{0.012} & \meanstd{0.338}{8.8e-03} & \meanstd{0.350}{0.010} & \meanstd{0.359}{0.019} & \meanstd{0.371}{0.014} \\
\addlinespace[0pt]
FaceAll & \meanstd{0.631}{9.4e-03} & \meanstd{0.570}{0.0} & \meanstd{0.684}{4.4e-03} & \meanstd{0.667}{0.015} & \meanstd{0.526}{0.025} & \meanstd{0.716}{0.016} & \meanstd{0.453}{0.033} & \meanstd{0.642}{0.012} & \meanstd{0.703}{0.023} \\
\addlinespace[0pt]
Fish & \meanstd{0.721}{0.032} & \meanstd{0.614}{0.0} & \meanstd{0.750}{0.039} & \meanstd{0.749}{0.015} & \meanstd{0.600}{0.023} & \meanstd{0.721}{0.037} & \meanstd{0.604}{0.017} & \meanstd{0.751}{0.026} & \meanstd{0.748}{0.027} \\
\addlinespace[0pt]
FordA & \meanstd{0.869}{2.9e-03} & \meanstd{0.699}{0.0} & \meanstd{0.857}{1.4e-03} & \meanstd{0.897}{3.3e-03} & \meanstd{0.905}{3.2e-03} & \meanstd{0.905}{9.3e-03} & \meanstd{0.908}{3.0e-03} & \meanstd{0.894}{9.7e-03} & \meanstd{0.909}{5.1e-03} \\
\addlinespace[0pt]
FordB & \meanstd{0.714}{0.014} & \meanstd{0.586}{0.0} & \meanstd{0.661}{2.5e-03} & \meanstd{0.745}{4.7e-03} & \meanstd{0.714}{0.024} & \meanstd{0.703}{0.029} & \meanstd{0.733}{9.7e-03} & \meanstd{0.729}{0.017} & \meanstd{0.759}{9.3e-03} \\
\addlinespace[0pt]
FreezerRegularTrain & \meanstd{0.957}{7.9e-03} & \meanstd{0.866}{9.1e-04} & \meanstd{0.996}{2.8e-04} & \meanstd{0.996}{4.7e-04} & \meanstd{0.997}{0.0} & \meanstd{0.997}{3.4e-04} & \meanstd{0.997}{0.0} & \meanstd{0.987}{6.0e-03} & \meanstd{0.988}{0.016} \\
\addlinespace[0pt]
GunPointAgeSpan & \meanstd{0.934}{1.1e-03} & \meanstd{0.800}{2.2e-03} & \meanstd{0.892}{0.018} & \meanstd{0.940}{0.019} & \meanstd{0.922}{0.026} & \meanstd{0.971}{4.8e-03} & \meanstd{0.960}{3.6e-03} & \meanstd{0.930}{9.5e-03} & \meanstd{0.942}{0.018} \\
\addlinespace[0pt]
GunPointMaleVsFemale & \meanstd{0.999}{1.3e-03} & \meanstd{0.981}{2.5e-03} & \meanstd{0.995}{2.9e-03} & \meanstd{0.994}{2.5e-03} & \meanstd{0.976}{3.8e-03} & \meanstd{0.984}{7.5e-03} & \meanstd{0.975}{2.9e-03} & \meanstd{0.992}{6.4e-03} & \meanstd{0.995}{4.0e-03} \\
\addlinespace[0pt]
GunPointOldVsYoung & \meanstd{1.00}{9.6e-04} & \meanstd{0.988}{0.0} & \meanstd{1.00}{9.6e-04} & \meanstd{1.00}{0.0} & \meanstd{1.00}{0.0} & \meanstd{1.00}{0.0} & \meanstd{1.00}{0.0} & \meanstd{0.999}{3.1e-03} & \meanstd{1.00}{0.0} \\
\addlinespace[0pt]
Ham & \meanstd{0.579}{0.044} & \meanstd{0.511}{0.0} & \meanstd{0.695}{0.017} & \meanstd{0.689}{0.053} & \meanstd{0.601}{0.021} & \meanstd{0.615}{0.019} & \meanstd{0.604}{0.020} & \meanstd{0.662}{0.029} & \meanstd{0.634}{0.047} \\
\addlinespace[0pt]
HandOutlines & \meanstd{0.787}{0.015} & \meanstd{0.622}{0.0} & \meanstd{0.814}{5.0e-03} & \meanstd{0.823}{0.013} & \meanstd{0.786}{0.013} & \meanstd{0.841}{7.3e-03} & \meanstd{0.806}{6.0e-03} & \meanstd{0.796}{0.020} & \meanstd{0.820}{0.014} \\
\addlinespace[0pt]
Haptics & \meanstd{0.410}{0.010} & \meanstd{0.374}{0.0} & \meanstd{0.399}{0.013} & \meanstd{0.385}{0.031} & \meanstd{0.356}{0.024} & \meanstd{0.480}{0.017} & \meanstd{0.406}{0.011} & \meanstd{0.383}{0.013} & \meanstd{0.430}{0.022} \\
\addlinespace[0pt]
InlineSkate & \meanstd{0.411}{0.017} & \meanstd{0.337}{4.0e-03} & \meanstd{0.356}{0.023} & \meanstd{0.379}{5.5e-03} & \meanstd{0.339}{0.036} & \meanstd{0.436}{0.022} & \meanstd{0.293}{0.017} & \meanstd{0.364}{0.024} & \meanstd{0.353}{0.022} \\
\addlinespace[0pt]
InsectWingbeatSound & \meanstd{0.444}{0.011} & \meanstd{0.507}{0.0} & \meanstd{0.450}{0.015} & \meanstd{0.468}{0.011} & \meanstd{0.436}{0.021} & \meanstd{0.526}{0.032} & \meanstd{0.405}{0.023} & \meanstd{0.429}{0.011} & \meanstd{0.488}{0.016} \\
\addlinespace[0pt]
LargeKitchenAppliances & \meanstd{0.719}{7.9e-03} & \meanstd{0.800}{0.0} & \meanstd{0.849}{0.014} & \meanstd{0.858}{5.8e-03} & \meanstd{0.785}{0.011} & \meanstd{0.840}{0.012} & \meanstd{0.764}{0.023} & \meanstd{0.826}{0.023} & \meanstd{0.812}{0.011} \\
\addlinespace[0pt]
MixedShapesRegTrain & \meanstd{0.898}{4.2e-03} & \meanstd{0.839}{0.0} & \meanstd{0.879}{3.3e-03} & \meanstd{0.890}{4.4e-03} & \meanstd{0.849}{6.0e-03} & \meanstd{0.901}{9.7e-03} & \meanstd{0.864}{3.3e-03} & \meanstd{0.875}{8.8e-03} & \meanstd{0.897}{8.4e-03} \\
\addlinespace[0pt]
MixedShapesSmallTrain & \meanstd{0.846}{5.2e-03} & \meanstd{0.832}{0.0} & \meanstd{0.831}{0.019} & \meanstd{0.830}{9.2e-03} & \meanstd{0.810}{0.013} & \meanstd{0.851}{9.3e-03} & \meanstd{0.793}{0.072} & \meanstd{0.830}{7.9e-03} & \meanstd{0.838}{0.015} \\
\addlinespace[0pt]
NonInvasiveFetalECG1 & \meanstd{0.765}{1.8e-03} & \meanstd{0.765}{9.2e-04} & \meanstd{0.812}{3.7e-03} & \meanstd{0.842}{3.1e-03} & \meanstd{0.697}{0.012} & \meanstd{0.797}{0.025} & \meanstd{0.153}{0.028} & \meanstd{0.792}{7.1e-03} & \meanstd{0.804}{0.012} \\
\addlinespace[0pt]
NonInvasiveFetalECG2 & \meanstd{0.826}{2.6e-03} & \meanstd{0.798}{2.8e-04} & \meanstd{0.835}{0.011} & \meanstd{0.852}{3.8e-03} & \meanstd{0.728}{0.016} & \meanstd{0.823}{0.022} & \meanstd{0.157}{0.034} & \meanstd{0.826}{8.9e-03} & \meanstd{0.845}{8.5e-03} \\
\addlinespace[0pt]
OSULeaf & \meanstd{0.647}{0.012} & \meanstd{0.618}{0.0} & \meanstd{0.667}{0.033} & \meanstd{0.715}{0.010} & \meanstd{0.558}{8.1e-03} & \meanstd{0.666}{0.014} & \meanstd{0.532}{0.045} & \meanstd{0.673}{0.019} & \meanstd{0.702}{0.030} \\
\addlinespace[0pt]
Plane & \meanstd{0.988}{4.3e-03} & \meanstd{0.912}{0.0} & \meanstd{0.934}{0.011} & \meanstd{0.935}{0.023} & \meanstd{0.917}{0.027} & \meanstd{0.988}{9.6e-03} & \meanstd{0.842}{0.16} & \meanstd{0.933}{0.030} & \meanstd{0.965}{0.029} \\
\addlinespace[0pt]
PowerCons & \meanstd{0.901}{0.018} & \meanstd{0.897}{0.0} & \meanstd{0.903}{0.018} & \meanstd{0.911}{0.013} & \meanstd{0.832}{0.024} & \meanstd{0.907}{0.019} & \meanstd{0.939}{8.2e-03} & \meanstd{0.904}{0.018} & \meanstd{0.910}{0.028} \\
\addlinespace[0pt]
RefrigerationDevices & \meanstd{0.531}{2.5e-03} & \meanstd{0.525}{0.0} & \meanstd{0.504}{0.013} & \meanstd{0.519}{0.034} & \meanstd{0.485}{0.042} & \meanstd{0.521}{7.4e-03} & \meanstd{0.485}{0.015} & \meanstd{0.519}{0.035} & \meanstd{0.486}{0.027} \\
\addlinespace[0pt]
ScreenType & \meanstd{0.444}{0.010} & \meanstd{0.389}{0.0} & \meanstd{0.475}{2.2e-03} & \meanstd{0.481}{0.014} & \meanstd{0.458}{0.026} & \meanstd{0.507}{0.017} & \meanstd{0.473}{0.033} & \meanstd{0.455}{0.020} & \meanstd{0.473}{0.030} \\
\addlinespace[0pt]
SemgHandGenderCh2 & \meanstd{0.904}{9.4e-03} & \meanstd{0.742}{0.0} & \meanstd{0.811}{8.6e-03} & \meanstd{0.889}{0.014} & \meanstd{0.822}{0.034} & \meanstd{0.887}{0.012} & \meanstd{0.865}{5.4e-03} & \meanstd{0.874}{0.019} & \meanstd{0.904}{8.4e-03} \\
\addlinespace[0pt]
SemgHandMovementCh2 & \meanstd{0.704}{0.016} & \meanstd{0.427}{0.0} & \meanstd{0.543}{0.014} & \meanstd{0.664}{0.011} & \meanstd{0.624}{0.022} & \meanstd{0.748}{0.035} & \meanstd{0.536}{0.022} & \meanstd{0.658}{0.020} & \meanstd{0.708}{0.023} \\
\addlinespace[0pt]
SemgHandSubjectCh2 & \meanstd{0.835}{6.6e-03} & \meanstd{0.618}{0.0} & \meanstd{0.712}{5.2e-03} & \meanstd{0.836}{0.013} & \meanstd{0.752}{8.4e-03} & \meanstd{0.813}{0.027} & \meanstd{0.673}{0.015} & \meanstd{0.806}{0.021} & \meanstd{0.840}{0.013} \\
\addlinespace[0pt]
ShapesAll & \meanstd{0.713}{6.7e-03} & \meanstd{0.688}{7.0e-04} & \meanstd{0.618}{0.012} & \meanstd{0.654}{0.014} & \meanstd{0.537}{0.018} & \meanstd{0.685}{0.031} & \meanstd{0.101}{0.017} & \meanstd{0.583}{0.017} & \meanstd{0.692}{0.011} \\
\addlinespace[0pt]
SmallKitchenAppliances & \meanstd{0.749}{8.0e-03} & \meanstd{0.741}{0.0} & \meanstd{0.798}{7.3e-03} & \meanstd{0.781}{0.011} & \meanstd{0.757}{0.019} & \meanstd{0.809}{7.7e-03} & \meanstd{0.781}{0.011} & \meanstd{0.768}{0.022} & \meanstd{0.777}{0.024} \\
\addlinespace[0pt]
StarLightCurves & \meanstd{0.929}{7.5e-04} & \meanstd{0.885}{3.7e-05} & \meanstd{0.934}{1.2e-03} & \meanstd{0.941}{1.5e-03} & \meanstd{0.933}{0.012} & \meanstd{0.943}{5.2e-03} & \meanstd{0.932}{7.4e-03} & \meanstd{0.927}{6.7e-03} & \meanstd{0.939}{4.4e-03} \\
\addlinespace[0pt]
Strawberry & \meanstd{0.910}{5.2e-03} & \meanstd{0.705}{0.0} & \meanstd{0.862}{1.3e-03} & \meanstd{0.923}{5.1e-03} & \meanstd{0.864}{0.018} & \meanstd{0.905}{0.010} & \meanstd{0.870}{7.4e-03} & \meanstd{0.904}{0.011} & \meanstd{0.906}{0.014} \\
\addlinespace[0pt]
SwedishLeaf & \meanstd{0.823}{0.021} & \meanstd{0.807}{1.4e-03} & \meanstd{0.838}{6.0e-03} & \meanstd{0.849}{6.7e-03} & \meanstd{0.785}{0.019} & \meanstd{0.862}{0.021} & \meanstd{0.559}{0.064} & \meanstd{0.813}{0.016} & \meanstd{0.857}{0.019} \\
\addlinespace[0pt]
Trace & \meanstd{1.00}{0.0} & \meanstd{0.913}{0.0} & \meanstd{0.999}{4.2e-03} & \meanstd{1.00}{0.0} & \meanstd{0.997}{9.8e-03} & \meanstd{0.967}{0.083} & \meanstd{0.812}{0.12} & \meanstd{0.996}{6.4e-03} & \meanstd{0.989}{0.018} \\
\addlinespace[0pt]
TwoPatterns & \meanstd{0.765}{3.1e-03} & \meanstd{0.787}{0.0} & \meanstd{0.805}{2.0e-03} & \meanstd{0.841}{7.1e-03} & \meanstd{0.831}{9.3e-03} & \meanstd{0.833}{4.9e-03} & \meanstd{0.746}{4.6e-03} & \meanstd{0.828}{7.7e-03} & \meanstd{0.852}{5.6e-03} \\
\addlinespace[0pt]
UWaveGestureLibraryAll & \meanstd{0.740}{6.3e-03} & \meanstd{0.667}{0.0} & \meanstd{0.716}{2.4e-03} & \meanstd{0.768}{2.0e-03} & \meanstd{0.679}{0.012} & \meanstd{0.793}{0.014} & \meanstd{0.681}{9.9e-03} & \meanstd{0.716}{0.012} & \meanstd{0.791}{7.2e-03} \\
\addlinespace[0pt]
UWaveGestureLibraryX & \meanstd{0.705}{9.4e-03} & \meanstd{0.662}{2.0e-03} & \meanstd{0.689}{3.0e-03} & \meanstd{0.718}{0.010} & \meanstd{0.668}{6.9e-03} & \meanstd{0.735}{7.5e-03} & \meanstd{0.600}{0.019} & \meanstd{0.678}{0.015} & \meanstd{0.728}{4.9e-03} \\
\addlinespace[0pt]
UWaveGestureLibraryY & \meanstd{0.647}{4.3e-03} & \meanstd{0.538}{2.7e-03} & \meanstd{0.602}{1.9e-03} & \meanstd{0.651}{7.0e-03} & \meanstd{0.577}{0.015} & \meanstd{0.682}{7.9e-03} & \meanstd{0.519}{0.017} & \meanstd{0.600}{0.011} & \meanstd{0.659}{0.010} \\
\addlinespace[0pt]
UWaveGestureLibraryZ & \meanstd{0.660}{9.6e-03} & \meanstd{0.617}{0.0} & \meanstd{0.643}{1.3e-03} & \meanstd{0.669}{0.011} & \meanstd{0.604}{4.8e-03} & \meanstd{0.667}{0.012} & \meanstd{0.592}{0.011} & \meanstd{0.645}{0.015} & \meanstd{0.674}{5.7e-03} \\
\addlinespace[0pt]
Wafer & \meanstd{0.961}{8.3e-04} & \meanstd{0.969}{0.0} & \meanstd{0.974}{3.3e-03} & \meanstd{0.978}{1.5e-03} & \meanstd{0.970}{9.6e-03} & \meanstd{0.986}{6.3e-03} & \meanstd{0.988}{6.6e-03} & \meanstd{0.974}{4.1e-03} & \meanstd{0.987}{5.4e-03} \\
\addlinespace[0pt]
Worms & \meanstd{0.712}{0.044} & \meanstd{0.635}{0.0} & \meanstd{0.641}{0.037} & \meanstd{0.675}{0.018} & \meanstd{0.532}{0.039} & \meanstd{0.639}{0.033} & \meanstd{0.471}{0.056} & \meanstd{0.592}{0.047} & \meanstd{0.627}{0.055} \\
\addlinespace[0pt]
WormsTwoClass & \meanstd{0.855}{0.023} & \meanstd{0.818}{0.0} & \meanstd{0.764}{7.2e-03} & \meanstd{0.766}{0.036} & \meanstd{0.740}{0.019} & \meanstd{0.841}{0.023} & \meanstd{0.722}{0.032} & \meanstd{0.703}{0.053} & \meanstd{0.791}{0.046} \\
\addlinespace[0pt]
Yoga & \meanstd{0.739}{3.3e-04} & \meanstd{0.582}{0.0} & \meanstd{0.673}{3.7e-03} & \meanstd{0.725}{0.013} & \meanstd{0.699}{0.018} & \meanstd{0.732}{0.015} & \meanstd{0.705}{8.7e-03} & \meanstd{0.694}{0.016} & \meanstd{0.721}{0.023} \\
\end{longtable}
}
\endgroup

\clearpage
\begin{table}[t]
\centering
\caption{Geometric mean of inference OPs for Catch22 across 10 random seeds assuming FP16 for floating-point and INT16 for integer arithmetic.}
\label{results-ops-catch22}
\footnotesize{
\begin{tabularx}{\linewidth}{l*{9}{Y}}
\toprule
& KNN & NB & LR & SVM & DT & RF & AB & MLP & DLN \\ 
\midrule
ACSF1 & 1.15M & 2.52M & 1.81M & 8.20M & 915 & 115K & 190K & 8.13M & 31.9K \\
\addlinespace[0pt]
ChlorineConcentration & 7.85M & 679K & 518K & 29.8M & 1.65K & 176K & 222K & 7.53M & 38.6K \\
\addlinespace[0pt]
Computers & 1.64M & 445K & 171K & 10.7M & 695 & 137K & 205K & 2.52M & 26.5K \\
\addlinespace[0pt]
CricketX & 2.69M & 2.71M & 2.07M & 63.1M & 1.23K & 186K & 252K & 4.66M & 85.2K \\
\addlinespace[0pt]
CricketY & 2.69M & 2.71M & 2.07M & 49.1M & 1.34K & 193K & 239K & 5.43M & 67.8K \\
\addlinespace[0pt]
CricketZ & 2.69M & 2.71M & 2.07M & 52.5M & 1.74K & 170K & 240K & 6.81M & 68.8K \\
\addlinespace[0pt]
Earthquakes & 4.59M & 452K & 173K & 16.9M & 512 & 25.9K & 234K & 4.37M & 13.6K \\
\addlinespace[0pt]
EOGHorizontalSignal & 2.49M & 2.71M & 2.07M & 38.9M & 1.50K & 161K & 251K & 4.57M & 58.2K \\
\addlinespace[0pt]
EOGVerticalSignal & 2.49M & 2.71M & 2.07M & 32.2M & 1.35K & 192K & 262K & 7.22M & 55.0K \\
\addlinespace[0pt]
EthanolLevel & 5.29M & 860K & 676K & 39.7M & 952 & 152K & 186K & 2.40M & 32.6K \\
\addlinespace[0pt]
FaceAll & 3.92M & 3.17M & 2.42M & 71.8M & 1.52K & 178K & 271K & 4.63M & 76.4K \\
\addlinespace[0pt]
Fish & 1.04M & 1.51M & 1.18M & 12.2M & 878 & 177K & 224K & 3.00M & 49.2K \\
\addlinespace[0pt]
FordA & 27.0M & 452K & 173K & 85.6M & 1.34K & 163K & 255K & 1.47M & 27.7K \\
\addlinespace[0pt]
FordB & 27.3M & 452K & 173K & 79.2M & 1.41K & 175K & 261K & 1.81M & 30.1K \\
\addlinespace[0pt]
FreezerRegularTrain & 1.95M & 475K & 176K & 932K & 183 & 12.6K & 136K & 6.54M & 7.57K \\
\addlinespace[0pt]
GunPointAgeSpan & 962K & 467K & 175K & 5.64M & 750 & 120K & 200K & 4.47M & 47.1K \\
\addlinespace[0pt]
GunPointMaleVersusFemale & 962K & 467K & 175K & 2.26M & 348 & 56.3K & 176K & 4.38M & 10.8K \\
\addlinespace[0pt]
GunPointOldVersusYoung & 1.29M & 460K & 174K & 3.31M & 183 & 20.7K & 136K & 5.14M & 8.39K \\
\addlinespace[0pt]
Ham & 1.25M & 452K & 173K & 3.01M & 988 & 86.4K & 143K & 4.03M & 25.3K \\
\addlinespace[0pt]
HandOutlines & 8.54M & 430K & 169K & 29.6M & 1.48K & 148K & 242K & 2.46M & 33.8K \\
\addlinespace[0pt]
Haptics & 1.03M & 1.13M & 863K & 15.1M & 878 & 130K & 181K & 2.89M & 37.8K \\
\addlinespace[0pt]
InlineSkate & 672K & 1.61M & 1.22M & 3.35M & 1.02K & 147K & 219K & 3.81M & 31.9K \\
\addlinespace[0pt]
InsectWingbeatSound & 1.49M & 2.49M & 1.90M & 23.8M & 878 & 87.8K & 232K & 6.13M & 40.9K \\
\addlinespace[0pt]
LargeKitchenAppliances & 2.59M & 679K & 518K & 7.17M & 1.10K & 174K & 248K & 3.66M & 34.7K \\
\addlinespace[0pt]
MixedShapesRegularTrain & 3.36M & 1.11M & 857K & 37.1M & 1.17K & 184K & 191K & 5.83M & 36.5K \\
\addlinespace[0pt]
MixedShapesSmallTrain & 630K & 1.11M & 857K & 5.40M & 677 & 86.9K & 170K & 3.47M & 31.1K \\
\addlinespace[0pt]
NonInvasiveFetalECGThorax1 & 13.2M & 9.50M & 7.25M & 102M & 1.85K & 152K & 245K & 5.25M & 185K \\
\addlinespace[0pt]
NonInvasiveFetalECGThorax2 & 13.2M & 9.50M & 7.25M & 93.4M & 1.90K & 139K & 240K & 5.44M & 160K \\
\addlinespace[0pt]
OSULeaf & 2.33M & 1.36M & 1.04M & 26.0M & 988 & 132K & 136K & 4.30M & 46.4K \\
\addlinespace[0pt]
Plane & 689K & 1.58M & 1.21M & 4.41M & 824 & 107K & 221K & 4.68M & 27.3K \\
\addlinespace[0pt]
PowerCons & 964K & 452K & 173K & 7.97M & 366 & 96.4K & 177K & 4.40M & 52.8K \\
\addlinespace[0pt]
RefrigerationDevices & 2.59M & 679K & 518K & 42.2M & 1.63K & 165K & 160K & 7.74M & 39.9K \\
\addlinespace[0pt]
ScreenType & 2.49M & 668K & 514K & 22.4M & 1.45K & 116K & 217K & 3.09M & 37.3K \\
\addlinespace[0pt]
SemgHandGenderCh2 & 2.05M & 452K & 173K & 18.1M & 1.24K & 123K & 236K & 4.33M & 34.0K \\
\addlinespace[0pt]
SemgHandMovementCh2 & 3.13M & 1.36M & 1.04M & 60.2M & 1.90K & 176K & 237K & 5.67M & 53.9K \\
\addlinespace[0pt]
SemgHandSubjectCh2 & 3.13M & 1.13M & 863K & 54.2M & 1.56K & 157K & 164K & 4.82M & 40.4K \\
\addlinespace[0pt]
ShapesAll & 4.21M & 13.6M & 10.4M & 77.1M & 1.63K & 164K & 247K & 10.4M & 203K \\
\addlinespace[0pt]
SmallKitchenAppliances & 2.59M & 679K & 518K & 28.0M & 622 & 144K & 208K & 4.13M & 39.2K \\
\addlinespace[0pt]
StarLightCurves & 6.91M & 668K & 514K & 6.11M & 714 & 103K & 213K & 1.94M & 39.3K \\
\addlinespace[0pt]
Strawberry & 4.76M & 475K & 176K & 26.6M & 1.43K & 184K & 257K & 5.86M & 27.8K \\
\addlinespace[0pt]
SwedishLeaf & 3.49M & 3.39M & 2.59M & 32.9M & 1.28K & 175K & 263K & 4.57M & 65.5K \\
\addlinespace[0pt]
Trace & 937K & 920K & 696K & 3.83M & 439 & 42.5K & 204K & 5.05M & 9.25K \\
\addlinespace[0pt]
TwoPatterns & 7.16M & 905K & 691K & 59.6M & 1.59K & 180K & 216K & 2.79M & 32.5K \\
\addlinespace[0pt]
UWaveGestureLibraryAll & 6.39M & 1.81M & 1.38M & 78.8M & 1.67K & 169K & 270K & 4.94M & 71.7K \\
\addlinespace[0pt]
UWaveGestureLibraryX & 6.39M & 1.81M & 1.38M & 69.2M & 1.15K & 178K & 257K & 3.53M & 61.6K \\
\addlinespace[0pt]
UWaveGestureLibraryY & 6.39M & 1.81M & 1.38M & 112M & 1.37K & 181K & 234K & 7.52M & 65.6K \\
\addlinespace[0pt]
UWaveGestureLibraryZ & 5.50M & 1.69M & 1.34M & 46.0M & 1.54K & 146K & 228K & 2.89M & 47.9K \\
\addlinespace[0pt]
Wafer & 7.16M & 452K & 173K & 15.3M & 805 & 79.0K & 199K & 4.08M & 20.0K \\
\addlinespace[0pt]
Worms & 2.22M & 1.13M & 863K & 8.53M & 586 & 79.1K & 178K & 4.67M & 30.7K \\
\addlinespace[0pt]
WormsTwoClass & 1.05M & 452K & 173K & 11.2M & 952 & 125K & 205K & 3.33M & 31.6K \\
\addlinespace[0pt]
Yoga & 2.20M & 467K & 175K & 29.0M & 1.12K & 129K & 211K & 7.98M & 26.2K \\
\bottomrule
\end{tabularx}
}
\end{table}



\clearpage
\begin{table}[t]
\centering
\caption{Best-of-10 test balanced accuracy for TSFresh-10.}
\label{results-best-acc-tsfresh10}
\footnotesize{
\begin{tabularx}{\linewidth}{l*{9}{Y}}
\toprule
& KNN & NB & LR & SVM & DT & RF & AB & MLP & DLN \\ 
\midrule
ACSF1 & 0.780 & 0.690 & 0.640 & 0.750 & 0.670 & 0.810 & 0.610 & 0.730 & 0.780 \\
\addlinespace[0pt]
ChlorineConcentration & 0.647 & 0.431 & 0.448 & 0.709 & 0.552 & 0.531 & 0.462 & 0.655 & 0.549 \\
\addlinespace[0pt]
Computers & 0.680 & 0.668 & 0.708 & 0.676 & 0.656 & 0.692 & 0.664 & 0.652 & 0.700 \\
\addlinespace[0pt]
CricketX & 0.559 & 0.417 & 0.511 & 0.577 & 0.416 & 0.545 & 0.358 & 0.522 & 0.522 \\
\addlinespace[0pt]
CricketY & 0.594 & 0.481 & 0.557 & 0.627 & 0.480 & 0.602 & 0.419 & 0.577 & 0.576 \\
\addlinespace[0pt]
CricketZ & 0.579 & 0.472 & 0.515 & 0.569 & 0.443 & 0.530 & 0.402 & 0.506 & 0.531 \\
\addlinespace[0pt]
Earthquakes & 0.585 & 0.656 & 0.646 & 0.674 & 0.636 & 0.660 & 0.595 & 0.646 & 0.632 \\
\addlinespace[0pt]
EOGHorizontalSignal & 0.442 & 0.447 & 0.517 & 0.471 & 0.464 & 0.561 & 0.358 & 0.508 & 0.471 \\
\addlinespace[0pt]
EOGVerticalSignal & 0.420 & 0.361 & 0.422 & 0.384 & 0.378 & 0.434 & 0.410 & 0.425 & 0.396 \\
\addlinespace[0pt]
EthanolLevel & 0.326 & 0.343 & 0.364 & 0.364 & 0.343 & 0.382 & 0.380 & 0.369 & 0.368 \\
\addlinespace[0pt]
FaceAll & 0.722 & 0.659 & 0.691 & 0.739 & 0.607 & 0.724 & 0.552 & 0.706 & 0.728 \\
\addlinespace[0pt]
Fish & 0.838 & 0.838 & 0.840 & 0.849 & 0.730 & 0.818 & 0.699 & 0.814 & 0.838 \\
\addlinespace[0pt]
FordA & 1.00 & 1.00 & 1.00 & 1.00 & 1.00 & 1.00 & 1.00 & 1.00 & 1.00 \\
\addlinespace[0pt]
FordB & 0.738 & 0.732 & 0.773 & 0.793 & 0.731 & 0.754 & 0.750 & 0.787 & 0.753 \\
\addlinespace[0pt]
FreezerRegularTrain & 0.906 & 0.891 & 0.944 & 0.942 & 0.941 & 0.958 & 0.946 & 0.942 & 0.944 \\
\addlinespace[0pt]
GunPointAgeSpan & 0.965 & 0.880 & 0.915 & 0.965 & 0.927 & 0.953 & 0.943 & 0.978 & 0.972 \\
\addlinespace[0pt]
GunPointMaleVersusFemale & 0.948 & 0.780 & 0.967 & 0.978 & 0.978 & 0.994 & 0.985 & 0.991 & 0.994 \\
\addlinespace[0pt]
GunPointOldVersusYoung & 0.997 & 1.00 & 1.00 & 1.00 & 1.00 & 1.00 & 1.00 & 1.00 & 1.00 \\
\addlinespace[0pt]
Ham & 0.705 & 0.727 & 0.717 & 0.747 & 0.708 & 0.744 & 0.705 & 0.690 & 0.678 \\
\addlinespace[0pt]
HandOutlines & 0.825 & 0.614 & 0.764 & 0.869 & 0.838 & 0.860 & 0.815 & 0.863 & 0.872 \\
\addlinespace[0pt]
Haptics & 0.416 & 0.413 & 0.454 & 0.446 & 0.403 & 0.466 & 0.393 & 0.431 & 0.454 \\
\addlinespace[0pt]
InlineSkate & 0.425 & 0.341 & 0.317 & 0.416 & 0.349 & 0.376 & 0.280 & 0.355 & 0.441 \\
\addlinespace[0pt]
InsectWingbeatSound & 0.583 & 0.568 & 0.581 & 0.571 & 0.454 & 0.581 & 0.481 & 0.539 & 0.546 \\
\addlinespace[0pt]
LargeKitchenAppliances & 0.755 & 0.579 & 0.672 & 0.749 & 0.728 & 0.784 & 0.771 & 0.779 & 0.760 \\
\addlinespace[0pt]
MixedShapesRegularTrain & 0.935 & 0.879 & 0.911 & 0.936 & 0.884 & 0.917 & 0.859 & 0.917 & 0.930 \\
\addlinespace[0pt]
MixedShapesSmallTrain & 0.835 & 0.845 & 0.848 & 0.862 & 0.757 & 0.848 & 0.777 & 0.833 & 0.835 \\
\addlinespace[0pt]
NonInvasiveFetalECGThorax1 & 0.787 & 0.775 & 0.813 & 0.847 & 0.681 & 0.795 & 0.227 & 0.816 & 0.787 \\
\addlinespace[0pt]
NonInvasiveFetalECGThorax2 & 0.835 & 0.833 & 0.853 & 0.883 & 0.751 & 0.827 & 0.263 & 0.827 & 0.811 \\
\addlinespace[0pt]
OSULeaf & 0.871 & 0.882 & 0.917 & 0.901 & 0.803 & 0.868 & 0.697 & 0.861 & 0.873 \\
\addlinespace[0pt]
Plane & 0.943 & 0.959 & 0.977 & 0.980 & 0.904 & 0.973 & 0.808 & 0.956 & 0.980 \\
\addlinespace[0pt]
PowerCons & 0.876 & 0.808 & 0.869 & 0.876 & 0.813 & 0.841 & 0.855 & 0.890 & 0.876 \\
\addlinespace[0pt]
RefrigerationDevices & 0.605 & 0.477 & 0.605 & 0.619 & 0.584 & 0.632 & 0.600 & 0.613 & 0.611 \\
\addlinespace[0pt]
ScreenType & 0.467 & 0.392 & 0.448 & 0.456 & 0.515 & 0.541 & 0.536 & 0.453 & 0.493 \\
\addlinespace[0pt]
SemgHandGenderCh2 & 0.795 & 0.676 & 0.799 & 0.843 & 0.823 & 0.852 & 0.832 & 0.817 & 0.842 \\
\addlinespace[0pt]
SemgHandMovementCh2 & 0.873 & 0.407 & 0.587 & 0.873 & 0.736 & 0.804 & 0.469 & 0.831 & 0.811 \\
\addlinespace[0pt]
SemgHandSubjectCh2 & 0.916 & 0.660 & 0.791 & 0.918 & 0.842 & 0.880 & 0.800 & 0.900 & 0.904 \\
\addlinespace[0pt]
ShapesAll & 0.757 & 0.703 & 0.737 & 0.757 & 0.608 & 0.720 & 0.107 & 0.707 & 0.745 \\
\addlinespace[0pt]
SmallKitchenAppliances & 0.677 & 0.611 & 0.672 & 0.723 & 0.725 & 0.755 & 0.691 & 0.720 & 0.725 \\
\addlinespace[0pt]
StarLightCurves & 0.956 & 0.899 & 0.946 & 0.961 & 0.950 & 0.955 & 0.954 & 0.958 & 0.957 \\
\addlinespace[0pt]
Strawberry & 0.953 & 0.844 & 0.840 & 0.946 & 0.934 & 0.941 & 0.939 & 0.950 & 0.933 \\
\addlinespace[0pt]
SwedishLeaf & 0.871 & 0.823 & 0.892 & 0.901 & 0.810 & 0.871 & 0.609 & 0.890 & 0.880 \\
\addlinespace[0pt]
Trace & 0.791 & 0.867 & 0.930 & 0.934 & 0.867 & 0.881 & 0.750 & 0.880 & 0.894 \\
\addlinespace[0pt]
TwoPatterns & 0.999 & 0.963 & 0.999 & 0.999 & 0.980 & 0.995 & 0.747 & 0.999 & 0.998 \\
\addlinespace[0pt]
UWaveGestureLibraryAll & 0.952 & 0.928 & 0.921 & 0.961 & 0.884 & 0.950 & 0.885 & 0.936 & 0.947 \\
\addlinespace[0pt]
UWaveGestureLibraryX & 0.780 & 0.717 & 0.754 & 0.800 & 0.702 & 0.780 & 0.694 & 0.764 & 0.768 \\
\addlinespace[0pt]
UWaveGestureLibraryY & 0.685 & 0.593 & 0.667 & 0.716 & 0.625 & 0.686 & 0.547 & 0.679 & 0.666 \\
\addlinespace[0pt]
UWaveGestureLibraryZ & 0.711 & 0.648 & 0.708 & 0.726 & 0.651 & 0.714 & 0.620 & 0.714 & 0.705 \\
\addlinespace[0pt]
Wafer & 1.00 & 0.992 & 1.00 & 1.00 & 0.995 & 1.00 & 0.995 & 1.00 & 1.00 \\
\addlinespace[0pt]
Worms & 0.521 & 0.449 & 0.479 & 0.551 & 0.489 & 0.520 & 0.486 & 0.601 & 0.568 \\
\addlinespace[0pt]
WormsTwoClass & 0.803 & 0.777 & 0.750 & 0.826 & 0.746 & 0.795 & 0.818 & 0.864 & 0.803 \\
\addlinespace[0pt]
Yoga & 0.897 & 0.865 & 0.905 & 0.910 & 0.882 & 0.904 & 0.896 & 0.906 & 0.900 \\
\bottomrule
\end{tabularx}
}
\end{table}

\clearpage
\begingroup
\setlength{\tabcolsep}{3pt}
\footnotesize{
\begin{longtable}[c]{@{}Z{0.22\textwidth}*{9}{c}@{}}
\caption{Mean and standard deviation of balanced accuracy for TSFresh-10 across 10 random seeds.\label{results-mean-acc-tsfresh10}}\\
\toprule
& KNN & NB & LR & SVM & DT & RF & AB & MLP & DLN \\ \midrule
\endfirsthead
\multicolumn{10}{c}{\tablename\ \thetable\ (continued)}\\
\toprule
& KNN & NB & LR & SVM & DT & RF & AB & MLP & DLN \\ \midrule
\endhead
\midrule
\multicolumn{10}{r}{continued on next page}\\
\endfoot
\bottomrule
\endlastfoot
ACSF1 & \meanstd{0.739}{0.033} & \meanstd{0.690}{0.0} & \meanstd{0.618}{0.019} & \meanstd{0.708}{0.034} & \meanstd{0.597}{0.058} & \meanstd{0.732}{0.058} & \meanstd{0.525}{0.047} & \meanstd{0.694}{0.027} & \meanstd{0.723}{0.047} \\
\addlinespace[0pt]
ChlorineConcentration & \meanstd{0.637}{0.021} & \meanstd{0.431}{0.0} & \meanstd{0.446}{2.3e-03} & \meanstd{0.697}{0.011} & \meanstd{0.482}{0.044} & \meanstd{0.514}{0.013} & \meanstd{0.451}{9.9e-03} & \meanstd{0.618}{0.028} & \meanstd{0.526}{0.014} \\
\addlinespace[0pt]
Computers & \meanstd{0.649}{0.038} & \meanstd{0.668}{0.0} & \meanstd{0.693}{6.3e-03} & \meanstd{0.658}{0.017} & \meanstd{0.608}{0.035} & \meanstd{0.673}{0.013} & \meanstd{0.648}{7.9e-03} & \meanstd{0.637}{0.014} & \meanstd{0.654}{0.024} \\
\addlinespace[0pt]
CricketX & \meanstd{0.547}{0.012} & \meanstd{0.417}{0.0} & \meanstd{0.507}{2.2e-03} & \meanstd{0.563}{8.9e-03} & \meanstd{0.385}{0.014} & \meanstd{0.533}{8.0e-03} & \meanstd{0.330}{0.017} & \meanstd{0.500}{0.013} & \meanstd{0.489}{0.016} \\
\addlinespace[0pt]
CricketY & \meanstd{0.584}{0.011} & \meanstd{0.481}{0.0} & \meanstd{0.556}{4.4e-03} & \meanstd{0.613}{0.016} & \meanstd{0.458}{0.025} & \meanstd{0.577}{0.014} & \meanstd{0.395}{0.014} & \meanstd{0.539}{0.027} & \meanstd{0.546}{0.028} \\
\addlinespace[0pt]
CricketZ & \meanstd{0.562}{0.013} & \meanstd{0.472}{0.0} & \meanstd{0.508}{7.1e-03} & \meanstd{0.549}{0.019} & \meanstd{0.435}{9.9e-03} & \meanstd{0.509}{0.019} & \meanstd{0.370}{0.019} & \meanstd{0.483}{0.020} & \meanstd{0.503}{0.018} \\
\addlinespace[0pt]
EOGHorizontalSignal & \meanstd{0.428}{0.014} & \meanstd{0.447}{0.0} & \meanstd{0.505}{8.5e-03} & \meanstd{0.446}{0.014} & \meanstd{0.442}{0.019} & \meanstd{0.503}{0.025} & \meanstd{0.323}{0.028} & \meanstd{0.434}{0.044} & \meanstd{0.428}{0.026} \\
\addlinespace[0pt]
EOGVerticalSignal & \meanstd{0.401}{0.016} & \meanstd{0.361}{0.0} & \meanstd{0.411}{6.7e-03} & \meanstd{0.357}{0.020} & \meanstd{0.349}{0.018} & \meanstd{0.419}{0.013} & \meanstd{0.343}{0.042} & \meanstd{0.366}{0.034} & \meanstd{0.379}{0.012} \\
\addlinespace[0pt]
Earthquakes & \meanstd{0.557}{0.033} & \meanstd{0.656}{0.0} & \meanstd{0.622}{0.018} & \meanstd{0.637}{0.019} & \meanstd{0.603}{0.043} & \meanstd{0.647}{8.4e-03} & \meanstd{0.584}{6.8e-03} & \meanstd{0.584}{0.037} & \meanstd{0.594}{0.020} \\
\addlinespace[0pt]
EthanolLevel & \meanstd{0.307}{0.012} & \meanstd{0.343}{0.0} & \meanstd{0.362}{2.8e-03} & \meanstd{0.356}{9.2e-03} & \meanstd{0.316}{0.022} & \meanstd{0.363}{0.012} & \meanstd{0.363}{0.013} & \meanstd{0.334}{0.021} & \meanstd{0.351}{0.014} \\
\addlinespace[0pt]
FaceAll & \meanstd{0.718}{4.7e-03} & \meanstd{0.659}{0.0} & \meanstd{0.684}{4.3e-03} & \meanstd{0.729}{0.013} & \meanstd{0.580}{0.015} & \meanstd{0.707}{0.010} & \meanstd{0.498}{0.027} & \meanstd{0.687}{0.013} & \meanstd{0.716}{0.010} \\
\addlinespace[0pt]
Fish & \meanstd{0.819}{0.019} & \meanstd{0.838}{0.0} & \meanstd{0.819}{0.023} & \meanstd{0.831}{0.012} & \meanstd{0.688}{0.032} & \meanstd{0.784}{0.032} & \meanstd{0.658}{0.029} & \meanstd{0.772}{0.031} & \meanstd{0.792}{0.032} \\
\addlinespace[0pt]
FordA & \meanstd{1.00}{0.0} & \meanstd{1.00}{0.0} & \meanstd{1.00}{0.0} & \meanstd{1.00}{0.0} & \meanstd{1.00}{0.0} & \meanstd{1.00}{0.0} & \meanstd{1.00}{0.0} & \meanstd{1.00}{0.0} & \meanstd{1.00}{0.0} \\
\addlinespace[0pt]
FordB & \meanstd{0.728}{6.6e-03} & \meanstd{0.732}{0.0} & \meanstd{0.771}{1.1e-03} & \meanstd{0.785}{7.1e-03} & \meanstd{0.724}{7.7e-03} & \meanstd{0.748}{3.7e-03} & \meanstd{0.745}{4.6e-03} & \meanstd{0.773}{9.6e-03} & \meanstd{0.744}{6.0e-03} \\
\addlinespace[0pt]
FreezerRegularTrain & \meanstd{0.901}{4.7e-03} & \meanstd{0.891}{0.0} & \meanstd{0.934}{5.1e-03} & \meanstd{0.940}{2.0e-03} & \meanstd{0.921}{6.9e-03} & \meanstd{0.936}{0.016} & \meanstd{0.929}{8.4e-03} & \meanstd{0.933}{5.7e-03} & \meanstd{0.936}{5.2e-03} \\
\addlinespace[0pt]
GunPointAgeSpan & \meanstd{0.952}{0.018} & \meanstd{0.880}{0.0} & \meanstd{0.892}{0.017} & \meanstd{0.943}{0.025} & \meanstd{0.909}{0.014} & \meanstd{0.943}{4.9e-03} & \meanstd{0.935}{7.4e-03} & \meanstd{0.965}{9.1e-03} & \meanstd{0.957}{0.012} \\
\addlinespace[0pt]
GunPointMaleVsFemale & \meanstd{0.942}{0.011} & \meanstd{0.780}{0.0} & \meanstd{0.960}{3.7e-03} & \meanstd{0.970}{5.4e-03} & \meanstd{0.978}{0.0} & \meanstd{0.988}{2.7e-03} & \meanstd{0.985}{0.0} & \meanstd{0.978}{8.8e-03} & \meanstd{0.973}{0.015} \\
\addlinespace[0pt]
GunPointOldVsYoung & \meanstd{0.997}{0.0} & \meanstd{1.00}{0.0} & \meanstd{1.00}{0.0} & \meanstd{1.00}{0.0} & \meanstd{1.00}{0.0} & \meanstd{1.00}{0.0} & \meanstd{1.00}{1.1e-03} & \meanstd{1.00}{0.0} & \meanstd{0.999}{1.6e-03} \\
\addlinespace[0pt]
Ham & \meanstd{0.666}{0.029} & \meanstd{0.727}{0.0} & \meanstd{0.682}{0.022} & \meanstd{0.662}{0.041} & \meanstd{0.645}{0.057} & \meanstd{0.700}{0.035} & \meanstd{0.658}{0.032} & \meanstd{0.603}{0.058} & \meanstd{0.621}{0.036} \\
\addlinespace[0pt]
HandOutlines & \meanstd{0.825}{0.0} & \meanstd{0.614}{0.0} & \meanstd{0.764}{0.0} & \meanstd{0.853}{0.012} & \meanstd{0.822}{0.010} & \meanstd{0.849}{6.3e-03} & \meanstd{0.811}{4.5e-03} & \meanstd{0.850}{9.6e-03} & \meanstd{0.854}{0.015} \\
\addlinespace[0pt]
Haptics & \meanstd{0.406}{9.2e-03} & \meanstd{0.413}{0.0} & \meanstd{0.431}{0.018} & \meanstd{0.417}{0.027} & \meanstd{0.375}{0.020} & \meanstd{0.445}{0.018} & \meanstd{0.372}{0.015} & \meanstd{0.406}{0.024} & \meanstd{0.415}{0.023} \\
\addlinespace[0pt]
InlineSkate & \meanstd{0.413}{0.012} & \meanstd{0.341}{0.0} & \meanstd{0.304}{7.3e-03} & \meanstd{0.364}{0.034} & \meanstd{0.325}{0.019} & \meanstd{0.352}{0.018} & \meanstd{0.245}{0.019} & \meanstd{0.332}{0.024} & \meanstd{0.404}{0.022} \\
\addlinespace[0pt]
InsectWingbeatSound & \meanstd{0.572}{0.013} & \meanstd{0.568}{0.0} & \meanstd{0.555}{0.026} & \meanstd{0.550}{0.015} & \meanstd{0.442}{0.013} & \meanstd{0.568}{6.7e-03} & \meanstd{0.445}{0.027} & \meanstd{0.514}{0.015} & \meanstd{0.528}{0.016} \\
\addlinespace[0pt]
LargeKitchenAppliances & \meanstd{0.747}{7.2e-03} & \meanstd{0.579}{0.0} & \meanstd{0.664}{5.6e-03} & \meanstd{0.723}{0.020} & \meanstd{0.700}{0.021} & \meanstd{0.766}{9.9e-03} & \meanstd{0.753}{0.014} & \meanstd{0.730}{0.029} & \meanstd{0.743}{0.013} \\
\addlinespace[0pt]
MixedShapesRegTrain & \meanstd{0.931}{3.1e-03} & \meanstd{0.879}{0.0} & \meanstd{0.909}{6.3e-04} & \meanstd{0.930}{4.4e-03} & \meanstd{0.873}{5.5e-03} & \meanstd{0.911}{3.5e-03} & \meanstd{0.843}{0.011} & \meanstd{0.913}{5.0e-03} & \meanstd{0.918}{6.7e-03} \\
\addlinespace[0pt]
MixedShapesSmallTrain & \meanstd{0.830}{2.9e-03} & \meanstd{0.845}{0.0} & \meanstd{0.827}{0.018} & \meanstd{0.847}{7.3e-03} & \meanstd{0.728}{0.021} & \meanstd{0.842}{5.3e-03} & \meanstd{0.746}{0.017} & \meanstd{0.816}{0.028} & \meanstd{0.820}{0.012} \\
\addlinespace[0pt]
NonInvasiveFetalECG1 & \meanstd{0.781}{3.3e-03} & \meanstd{0.775}{0.0} & \meanstd{0.809}{1.3e-03} & \meanstd{0.837}{5.2e-03} & \meanstd{0.666}{8.1e-03} & \meanstd{0.767}{0.025} & \meanstd{0.172}{0.031} & \meanstd{0.804}{0.010} & \meanstd{0.766}{0.016} \\
\addlinespace[0pt]
NonInvasiveFetalECG2 & \meanstd{0.833}{2.1e-03} & \meanstd{0.833}{0.0} & \meanstd{0.840}{8.8e-03} & \meanstd{0.875}{5.5e-03} & \meanstd{0.740}{0.012} & \meanstd{0.806}{0.025} & \meanstd{0.161}{0.053} & \meanstd{0.815}{9.8e-03} & \meanstd{0.792}{0.017} \\
\addlinespace[0pt]
OSULeaf & \meanstd{0.863}{6.7e-03} & \meanstd{0.882}{0.0} & \meanstd{0.879}{0.043} & \meanstd{0.894}{5.7e-03} & \meanstd{0.788}{6.1e-03} & \meanstd{0.852}{0.011} & \meanstd{0.635}{0.033} & \meanstd{0.833}{0.025} & \meanstd{0.845}{0.019} \\
\addlinespace[0pt]
Plane & \meanstd{0.923}{0.016} & \meanstd{0.959}{0.0} & \meanstd{0.943}{0.039} & \meanstd{0.970}{5.0e-03} & \meanstd{0.890}{0.013} & \meanstd{0.950}{0.015} & \meanstd{0.674}{0.15} & \meanstd{0.926}{0.021} & \meanstd{0.960}{0.022} \\
\addlinespace[0pt]
PowerCons & \meanstd{0.849}{0.026} & \meanstd{0.808}{0.0} & \meanstd{0.863}{7.8e-03} & \meanstd{0.867}{5.8e-03} & \meanstd{0.812}{4.3e-03} & \meanstd{0.815}{0.013} & \meanstd{0.828}{0.016} & \meanstd{0.862}{0.024} & \meanstd{0.837}{0.021} \\
\addlinespace[0pt]
RefrigerationDevices & \meanstd{0.585}{0.021} & \meanstd{0.477}{0.0} & \meanstd{0.596}{0.014} & \meanstd{0.591}{0.028} & \meanstd{0.541}{0.023} & \meanstd{0.609}{0.011} & \meanstd{0.566}{0.023} & \meanstd{0.580}{0.021} & \meanstd{0.579}{0.020} \\
\addlinespace[0pt]
ScreenType & \meanstd{0.448}{0.017} & \meanstd{0.392}{0.0} & \meanstd{0.442}{2.8e-03} & \meanstd{0.417}{0.023} & \meanstd{0.483}{0.029} & \meanstd{0.532}{8.5e-03} & \meanstd{0.507}{0.021} & \meanstd{0.412}{0.023} & \meanstd{0.459}{0.026} \\
\addlinespace[0pt]
SemgHandGenderCh2 & \meanstd{0.789}{6.6e-03} & \meanstd{0.676}{0.0} & \meanstd{0.774}{0.022} & \meanstd{0.835}{8.8e-03} & \meanstd{0.796}{0.022} & \meanstd{0.840}{9.1e-03} & \meanstd{0.823}{8.7e-03} & \meanstd{0.785}{0.014} & \meanstd{0.827}{0.012} \\
\addlinespace[0pt]
SemgHandMovementCh2 & \meanstd{0.853}{0.017} & \meanstd{0.407}{0.0} & \meanstd{0.565}{0.010} & \meanstd{0.849}{0.021} & \meanstd{0.684}{0.025} & \meanstd{0.779}{0.038} & \meanstd{0.442}{0.020} & \meanstd{0.797}{0.020} & \meanstd{0.776}{0.017} \\
\addlinespace[0pt]
SemgHandSubjectCh2 & \meanstd{0.905}{5.8e-03} & \meanstd{0.660}{0.0} & \meanstd{0.787}{1.8e-03} & \meanstd{0.898}{0.023} & \meanstd{0.822}{0.017} & \meanstd{0.869}{7.6e-03} & \meanstd{0.778}{0.013} & \meanstd{0.865}{0.015} & \meanstd{0.882}{0.016} \\
\addlinespace[0pt]
ShapesAll & \meanstd{0.753}{2.7e-03} & \meanstd{0.703}{0.0} & \meanstd{0.727}{7.4e-03} & \meanstd{0.739}{7.0e-03} & \meanstd{0.580}{0.017} & \meanstd{0.678}{0.049} & \meanstd{0.0915}{0.013} & \meanstd{0.688}{0.013} & \meanstd{0.692}{0.036} \\
\addlinespace[0pt]
SmallKitchenAppliances & \meanstd{0.659}{0.015} & \meanstd{0.611}{0.0} & \meanstd{0.670}{1.1e-03} & \meanstd{0.704}{0.020} & \meanstd{0.706}{0.015} & \meanstd{0.737}{9.3e-03} & \meanstd{0.667}{0.014} & \meanstd{0.687}{0.028} & \meanstd{0.697}{0.014} \\
\addlinespace[0pt]
StarLightCurves & \meanstd{0.955}{9.3e-04} & \meanstd{0.899}{0.0} & \meanstd{0.944}{2.2e-03} & \meanstd{0.957}{2.4e-03} & \meanstd{0.949}{1.0e-03} & \meanstd{0.923}{0.093} & \meanstd{0.949}{3.3e-03} & \meanstd{0.953}{5.6e-03} & \meanstd{0.954}{2.6e-03} \\
\addlinespace[0pt]
Strawberry & \meanstd{0.946}{3.7e-03} & \meanstd{0.844}{0.0} & \meanstd{0.835}{3.0e-03} & \meanstd{0.937}{5.6e-03} & \meanstd{0.920}{0.011} & \meanstd{0.930}{8.0e-03} & \meanstd{0.921}{0.014} & \meanstd{0.927}{0.012} & \meanstd{0.926}{5.7e-03} \\
\addlinespace[0pt]
SwedishLeaf & \meanstd{0.862}{4.5e-03} & \meanstd{0.823}{0.0} & \meanstd{0.890}{1.1e-03} & \meanstd{0.896}{3.9e-03} & \meanstd{0.778}{0.016} & \meanstd{0.855}{0.016} & \meanstd{0.490}{0.079} & \meanstd{0.877}{8.2e-03} & \meanstd{0.860}{0.013} \\
\addlinespace[0pt]
Trace & \meanstd{0.777}{0.013} & \meanstd{0.867}{0.0} & \meanstd{0.918}{0.010} & \meanstd{0.918}{0.012} & \meanstd{0.858}{0.011} & \meanstd{0.854}{0.018} & \meanstd{0.750}{0.0} & \meanstd{0.859}{0.016} & \meanstd{0.876}{0.017} \\
\addlinespace[0pt]
TwoPatterns & \meanstd{0.998}{4.0e-04} & \meanstd{0.963}{0.0} & \meanstd{0.996}{3.2e-03} & \meanstd{0.999}{5.4e-04} & \meanstd{0.978}{1.5e-03} & \meanstd{0.992}{2.1e-03} & \meanstd{0.746}{9.9e-04} & \meanstd{0.997}{1.7e-03} & \meanstd{0.996}{9.4e-04} \\
\addlinespace[0pt]
UWaveGestureLibraryAll & \meanstd{0.947}{2.2e-03} & \meanstd{0.928}{0.0} & \meanstd{0.919}{2.7e-03} & \meanstd{0.957}{8.6e-03} & \meanstd{0.874}{7.7e-03} & \meanstd{0.937}{9.9e-03} & \meanstd{0.871}{8.2e-03} & \meanstd{0.929}{5.4e-03} & \meanstd{0.939}{5.6e-03} \\
\addlinespace[0pt]
UWaveGestureLibraryX & \meanstd{0.774}{3.7e-03} & \meanstd{0.717}{0.0} & \meanstd{0.751}{2.5e-03} & \meanstd{0.787}{6.9e-03} & \meanstd{0.695}{5.4e-03} & \meanstd{0.769}{9.2e-03} & \meanstd{0.683}{0.010} & \meanstd{0.743}{0.014} & \meanstd{0.760}{8.2e-03} \\
\addlinespace[0pt]
UWaveGestureLibraryY & \meanstd{0.680}{3.3e-03} & \meanstd{0.593}{0.0} & \meanstd{0.665}{7.6e-04} & \meanstd{0.705}{7.5e-03} & \meanstd{0.605}{0.012} & \meanstd{0.682}{5.1e-03} & \meanstd{0.535}{8.3e-03} & \meanstd{0.669}{6.0e-03} & \meanstd{0.655}{8.7e-03} \\
\addlinespace[0pt]
UWaveGestureLibraryZ & \meanstd{0.706}{6.0e-03} & \meanstd{0.648}{0.0} & \meanstd{0.703}{7.4e-03} & \meanstd{0.721}{5.6e-03} & \meanstd{0.640}{7.4e-03} & \meanstd{0.704}{0.011} & \meanstd{0.606}{0.010} & \meanstd{0.701}{0.010} & \meanstd{0.690}{0.011} \\
\addlinespace[0pt]
Wafer & \meanstd{1.00}{0.0} & \meanstd{0.992}{0.0} & \meanstd{1.00}{0.0} & \meanstd{1.00}{0.0} & \meanstd{0.993}{1.3e-03} & \meanstd{0.999}{1.7e-03} & \meanstd{0.995}{1.3e-03} & \meanstd{1.00}{0.0} & \meanstd{0.997}{5.2e-03} \\
\addlinespace[0pt]
Worms & \meanstd{0.484}{0.077} & \meanstd{0.449}{0.0} & \meanstd{0.457}{0.017} & \meanstd{0.526}{0.016} & \meanstd{0.405}{0.049} & \meanstd{0.442}{0.034} & \meanstd{0.428}{0.037} & \meanstd{0.501}{0.046} & \meanstd{0.440}{0.11} \\
\addlinespace[0pt]
WormsTwoClass & \meanstd{0.776}{0.018} & \meanstd{0.777}{0.0} & \meanstd{0.740}{3.6e-03} & \meanstd{0.782}{0.032} & \meanstd{0.718}{0.025} & \meanstd{0.764}{0.031} & \meanstd{0.771}{0.024} & \meanstd{0.761}{0.051} & \meanstd{0.756}{0.039} \\
\addlinespace[0pt]
Yoga & \meanstd{0.893}{3.3e-03} & \meanstd{0.865}{0.0} & \meanstd{0.905}{4.5e-04} & \meanstd{0.906}{2.1e-03} & \meanstd{0.865}{0.012} & \meanstd{0.902}{1.4e-03} & \meanstd{0.890}{3.2e-03} & \meanstd{0.896}{0.010} & \meanstd{0.890}{0.012} \\
\end{longtable}
}
\endgroup

\clearpage
\begin{table}[t]
\centering
\caption{Geometric mean of inference OPs for TSFresh-10 across 10 random seeds assuming FP16 for floating-point and INT16 for integer arithmetic.}
\label{results-ops-tsfresh10}
\footnotesize{
\begin{tabularx}{\linewidth}{l*{9}{Y}}
\toprule
& KNN & NB & LR & SVM & DT & RF & AB & MLP & DLN \\ 
\midrule
ACSF1 & 1.14M & 1.82M & 1.58M & 8.21M & 1.13K & 132K & 219K & 1.26M & 24.2K \\
\addlinespace[0pt]
ChlorineConcentration & 6.57M & 544K & 473K & 43.2M & 1.56K & 170K & 247K & 2.26M & 25.7K \\
\addlinespace[0pt]
Computers & 1.79M & 363K & 158K & 23.5M & 714 & 95.9K & 158K & 792K & 10.8K \\
\addlinespace[0pt]
CricketX & 2.84M & 2.18M & 1.89M & 50.4M & 1.56K & 170K & 229K & 1.60M & 38.9K \\
\addlinespace[0pt]
CricketY & 1.97M & 2.18M & 1.89M & 48.1M & 1.54K & 152K & 222K & 1.46M & 39.4K \\
\addlinespace[0pt]
CricketZ & 1.97M & 2.18M & 1.89M & 51.1M & 1.56K & 180K & 256K & 1.43M & 33.6K \\
\addlinespace[0pt]
Earthquakes & 4.13M & 363K & 158K & 15.0M & 549 & 33.2K & 227K & 740K & 4.91K \\
\addlinespace[0pt]
EOGHorizontalSignal & 1.42M & 2.18M & 1.89M & 36.5M & 1.30K & 163K & 239K & 1.73M & 39.1K \\
\addlinespace[0pt]
EOGVerticalSignal & 3.44M & 2.18M & 1.89M & 39.0M & 1.39K & 214K & 248K & 1.45M & 33.1K \\
\addlinespace[0pt]
EthanolLevel & 5.40M & 726K & 631K & 36.7M & 1.02K & 151K & 232K & 783K & 14.8K \\
\addlinespace[0pt]
FaceAll & 2.25M & 2.54M & 2.21M & 48.2M & 1.52K & 173K & 274K & 1.44M & 46.0K \\
\addlinespace[0pt]
Fish & 2.41M & 1.27M & 1.10M & 17.7M & 1.24K & 167K & 221K & 1.25M & 22.2K \\
\addlinespace[0pt]
FordA & 20.3M & 363K & 158K & 24.7M & 183 & 21.8K & 136K & 1.05M & 1.55K \\
\addlinespace[0pt]
FordB & 15.5M & 356K & 156K & 30.0M & 1.13K & 187K & 265K & 304K & 24.8K \\
\addlinespace[0pt]
FreezerRegularTrain & 2.23M & 363K & 158K & 1.93M & 238 & 37.5K & 191K & 977K & 7.96K \\
\addlinespace[0pt]
GunPointAgeSpan & 795K & 363K & 158K & 4.25M & 531 & 81.5K & 174K & 1.18M & 19.1K \\
\addlinespace[0pt]
GunPointMaleVersusFemale & 526K & 370K & 159K & 300K & 549 & 93.4K & 179K & 868K & 8.49K \\
\addlinespace[0pt]
GunPointOldVersusYoung & 747K & 378K & 160K & 1.45M & 183 & 23.9K & 136K & 1.48M & 1.57K \\
\addlinespace[0pt]
Ham & 1.12M & 363K & 158K & 3.81M & 586 & 92.5K & 170K & 910K & 13.7K \\
\addlinespace[0pt]
HandOutlines & 9.78M & 363K & 158K & 47.3M & 1.63K & 189K & 235K & 715K & 23.3K \\
\addlinespace[0pt]
Haptics & 745K & 908K & 789K & 11.2M & 878 & 92.3K & 187K & 1.59M & 17.3K \\
\addlinespace[0pt]
InlineSkate & 464K & 1.27M & 1.10M & 6.26M & 952 & 75.9K & 157K & 1.52M & 17.0K \\
\addlinespace[0pt]
InsectWingbeatSound & 1.08M & 2.00M & 1.74M & 10.9M & 1.32K & 135K & 248K & 1.24M & 29.0K \\
\addlinespace[0pt]
LargeKitchenAppliances & 1.47M & 544K & 473K & 23.8M & 1.19K & 143K & 227K & 1.78M & 19.0K \\
\addlinespace[0pt]
MixedShapesRegularTrain & 6.48M & 908K & 789K & 30.8M & 1.26K & 176K & 205K & 1.06M & 22.8K \\
\addlinespace[0pt]
MixedShapesSmallTrain & 469K & 908K & 789K & 7.59M & 732 & 93.5K & 212K & 1.12M & 16.4K \\
\addlinespace[0pt]
NonInvasiveFetalECGThorax1 & 23.9M & 7.62M & 6.62M & 180M & 1.96K & 159K & 261K & 2.12M & 98.9K \\
\addlinespace[0pt]
NonInvasiveFetalECGThorax2 & 7.80M & 7.62M & 6.62M & 153M & 1.99K & 164K & 270K & 2.16M & 76.3K \\
\addlinespace[0pt]
OSULeaf & 975K & 1.09M & 946K & 14.2M & 695 & 121K & 132K & 1.41M & 22.1K \\
\addlinespace[0pt]
Plane & 494K & 1.27M & 1.10M & 1.20M & 1.04K & 99.7K & 192K & 1.01M & 14.5K \\
\addlinespace[0pt]
PowerCons & 568K & 370K & 159K & 2.88M & 238 & 51.4K & 200K & 664K & 6.50K \\
\addlinespace[0pt]
RefrigerationDevices & 1.47M & 544K & 473K & 17.2M & 1.35K & 162K & 140K & 1.02M & 22.8K \\
\addlinespace[0pt]
ScreenType & 1.47M & 544K & 473K & 24.7M & 1.04K & 165K & 219K & 861K & 18.5K \\
\addlinespace[0pt]
SemgHandGenderCh2 & 1.83M & 363K & 158K & 19.4M & 878 & 128K & 199K & 1.36M & 14.2K \\
\addlinespace[0pt]
SemgHandMovementCh2 & 1.79M & 1.09M & 946K & 42.5M & 1.78K & 151K & 252K & 2.70M & 27.6K \\
\addlinespace[0pt]
SemgHandSubjectCh2 & 2.79M & 908K & 789K & 34.3M & 1.41K & 184K & 180K & 1.41M & 28.2K \\
\addlinespace[0pt]
ShapesAll & 2.43M & 10.9M & 9.46M & 55.4M & 1.68K & 184K & 240K & 2.54M & 118K \\
\addlinespace[0pt]
SmallKitchenAppliances & 3.15M & 544K & 473K & 33.1M & 1.21K & 141K & 178K & 1.21M & 19.2K \\
\addlinespace[0pt]
StarLightCurves & 9.78M & 544K & 473K & 13.7M & 549 & 123K & 174K & 591K & 7.71K \\
\addlinespace[0pt]
Strawberry & 5.92M & 363K & 158K & 21.1M & 1.12K & 129K & 251K & 1.22M & 18.7K \\
\addlinespace[0pt]
SwedishLeaf & 7.04M & 2.72M & 2.37M & 15.5M & 1.39K & 135K & 265K & 1.48M & 44.4K \\
\addlinespace[0pt]
Trace & 635K & 711K & 626K & 1.49M & 512 & 65.9K & 220K & 748K & 8.12K \\
\addlinespace[0pt]
TwoPatterns & 5.30M & 726K & 631K & 5.18M & 1.37K & 159K & 193K & 933K & 23.6K \\
\addlinespace[0pt]
UWaveGestureLibraryAll & 10.7M & 1.45M & 1.26M & 63.7M & 1.39K & 154K & 247K & 1.95M & 27.0K \\
\addlinespace[0pt]
UWaveGestureLibraryX & 6.73M & 1.45M & 1.26M & 59.4M & 1.74K & 168K & 210K & 1.36M & 33.8K \\
\addlinespace[0pt]
UWaveGestureLibraryY & 8.74M & 1.45M & 1.26M & 79.5M & 1.54K & 186K & 195K & 1.64M & 30.9K \\
\addlinespace[0pt]
UWaveGestureLibraryZ & 11.7M & 1.45M & 1.26M & 59.2M & 1.88K & 191K & 227K & 1.34M & 30.8K \\
\addlinespace[0pt]
Wafer & 5.30M & 363K & 158K & 8.31M & 183 & 24.5K & 136K & 1.05M & 2.60K \\
\addlinespace[0pt]
Worms & 1.47M & 908K & 789K & 13.8M & 695 & 50.0K & 215K & 779K & 9.47K \\
\addlinespace[0pt]
WormsTwoClass & 1.02M & 356K & 156K & 8.47M & 933 & 76.0K & 188K & 744K & 35.5K \\
\addlinespace[0pt]
Yoga & 1.60M & 370K & 159K & 3.82M & 805 & 132K & 157K & 1.22M & 6.91K \\
\bottomrule
\end{tabularx}
}
\end{table}



\clearpage
\begin{table}[t]
\centering
\caption{Best-of-10 test balanced accuracy for TSFresh-20.}
\label{results-best-acc-tsfresh20}
\footnotesize{
\begin{tabularx}{\linewidth}{l*{9}{Y}}
\toprule
& KNN & NB & LR & SVM & DT & RF & AB & MLP & DLN \\ 
\midrule
ACSF1 & 0.790 & 0.720 & 0.750 & 0.800 & 0.660 & 0.860 & 0.690 & 0.780 & 0.800 \\
\addlinespace[0pt]
ChlorineConcentration & 0.664 & 0.423 & 0.475 & 0.749 & 0.574 & 0.581 & 0.467 & 0.682 & 0.599 \\
\addlinespace[0pt]
Computers & 0.652 & 0.576 & 0.700 & 0.676 & 0.608 & 0.712 & 0.728 & 0.644 & 0.708 \\
\addlinespace[0pt]
CricketX & 0.654 & 0.542 & 0.594 & 0.671 & 0.463 & 0.647 & 0.465 & 0.626 & 0.594 \\
\addlinespace[0pt]
CricketY & 0.670 & 0.531 & 0.633 & 0.690 & 0.534 & 0.648 & 0.444 & 0.631 & 0.647 \\
\addlinespace[0pt]
CricketZ & 0.655 & 0.567 & 0.581 & 0.664 & 0.491 & 0.590 & 0.452 & 0.605 & 0.606 \\
\addlinespace[0pt]
Earthquakes & 0.580 & 0.650 & 0.679 & 0.674 & 0.636 & 0.650 & 0.595 & 0.685 & 0.660 \\
\addlinespace[0pt]
EOGHorizontalSignal & 0.459 & 0.478 & 0.499 & 0.490 & 0.512 & 0.542 & 0.348 & 0.515 & 0.503 \\
\addlinespace[0pt]
EOGVerticalSignal & 0.439 & 0.391 & 0.461 & 0.485 & 0.457 & 0.501 & 0.456 & 0.477 & 0.482 \\
\addlinespace[0pt]
EthanolLevel & 0.372 & 0.414 & 0.468 & 0.426 & 0.366 & 0.415 & 0.373 & 0.446 & 0.442 \\
\addlinespace[0pt]
FaceAll & 0.817 & 0.707 & 0.784 & 0.835 & 0.648 & 0.790 & 0.598 & 0.791 & 0.792 \\
\addlinespace[0pt]
Fish & 0.892 & 0.855 & 0.866 & 0.884 & 0.733 & 0.885 & 0.753 & 0.880 & 0.883 \\
\addlinespace[0pt]
FordA & 1.00 & 1.00 & 1.00 & 1.00 & 1.00 & 1.00 & 1.00 & 1.00 & 1.00 \\
\addlinespace[0pt]
FordB & 0.791 & 0.699 & 0.789 & 0.814 & 0.752 & 0.794 & 0.773 & 0.788 & 0.796 \\
\addlinespace[0pt]
FreezerRegularTrain & 0.902 & 0.890 & 0.939 & 0.943 & 0.919 & 0.970 & 0.964 & 0.940 & 0.945 \\
\addlinespace[0pt]
GunPointAgeSpan & 0.984 & 0.915 & 0.949 & 0.949 & 0.940 & 0.969 & 0.956 & 0.975 & 0.978 \\
\addlinespace[0pt]
GunPointMaleVersusFemale & 0.990 & 0.973 & 0.991 & 0.994 & 0.965 & 0.972 & 0.972 & 0.997 & 1.00 \\
\addlinespace[0pt]
GunPointOldVersusYoung & 1.00 & 1.00 & 1.00 & 1.00 & 1.00 & 1.00 & 1.00 & 1.00 & 1.00 \\
\addlinespace[0pt]
Ham & 0.692 & 0.722 & 0.667 & 0.667 & 0.693 & 0.725 & 0.668 & 0.630 & 0.681 \\
\addlinespace[0pt]
HandOutlines & 0.861 & 0.781 & 0.833 & 0.866 & 0.867 & 0.904 & 0.861 & 0.870 & 0.889 \\
\addlinespace[0pt]
Haptics & 0.419 & 0.365 & 0.441 & 0.502 & 0.412 & 0.480 & 0.390 & 0.465 & 0.480 \\
\addlinespace[0pt]
InlineSkate & 0.394 & 0.327 & 0.339 & 0.362 & 0.358 & 0.426 & 0.281 & 0.372 & 0.434 \\
\addlinespace[0pt]
InsectWingbeatSound & 0.617 & 0.593 & 0.602 & 0.612 & 0.509 & 0.639 & 0.513 & 0.571 & 0.595 \\
\addlinespace[0pt]
LargeKitchenAppliances & 0.853 & 0.672 & 0.805 & 0.867 & 0.829 & 0.891 & 0.837 & 0.875 & 0.864 \\
\addlinespace[0pt]
MixedShapesRegularTrain & 0.922 & 0.886 & 0.926 & 0.943 & 0.880 & 0.925 & 0.881 & 0.931 & 0.941 \\
\addlinespace[0pt]
MixedShapesSmallTrain & 0.879 & 0.879 & 0.894 & 0.901 & 0.813 & 0.883 & 0.816 & 0.891 & 0.882 \\
\addlinespace[0pt]
NonInvasiveFetalECGThorax1 & 0.828 & 0.821 & 0.861 & 0.878 & 0.744 & 0.844 & 0.160 & 0.860 & 0.832 \\
\addlinespace[0pt]
NonInvasiveFetalECGThorax2 & 0.870 & 0.881 & 0.902 & 0.910 & 0.778 & 0.871 & 0.227 & 0.885 & 0.877 \\
\addlinespace[0pt]
OSULeaf & 0.886 & 0.905 & 0.904 & 0.900 & 0.822 & 0.898 & 0.773 & 0.894 & 0.900 \\
\addlinespace[0pt]
Plane & 0.984 & 0.973 & 0.984 & 0.977 & 0.953 & 0.991 & 0.820 & 0.968 & 1.00 \\
\addlinespace[0pt]
PowerCons & 0.965 & 0.863 & 0.973 & 0.972 & 0.959 & 0.979 & 0.979 & 0.979 & 0.979 \\
\addlinespace[0pt]
RefrigerationDevices & 0.576 & 0.565 & 0.600 & 0.605 & 0.597 & 0.611 & 0.589 & 0.571 & 0.560 \\
\addlinespace[0pt]
ScreenType & 0.472 & 0.397 & 0.525 & 0.491 & 0.533 & 0.624 & 0.528 & 0.491 & 0.512 \\
\addlinespace[0pt]
SemgHandGenderCh2 & 0.845 & 0.660 & 0.773 & 0.862 & 0.868 & 0.885 & 0.878 & 0.854 & 0.879 \\
\addlinespace[0pt]
SemgHandMovementCh2 & 0.884 & 0.422 & 0.656 & 0.867 & 0.724 & 0.840 & 0.511 & 0.849 & 0.836 \\
\addlinespace[0pt]
SemgHandSubjectCh2 & 0.907 & 0.676 & 0.829 & 0.900 & 0.829 & 0.898 & 0.818 & 0.889 & 0.913 \\
\addlinespace[0pt]
ShapesAll & 0.832 & 0.742 & 0.820 & 0.823 & 0.620 & 0.777 & 0.130 & 0.800 & 0.798 \\
\addlinespace[0pt]
SmallKitchenAppliances & 0.643 & 0.613 & 0.661 & 0.669 & 0.731 & 0.752 & 0.693 & 0.661 & 0.704 \\
\addlinespace[0pt]
StarLightCurves & 0.961 & 0.896 & 0.946 & 0.959 & 0.946 & 0.956 & 0.953 & 0.959 & 0.961 \\
\addlinespace[0pt]
Strawberry & 0.962 & 0.835 & 0.918 & 0.967 & 0.940 & 0.936 & 0.960 & 0.958 & 0.952 \\
\addlinespace[0pt]
SwedishLeaf & 0.937 & 0.904 & 0.930 & 0.957 & 0.844 & 0.927 & 0.682 & 0.944 & 0.923 \\
\addlinespace[0pt]
Trace & 0.827 & 0.942 & 0.934 & 0.947 & 0.867 & 0.929 & 0.750 & 0.915 & 0.916 \\
\addlinespace[0pt]
TwoPatterns & 0.999 & 0.959 & 0.999 & 1.00 & 0.979 & 0.995 & 0.747 & 0.999 & 0.999 \\
\addlinespace[0pt]
UWaveGestureLibraryAll & 0.962 & 0.928 & 0.949 & 0.972 & 0.879 & 0.955 & 0.894 & 0.953 & 0.960 \\
\addlinespace[0pt]
UWaveGestureLibraryX & 0.787 & 0.717 & 0.775 & 0.817 & 0.718 & 0.800 & 0.714 & 0.780 & 0.789 \\
\addlinespace[0pt]
UWaveGestureLibraryY & 0.743 & 0.644 & 0.711 & 0.764 & 0.646 & 0.738 & 0.592 & 0.729 & 0.734 \\
\addlinespace[0pt]
UWaveGestureLibraryZ & 0.737 & 0.651 & 0.729 & 0.752 & 0.655 & 0.740 & 0.630 & 0.738 & 0.730 \\
\addlinespace[0pt]
Wafer & 1.00 & 0.992 & 1.00 & 1.00 & 0.995 & 1.00 & 0.995 & 1.00 & 1.00 \\
\addlinespace[0pt]
Worms & 0.597 & 0.470 & 0.644 & 0.662 & 0.571 & 0.599 & 0.394 & 0.659 & 0.602 \\
\addlinespace[0pt]
WormsTwoClass & 0.814 & 0.723 & 0.765 & 0.807 & 0.731 & 0.811 & 0.792 & 0.792 & 0.811 \\
\addlinespace[0pt]
Yoga & 0.896 & 0.879 & 0.904 & 0.904 & 0.868 & 0.908 & 0.893 & 0.898 & 0.907 \\
\bottomrule
\end{tabularx}
}
\end{table}

\clearpage
\begingroup
\setlength{\tabcolsep}{3pt}
\footnotesize{
\begin{longtable}[c]{@{}Z{0.22\textwidth}*{9}{c}@{}}
\caption{Mean and standard deviation of balanced accuracy for TSFresh-20 across 10 random seeds.\label{results-mean-acc-tsfresh20}}\\
\toprule
& KNN & NB & LR & SVM & DT & RF & AB & MLP & DLN \\ \midrule
\endfirsthead
\multicolumn{10}{c}{\tablename\ \thetable\ (continued)}\\
\toprule
& KNN & NB & LR & SVM & DT & RF & AB & MLP & DLN \\ \midrule
\endhead
\midrule
\multicolumn{10}{r}{continued on next page}\\
\endfoot
\bottomrule
\endlastfoot
ACSF1 & \meanstd{0.763}{0.024} & \meanstd{0.720}{0.0} & \meanstd{0.734}{9.7e-03} & \meanstd{0.778}{0.018} & \meanstd{0.630}{0.029} & \meanstd{0.837}{0.019} & \meanstd{0.554}{0.063} & \meanstd{0.763}{0.016} & \meanstd{0.745}{0.038} \\
\addlinespace[0pt]
ChlorineConcentration & \meanstd{0.662}{1.1e-03} & \meanstd{0.423}{0.0} & \meanstd{0.468}{6.9e-03} & \meanstd{0.744}{9.7e-03} & \meanstd{0.518}{0.043} & \meanstd{0.536}{0.028} & \meanstd{0.455}{7.4e-03} & \meanstd{0.662}{0.014} & \meanstd{0.570}{0.023} \\
\addlinespace[0pt]
Computers & \meanstd{0.641}{7.6e-03} & \meanstd{0.576}{0.0} & \meanstd{0.692}{3.0e-03} & \meanstd{0.663}{0.015} & \meanstd{0.576}{0.026} & \meanstd{0.683}{0.018} & \meanstd{0.696}{0.017} & \meanstd{0.621}{0.020} & \meanstd{0.650}{0.024} \\
\addlinespace[0pt]
CricketX & \meanstd{0.637}{0.020} & \meanstd{0.542}{0.0} & \meanstd{0.583}{6.7e-03} & \meanstd{0.636}{0.022} & \meanstd{0.447}{0.017} & \meanstd{0.601}{0.029} & \meanstd{0.417}{0.025} & \meanstd{0.590}{0.019} & \meanstd{0.574}{0.018} \\
\addlinespace[0pt]
CricketY & \meanstd{0.665}{3.1e-03} & \meanstd{0.531}{0.0} & \meanstd{0.600}{0.025} & \meanstd{0.674}{0.012} & \meanstd{0.495}{0.029} & \meanstd{0.622}{0.029} & \meanstd{0.410}{0.024} & \meanstd{0.621}{7.9e-03} & \meanstd{0.583}{0.031} \\
\addlinespace[0pt]
CricketZ & \meanstd{0.649}{6.2e-03} & \meanstd{0.567}{0.0} & \meanstd{0.549}{0.035} & \meanstd{0.644}{0.019} & \meanstd{0.472}{0.013} & \meanstd{0.570}{0.010} & \meanstd{0.412}{0.021} & \meanstd{0.571}{0.023} & \meanstd{0.574}{0.019} \\
\addlinespace[0pt]
EOGHorizontalSignal & \meanstd{0.422}{0.022} & \meanstd{0.478}{0.0} & \meanstd{0.497}{3.1e-03} & \meanstd{0.470}{0.013} & \meanstd{0.488}{0.024} & \meanstd{0.529}{0.014} & \meanstd{0.289}{0.050} & \meanstd{0.477}{0.026} & \meanstd{0.468}{0.025} \\
\addlinespace[0pt]
EOGVerticalSignal & \meanstd{0.425}{0.012} & \meanstd{0.391}{0.0} & \meanstd{0.455}{4.8e-03} & \meanstd{0.440}{0.033} & \meanstd{0.447}{7.9e-03} & \meanstd{0.478}{0.014} & \meanstd{0.406}{0.035} & \meanstd{0.441}{0.026} & \meanstd{0.447}{0.015} \\
\addlinespace[0pt]
Earthquakes & \meanstd{0.552}{0.024} & \meanstd{0.650}{0.0} & \meanstd{0.648}{0.018} & \meanstd{0.645}{0.017} & \meanstd{0.597}{0.047} & \meanstd{0.628}{0.015} & \meanstd{0.579}{0.011} & \meanstd{0.573}{0.062} & \meanstd{0.610}{0.034} \\
\addlinespace[0pt]
EthanolLevel & \meanstd{0.354}{0.010} & \meanstd{0.414}{0.0} & \meanstd{0.451}{0.015} & \meanstd{0.411}{8.5e-03} & \meanstd{0.344}{0.016} & \meanstd{0.391}{0.013} & \meanstd{0.349}{0.013} & \meanstd{0.408}{0.023} & \meanstd{0.412}{0.019} \\
\addlinespace[0pt]
FaceAll & \meanstd{0.811}{4.1e-03} & \meanstd{0.707}{0.0} & \meanstd{0.773}{8.7e-03} & \meanstd{0.826}{7.9e-03} & \meanstd{0.634}{7.6e-03} & \meanstd{0.775}{0.011} & \meanstd{0.525}{0.041} & \meanstd{0.779}{9.0e-03} & \meanstd{0.781}{8.2e-03} \\
\addlinespace[0pt]
Fish & \meanstd{0.867}{0.015} & \meanstd{0.855}{0.0} & \meanstd{0.848}{0.019} & \meanstd{0.859}{0.015} & \meanstd{0.688}{0.029} & \meanstd{0.852}{0.015} & \meanstd{0.709}{0.028} & \meanstd{0.841}{0.022} & \meanstd{0.841}{0.018} \\
\addlinespace[0pt]
FordA & \meanstd{1.00}{0.0} & \meanstd{1.00}{0.0} & \meanstd{1.00}{0.0} & \meanstd{1.00}{0.0} & \meanstd{1.00}{0.0} & \meanstd{1.00}{0.0} & \meanstd{1.00}{0.0} & \meanstd{1.00}{0.0} & \meanstd{1.00}{0.0} \\
\addlinespace[0pt]
FordB & \meanstd{0.781}{7.8e-03} & \meanstd{0.699}{0.0} & \meanstd{0.787}{8.7e-04} & \meanstd{0.807}{6.5e-03} & \meanstd{0.732}{0.014} & \meanstd{0.777}{0.014} & \meanstd{0.764}{7.2e-03} & \meanstd{0.769}{0.014} & \meanstd{0.782}{0.011} \\
\addlinespace[0pt]
FreezerRegularTrain & \meanstd{0.896}{3.0e-03} & \meanstd{0.890}{0.0} & \meanstd{0.931}{8.7e-03} & \meanstd{0.940}{3.1e-03} & \meanstd{0.919}{0.0} & \meanstd{0.949}{0.014} & \meanstd{0.946}{0.012} & \meanstd{0.929}{6.8e-03} & \meanstd{0.939}{6.3e-03} \\
\addlinespace[0pt]
GunPointAgeSpan & \meanstd{0.969}{8.6e-03} & \meanstd{0.915}{0.0} & \meanstd{0.937}{8.1e-03} & \meanstd{0.933}{9.8e-03} & \meanstd{0.939}{2.7e-03} & \meanstd{0.956}{6.4e-03} & \meanstd{0.950}{2.7e-03} & \meanstd{0.965}{9.7e-03} & \meanstd{0.955}{0.012} \\
\addlinespace[0pt]
GunPointMaleVsFemale & \meanstd{0.983}{5.7e-03} & \meanstd{0.973}{0.0} & \meanstd{0.991}{1.7e-04} & \meanstd{0.992}{1.6e-03} & \meanstd{0.965}{0.0} & \meanstd{0.968}{3.6e-03} & \meanstd{0.972}{0.0} & \meanstd{0.989}{5.4e-03} & \meanstd{0.993}{7.9e-03} \\
\addlinespace[0pt]
GunPointOldVsYoung & \meanstd{1.00}{0.0} & \meanstd{1.00}{0.0} & \meanstd{1.00}{0.0} & \meanstd{1.00}{0.0} & \meanstd{0.999}{1.4e-03} & \meanstd{1.00}{0.0} & \meanstd{0.999}{1.4e-03} & \meanstd{1.00}{0.0} & \meanstd{1.00}{1.1e-03} \\
\addlinespace[0pt]
Ham & \meanstd{0.629}{0.035} & \meanstd{0.722}{0.0} & \meanstd{0.607}{0.041} & \meanstd{0.616}{0.021} & \meanstd{0.599}{0.076} & \meanstd{0.669}{0.028} & \meanstd{0.641}{0.021} & \meanstd{0.577}{0.046} & \meanstd{0.640}{0.030} \\
\addlinespace[0pt]
HandOutlines & \meanstd{0.851}{8.9e-03} & \meanstd{0.781}{0.0} & \meanstd{0.830}{1.2e-03} & \meanstd{0.858}{7.1e-03} & \meanstd{0.841}{0.015} & \meanstd{0.897}{6.6e-03} & \meanstd{0.837}{0.010} & \meanstd{0.852}{0.017} & \meanstd{0.877}{9.6e-03} \\
\addlinespace[0pt]
Haptics & \meanstd{0.410}{5.1e-03} & \meanstd{0.365}{0.0} & \meanstd{0.434}{5.0e-03} & \meanstd{0.461}{0.023} & \meanstd{0.362}{0.023} & \meanstd{0.449}{0.022} & \meanstd{0.377}{0.012} & \meanstd{0.419}{0.035} & \meanstd{0.437}{0.023} \\
\addlinespace[0pt]
InlineSkate & \meanstd{0.375}{0.033} & \meanstd{0.327}{0.0} & \meanstd{0.335}{5.7e-03} & \meanstd{0.346}{0.014} & \meanstd{0.335}{0.014} & \meanstd{0.398}{0.029} & \meanstd{0.261}{0.025} & \meanstd{0.335}{0.027} & \meanstd{0.374}{0.043} \\
\addlinespace[0pt]
InsectWingbeatSound & \meanstd{0.608}{0.012} & \meanstd{0.593}{0.0} & \meanstd{0.556}{0.039} & \meanstd{0.561}{0.021} & \meanstd{0.500}{8.0e-03} & \meanstd{0.625}{8.0e-03} & \meanstd{0.452}{0.045} & \meanstd{0.543}{0.019} & \meanstd{0.572}{0.015} \\
\addlinespace[0pt]
LargeKitchenAppliances & \meanstd{0.841}{8.2e-03} & \meanstd{0.672}{0.0} & \meanstd{0.803}{3.3e-03} & \meanstd{0.846}{0.010} & \meanstd{0.764}{0.025} & \meanstd{0.878}{9.1e-03} & \meanstd{0.825}{8.7e-03} & \meanstd{0.849}{0.014} & \meanstd{0.851}{8.8e-03} \\
\addlinespace[0pt]
MixedShapesRegTrain & \meanstd{0.914}{7.2e-03} & \meanstd{0.886}{0.0} & \meanstd{0.918}{7.7e-03} & \meanstd{0.939}{1.8e-03} & \meanstd{0.864}{0.012} & \meanstd{0.921}{3.6e-03} & \meanstd{0.867}{0.011} & \meanstd{0.926}{4.4e-03} & \meanstd{0.933}{5.4e-03} \\
\addlinespace[0pt]
MixedShapesSmallTrain & \meanstd{0.876}{3.1e-03} & \meanstd{0.879}{0.0} & \meanstd{0.876}{0.013} & \meanstd{0.896}{2.9e-03} & \meanstd{0.784}{0.020} & \meanstd{0.874}{6.7e-03} & \meanstd{0.804}{9.3e-03} & \meanstd{0.879}{8.5e-03} & \meanstd{0.863}{0.014} \\
\addlinespace[0pt]
NonInvasiveFetalECG1 & \meanstd{0.824}{3.2e-03} & \meanstd{0.821}{0.0} & \meanstd{0.853}{7.2e-03} & \meanstd{0.876}{1.9e-03} & \meanstd{0.724}{0.012} & \meanstd{0.833}{0.012} & \meanstd{0.111}{0.025} & \meanstd{0.852}{6.2e-03} & \meanstd{0.825}{9.3e-03} \\
\addlinespace[0pt]
NonInvasiveFetalECG2 & \meanstd{0.867}{2.1e-03} & \meanstd{0.881}{0.0} & \meanstd{0.890}{9.8e-03} & \meanstd{0.906}{1.9e-03} & \meanstd{0.764}{0.013} & \meanstd{0.841}{0.031} & \meanstd{0.202}{0.023} & \meanstd{0.876}{6.9e-03} & \meanstd{0.861}{0.017} \\
\addlinespace[0pt]
OSULeaf & \meanstd{0.863}{0.013} & \meanstd{0.905}{0.0} & \meanstd{0.888}{0.023} & \meanstd{0.896}{3.7e-03} & \meanstd{0.801}{0.022} & \meanstd{0.887}{0.010} & \meanstd{0.763}{9.7e-03} & \meanstd{0.866}{0.018} & \meanstd{0.864}{0.020} \\
\addlinespace[0pt]
Plane & \meanstd{0.946}{0.024} & \meanstd{0.973}{0.0} & \meanstd{0.970}{0.013} & \meanstd{0.973}{4.3e-03} & \meanstd{0.898}{0.023} & \meanstd{0.980}{0.016} & \meanstd{0.713}{0.15} & \meanstd{0.951}{0.013} & \meanstd{0.974}{0.013} \\
\addlinespace[0pt]
PowerCons & \meanstd{0.952}{9.9e-03} & \meanstd{0.863}{0.0} & \meanstd{0.963}{8.7e-03} & \meanstd{0.966}{7.9e-03} & \meanstd{0.943}{0.020} & \meanstd{0.959}{0.025} & \meanstd{0.974}{4.7e-03} & \meanstd{0.968}{8.0e-03} & \meanstd{0.968}{9.3e-03} \\
\addlinespace[0pt]
RefrigerationDevices & \meanstd{0.552}{0.014} & \meanstd{0.565}{0.0} & \meanstd{0.587}{0.013} & \meanstd{0.577}{0.029} & \meanstd{0.555}{0.021} & \meanstd{0.593}{0.011} & \meanstd{0.550}{0.025} & \meanstd{0.534}{0.025} & \meanstd{0.541}{0.018} \\
\addlinespace[0pt]
ScreenType & \meanstd{0.458}{0.010} & \meanstd{0.397}{0.0} & \meanstd{0.498}{0.011} & \meanstd{0.473}{0.017} & \meanstd{0.502}{0.037} & \meanstd{0.570}{0.027} & \meanstd{0.507}{8.7e-03} & \meanstd{0.475}{0.012} & \meanstd{0.485}{0.020} \\
\addlinespace[0pt]
SemgHandGenderCh2 & \meanstd{0.826}{9.2e-03} & \meanstd{0.660}{0.0} & \meanstd{0.769}{2.9e-03} & \meanstd{0.847}{8.8e-03} & \meanstd{0.853}{0.015} & \meanstd{0.870}{0.012} & \meanstd{0.868}{8.3e-03} & \meanstd{0.835}{0.012} & \meanstd{0.869}{0.010} \\
\addlinespace[0pt]
SemgHandMovementCh2 & \meanstd{0.860}{0.016} & \meanstd{0.422}{0.0} & \meanstd{0.645}{3.7e-03} & \meanstd{0.843}{0.020} & \meanstd{0.716}{7.3e-03} & \meanstd{0.822}{0.017} & \meanstd{0.467}{0.029} & \meanstd{0.830}{0.011} & \meanstd{0.798}{0.027} \\
\addlinespace[0pt]
SemgHandSubjectCh2 & \meanstd{0.895}{9.5e-03} & \meanstd{0.676}{0.0} & \meanstd{0.820}{9.9e-03} & \meanstd{0.888}{0.015} & \meanstd{0.807}{0.016} & \meanstd{0.887}{0.011} & \meanstd{0.797}{0.012} & \meanstd{0.866}{0.020} & \meanstd{0.901}{6.9e-03} \\
\addlinespace[0pt]
ShapesAll & \meanstd{0.832}{0.0} & \meanstd{0.741}{8.1e-04} & \meanstd{0.802}{0.016} & \meanstd{0.815}{3.8e-03} & \meanstd{0.608}{6.7e-03} & \meanstd{0.712}{0.054} & \meanstd{0.103}{0.013} & \meanstd{0.782}{0.013} & \meanstd{0.784}{7.2e-03} \\
\addlinespace[0pt]
SmallKitchenAppliances & \meanstd{0.634}{0.014} & \meanstd{0.613}{0.0} & \meanstd{0.644}{6.3e-03} & \meanstd{0.661}{7.3e-03} & \meanstd{0.703}{0.023} & \meanstd{0.741}{6.1e-03} & \meanstd{0.673}{0.015} & \meanstd{0.628}{0.024} & \meanstd{0.668}{0.019} \\
\addlinespace[0pt]
StarLightCurves & \meanstd{0.960}{5.9e-04} & \meanstd{0.896}{0.0} & \meanstd{0.944}{2.1e-03} & \meanstd{0.958}{1.1e-03} & \meanstd{0.944}{2.0e-03} & \meanstd{0.954}{1.8e-03} & \meanstd{0.950}{2.9e-03} & \meanstd{0.955}{3.4e-03} & \meanstd{0.955}{3.2e-03} \\
\addlinespace[0pt]
Strawberry & \meanstd{0.957}{5.1e-03} & \meanstd{0.835}{0.0} & \meanstd{0.914}{2.0e-03} & \meanstd{0.957}{7.1e-03} & \meanstd{0.925}{7.7e-03} & \meanstd{0.928}{5.6e-03} & \meanstd{0.951}{4.9e-03} & \meanstd{0.945}{0.010} & \meanstd{0.942}{5.6e-03} \\
\addlinespace[0pt]
SwedishLeaf & \meanstd{0.927}{6.3e-03} & \meanstd{0.904}{0.0} & \meanstd{0.925}{3.2e-03} & \meanstd{0.942}{7.5e-03} & \meanstd{0.826}{0.014} & \meanstd{0.915}{9.8e-03} & \meanstd{0.604}{0.066} & \meanstd{0.924}{0.010} & \meanstd{0.912}{0.013} \\
\addlinespace[0pt]
Trace & \meanstd{0.820}{0.017} & \meanstd{0.942}{0.0} & \meanstd{0.900}{0.030} & \meanstd{0.935}{8.6e-03} & \meanstd{0.851}{0.013} & \meanstd{0.884}{0.029} & \meanstd{0.716}{0.013} & \meanstd{0.874}{0.022} & \meanstd{0.899}{0.021} \\
\addlinespace[0pt]
TwoPatterns & \meanstd{0.998}{1.8e-03} & \meanstd{0.959}{0.0} & \meanstd{0.995}{4.9e-03} & \meanstd{0.999}{3.2e-04} & \meanstd{0.978}{1.2e-03} & \meanstd{0.993}{2.8e-03} & \meanstd{0.746}{9.3e-04} & \meanstd{0.997}{2.3e-03} & \meanstd{0.997}{1.6e-03} \\
\addlinespace[0pt]
UWaveGestureLibraryAll & \meanstd{0.959}{2.2e-03} & \meanstd{0.928}{0.0} & \meanstd{0.944}{5.3e-03} & \meanstd{0.968}{5.0e-03} & \meanstd{0.874}{3.9e-03} & \meanstd{0.947}{4.5e-03} & \meanstd{0.879}{8.5e-03} & \meanstd{0.947}{5.3e-03} & \meanstd{0.953}{4.5e-03} \\
\addlinespace[0pt]
UWaveGestureLibraryX & \meanstd{0.786}{7.4e-04} & \meanstd{0.717}{0.0} & \meanstd{0.766}{4.9e-03} & \meanstd{0.808}{7.9e-03} & \meanstd{0.706}{6.2e-03} & \meanstd{0.785}{0.016} & \meanstd{0.707}{6.8e-03} & \meanstd{0.763}{8.3e-03} & \meanstd{0.778}{8.6e-03} \\
\addlinespace[0pt]
UWaveGestureLibraryY & \meanstd{0.742}{2.1e-03} & \meanstd{0.644}{0.0} & \meanstd{0.706}{2.8e-03} & \meanstd{0.755}{0.011} & \meanstd{0.639}{5.9e-03} & \meanstd{0.728}{7.8e-03} & \meanstd{0.567}{0.014} & \meanstd{0.713}{7.5e-03} & \meanstd{0.720}{7.6e-03} \\
\addlinespace[0pt]
UWaveGestureLibraryZ & \meanstd{0.730}{5.2e-03} & \meanstd{0.651}{0.0} & \meanstd{0.722}{7.2e-03} & \meanstd{0.745}{4.6e-03} & \meanstd{0.637}{0.011} & \meanstd{0.724}{0.016} & \meanstd{0.613}{0.014} & \meanstd{0.723}{9.2e-03} & \meanstd{0.721}{5.3e-03} \\
\addlinespace[0pt]
Wafer & \meanstd{1.00}{0.0} & \meanstd{0.992}{0.0} & \meanstd{1.00}{0.0} & \meanstd{1.00}{0.0} & \meanstd{0.994}{1.6e-03} & \meanstd{0.999}{7.8e-04} & \meanstd{0.994}{1.6e-03} & \meanstd{1.00}{5.1e-04} & \meanstd{0.998}{2.9e-03} \\
\addlinespace[0pt]
Worms & \meanstd{0.561}{0.072} & \meanstd{0.470}{0.0} & \meanstd{0.612}{0.046} & \meanstd{0.622}{0.022} & \meanstd{0.515}{0.031} & \meanstd{0.536}{0.029} & \meanstd{0.360}{0.029} & \meanstd{0.560}{0.064} & \meanstd{0.525}{0.055} \\
\addlinespace[0pt]
WormsTwoClass & \meanstd{0.769}{0.028} & \meanstd{0.723}{0.0} & \meanstd{0.755}{7.8e-03} & \meanstd{0.780}{0.028} & \meanstd{0.706}{0.029} & \meanstd{0.767}{0.031} & \meanstd{0.777}{0.013} & \meanstd{0.742}{0.035} & \meanstd{0.792}{0.013} \\
\addlinespace[0pt]
Yoga & \meanstd{0.895}{1.2e-03} & \meanstd{0.879}{0.0} & \meanstd{0.897}{4.2e-03} & \meanstd{0.899}{3.5e-03} & \meanstd{0.856}{8.9e-03} & \meanstd{0.902}{3.5e-03} & \meanstd{0.888}{3.7e-03} & \meanstd{0.888}{9.1e-03} & \meanstd{0.891}{9.6e-03} \\
\end{longtable}
}
\endgroup

\clearpage
\begin{table}[t]
\centering
\caption{Geometric mean of inference OPs for TSFresh-20 across 10 random seeds assuming FP16 for floating-point and INT16 for integer arithmetic.}
\label{results-ops-tsfresh20}
\footnotesize{
\begin{tabularx}{\linewidth}{l*{9}{Y}}
\toprule
& KNN & NB & LR & SVM & DT & RF & AB & MLP & DLN \\ 
\midrule
ACSF1 & 829K & 2.19M & 1.70M & 7.10M & 897 & 161K & 227K & 3.68M & 43.1K \\
\addlinespace[0pt]
ChlorineConcentration & 11.4M & 656K & 511K & 17.6M & 1.83K & 116K & 244K & 6.03M & 50.2K \\
\addlinespace[0pt]
Computers & 1.57M & 438K & 170K & 11.7M & 897 & 121K & 195K & 2.91M & 28.1K \\
\addlinespace[0pt]
CricketX & 2.50M & 2.63M & 2.04M & 37.7M & 1.50K & 171K & 248K & 4.79M & 64.7K \\
\addlinespace[0pt]
CricketY & 4.25M & 2.63M & 2.04M & 60.9M & 1.45K & 182K & 258K & 4.30M & 63.5K \\
\addlinespace[0pt]
CricketZ & 2.50M & 2.63M & 2.04M & 64.4M & 1.56K & 152K & 274K & 4.71M & 68.9K \\
\addlinespace[0pt]
Earthquakes & 6.43M & 423K & 168K & 10.3M & 897 & 15.7K & 217K & 1.64M & 24.2K \\
\addlinespace[0pt]
EOGHorizontalSignal & 2.31M & 2.63M & 2.04M & 37.7M & 1.24K & 176K & 214K & 4.47M & 65.3K \\
\addlinespace[0pt]
EOGVerticalSignal & 2.31M & 2.63M & 2.04M & 35.0M & 1.45K & 190K & 257K & 3.18M & 52.8K \\
\addlinespace[0pt]
EthanolLevel & 4.39M & 875K & 681K & 24.3M & 750 & 160K & 163K & 2.66M & 33.9K \\
\addlinespace[0pt]
FaceAll & 3.64M & 3.06M & 2.38M & 68.0M & 1.54K & 198K & 263K & 6.23M & 66.9K \\
\addlinespace[0pt]
Fish & 1.08M & 1.53M & 1.19M & 10.5M & 1.12K & 152K & 229K & 4.10M & 38.0K \\
\addlinespace[0pt]
FordA & 33.3M & 438K & 170K & 28.1M & 183 & 19.3K & 136K & 3.98M & 2.24K \\
\addlinespace[0pt]
FordB & 40.0M & 430K & 169K & 40.6M & 1.32K & 181K & 260K & 2.35M & 41.1K \\
\addlinespace[0pt]
FreezerRegularTrain & 2.94M & 438K & 170K & 2.02M & 183 & 49.7K & 159K & 3.35M & 8.21K \\
\addlinespace[0pt]
GunPointAgeSpan & 1.18M & 445K & 171K & 1.54M & 549 & 84.3K & 194K & 2.41M & 22.8K \\
\addlinespace[0pt]
GunPointMaleVersusFemale & 1.81M & 445K & 171K & 1.92M & 183 & 32.6K & 146K & 4.30M & 9.99K \\
\addlinespace[0pt]
GunPointOldVersusYoung & 1.24M & 452K & 173K & 3.37M & 183 & 24.1K & 136K & 4.78M & 6.71K \\
\addlinespace[0pt]
Ham & 906K & 438K & 170K & 2.73M & 512 & 88.6K & 161K & 3.16M & 29.7K \\
\addlinespace[0pt]
HandOutlines & 8.90M & 438K & 170K & 59.9M & 1.23K & 160K & 263K & 2.52M & 30.2K \\
\addlinespace[0pt]
Haptics & 2.34M & 1.09M & 851K & 17.1M & 952 & 126K & 225K & 3.30M & 34.1K \\
\addlinespace[0pt]
InlineSkate & 821K & 1.53M & 1.19M & 3.18M & 1.01K & 119K & 226K & 4.68M & 39.7K \\
\addlinespace[0pt]
InsectWingbeatSound & 1.38M & 2.41M & 1.87M & 25.8M & 1.15K & 153K & 256K & 4.49M & 52.7K \\
\addlinespace[0pt]
LargeKitchenAppliances & 2.40M & 656K & 511K & 15.8M & 1.13K & 177K & 241K & 5.25M & 39.3K \\
\addlinespace[0pt]
MixedShapesRegularTrain & 8.84M & 1.09M & 851K & 32.1M & 1.32K & 185K & 222K & 5.82M & 40.6K \\
\addlinespace[0pt]
MixedShapesSmallTrain & 2.85M & 1.09M & 851K & 3.97M & 677 & 102K & 188K & 3.45M & 39.8K \\
\addlinespace[0pt]
NonInvasiveFetalECGThorax1 & 12.3M & 9.19M & 7.15M & 250M & 1.88K & 191K & 246K & 5.46M & 175K \\
\addlinespace[0pt]
NonInvasiveFetalECGThorax2 & 12.3M & 9.19M & 7.15M & 139M & 1.90K & 148K & 251K & 5.26M & 162K \\
\addlinespace[0pt]
OSULeaf & 2.59M & 1.31M & 1.02M & 13.3M & 695 & 121K & 185K & 3.39M & 45.5K \\
\addlinespace[0pt]
Plane & 637K & 1.53M & 1.19M & 5.05M & 1.04K & 85.0K & 222K & 3.12M & 28.4K \\
\addlinespace[0pt]
PowerCons & 1.95M & 445K & 171K & 2.43M & 586 & 62.5K & 183K & 4.00M & 15.0K \\
\addlinespace[0pt]
RefrigerationDevices & 2.40M & 656K & 511K & 27.6M & 1.30K & 176K & 227K & 3.69M & 33.4K \\
\addlinespace[0pt]
ScreenType & 2.40M & 656K & 511K & 18.9M & 878 & 162K & 218K & 2.67M & 39.2K \\
\addlinespace[0pt]
SemgHandGenderCh2 & 1.90M & 438K & 170K & 13.1M & 952 & 143K & 210K & 2.43M & 53.6K \\
\addlinespace[0pt]
SemgHandMovementCh2 & 2.90M & 1.31M & 1.02M & 41.9M & 1.41K & 153K & 236K & 7.55M & 54.2K \\
\addlinespace[0pt]
SemgHandSubjectCh2 & 2.90M & 1.09M & 851K & 45.2M & 1.59K & 206K & 224K & 4.62M & 73.1K \\
\addlinespace[0pt]
ShapesAll & 4.06M & 13.4M & 10.3M & 87.8M & 1.72K & 160K & 245K & 7.68M & 204K \\
\addlinespace[0pt]
SmallKitchenAppliances & 3.24M & 656K & 511K & 24.7M & 915 & 180K & 175K & 2.98M & 30.9K \\
\addlinespace[0pt]
StarLightCurves & 6.66M & 656K & 511K & 27.4M & 403 & 120K & 149K & 4.72M & 24.0K \\
\addlinespace[0pt]
Strawberry & 10.9M & 438K & 170K & 14.9M & 1.10K & 120K & 235K & 4.41M & 24.6K \\
\addlinespace[0pt]
SwedishLeaf & 4.36M & 3.28M & 2.55M & 43.2M & 1.21K & 168K & 255K & 7.92M & 76.7K \\
\addlinespace[0pt]
Trace & 721K & 830K & 666K & 1.94M & 586 & 95.5K & 193K & 2.33M & 17.1K \\
\addlinespace[0pt]
TwoPatterns & 8.90M & 875K & 681K & 8.38M & 1.19K & 167K & 197K & 4.50M & 28.4K \\
\addlinespace[0pt]
UWaveGestureLibraryAll & 5.94M & 1.75M & 1.36M & 64.1M & 1.23K & 197K & 258K & 6.05M & 56.7K \\
\addlinespace[0pt]
UWaveGestureLibraryX & 5.94M & 1.75M & 1.36M & 85.7M & 1.61K & 164K & 242K & 3.15M & 54.3K \\
\addlinespace[0pt]
UWaveGestureLibraryY & 5.94M & 1.75M & 1.36M & 110M & 1.85K & 149K & 197K & 5.12M & 66.0K \\
\addlinespace[0pt]
UWaveGestureLibraryZ & 9.96M & 1.75M & 1.36M & 92.0M & 1.56K & 171K & 236K & 4.40M & 82.0K \\
\addlinespace[0pt]
Wafer & 8.90M & 438K & 170K & 9.72M & 183 & 23.7K & 136K & 3.98M & 4.59K \\
\addlinespace[0pt]
Worms & 4.16M & 1.09M & 851K & 15.2M & 512 & 70.3K & 211K & 2.53M & 25.7K \\
\addlinespace[0pt]
WormsTwoClass & 1.27M & 430K & 169K & 17.2M & 787 & 56.2K & 191K & 2.49M & 29.8K \\
\addlinespace[0pt]
Yoga & 1.98M & 445K & 171K & 6.16M & 897 & 156K & 195K & 3.83M & 17.2K \\
\bottomrule
\end{tabularx}
}
\end{table}


\clearpage
\begin{table}[t]
\centering
\caption{Best-of-10 test balanced accuracy for TSFresh-40.}
\label{results-best-acc-tsfresh40}
\footnotesize{
\begin{tabularx}{\linewidth}{l*{9}{Y}}
\toprule
& KNN & NB & LR & SVM & DT & RF & AB & MLP & DLN \\ 
\midrule
ACSF1 & 0.750 & 0.740 & 0.730 & 0.770 & 0.710 & 0.880 & 0.630 & 0.770 & 0.810 \\
\addlinespace[0pt]
ChlorineConcentration & 0.682 & 0.408 & 0.574 & 0.784 & 0.560 & 0.583 & 0.500 & 0.736 & 0.665 \\
\addlinespace[0pt]
Computers & 0.688 & 0.584 & 0.680 & 0.700 & 0.660 & 0.696 & 0.688 & 0.676 & 0.716 \\
\addlinespace[0pt]
CricketX & 0.642 & 0.555 & 0.624 & 0.657 & 0.478 & 0.624 & 0.458 & 0.632 & 0.631 \\
\addlinespace[0pt]
CricketY & 0.721 & 0.544 & 0.657 & 0.765 & 0.532 & 0.664 & 0.484 & 0.697 & 0.679 \\
\addlinespace[0pt]
CricketZ & 0.683 & 0.615 & 0.619 & 0.643 & 0.501 & 0.615 & 0.443 & 0.628 & 0.642 \\
\addlinespace[0pt]
Earthquakes & 0.566 & 0.659 & 0.689 & 0.707 & 0.702 & 0.689 & 0.557 & 0.600 & 0.718 \\
\addlinespace[0pt]
EOGHorizontalSignal & 0.456 & 0.492 & 0.537 & 0.537 & 0.551 & 0.562 & 0.338 & 0.554 & 0.573 \\
\addlinespace[0pt]
EOGVerticalSignal & 0.471 & 0.375 & 0.460 & 0.468 & 0.476 & 0.496 & 0.475 & 0.469 & 0.480 \\
\addlinespace[0pt]
EthanolLevel & 0.340 & 0.431 & 0.466 & 0.438 & 0.364 & 0.390 & 0.361 & 0.464 & 0.421 \\
\addlinespace[0pt]
FaceAll & 0.832 & 0.824 & 0.834 & 0.856 & 0.674 & 0.819 & 0.642 & 0.841 & 0.838 \\
\addlinespace[0pt]
Fish & 0.866 & 0.792 & 0.872 & 0.867 & 0.709 & 0.903 & 0.735 & 0.886 & 0.891 \\
\addlinespace[0pt]
FordA & 1.00 & 0.999 & 1.00 & 1.00 & 1.00 & 1.00 & 1.00 & 1.00 & 1.00 \\
\addlinespace[0pt]
FordB & 0.808 & 0.621 & 0.802 & 0.837 & 0.760 & 0.817 & 0.783 & 0.830 & 0.817 \\
\addlinespace[0pt]
FreezerRegularTrain & 0.870 & 0.861 & 0.946 & 0.953 & 0.945 & 0.976 & 0.974 & 0.945 & 0.951 \\
\addlinespace[0pt]
GunPointAgeSpan & 0.990 & 0.914 & 0.962 & 0.962 & 0.940 & 0.969 & 0.972 & 0.978 & 0.974 \\
\addlinespace[0pt]
GunPointMaleVersusFemale & 0.993 & 0.967 & 0.994 & 0.997 & 0.965 & 0.975 & 0.972 & 0.997 & 1.00 \\
\addlinespace[0pt]
GunPointOldVersusYoung & 1.00 & 1.00 & 1.00 & 1.00 & 1.00 & 1.00 & 1.00 & 1.00 & 1.00 \\
\addlinespace[0pt]
Ham & 0.654 & 0.733 & 0.771 & 0.655 & 0.697 & 0.742 & 0.715 & 0.643 & 0.711 \\
\addlinespace[0pt]
HandOutlines & 0.851 & 0.725 & 0.870 & 0.895 & 0.879 & 0.896 & 0.859 & 0.904 & 0.899 \\
\addlinespace[0pt]
Haptics & 0.438 & 0.361 & 0.474 & 0.495 & 0.434 & 0.508 & 0.459 & 0.487 & 0.492 \\
\addlinespace[0pt]
InlineSkate & 0.446 & 0.328 & 0.369 & 0.404 & 0.395 & 0.477 & 0.348 & 0.414 & 0.462 \\
\addlinespace[0pt]
InsectWingbeatSound & 0.636 & 0.604 & 0.630 & 0.636 & 0.497 & 0.656 & 0.472 & 0.599 & 0.637 \\
\addlinespace[0pt]
LargeKitchenAppliances & 0.821 & 0.683 & 0.816 & 0.861 & 0.861 & 0.891 & 0.872 & 0.835 & 0.869 \\
\addlinespace[0pt]
MixedShapesRegularTrain & 0.932 & 0.888 & 0.932 & 0.944 & 0.871 & 0.928 & 0.897 & 0.931 & 0.941 \\
\addlinespace[0pt]
MixedShapesSmallTrain & 0.872 & 0.877 & 0.889 & 0.901 & 0.813 & 0.890 & 0.823 & 0.897 & 0.888 \\
\addlinespace[0pt]
NonInvasiveFetalECGThorax1 & 0.860 & 0.860 & 0.891 & 0.901 & 0.765 & 0.873 & 0.216 & 0.893 & 0.883 \\
\addlinespace[0pt]
NonInvasiveFetalECGThorax2 & 0.891 & 0.895 & 0.915 & 0.924 & 0.798 & 0.894 & 0.296 & 0.904 & 0.915 \\
\addlinespace[0pt]
OSULeaf & 0.852 & 0.852 & 0.892 & 0.903 & 0.828 & 0.898 & 0.773 & 0.875 & 0.877 \\
\addlinespace[0pt]
Plane & 1.00 & 1.00 & 1.00 & 1.00 & 0.988 & 1.00 & 0.865 & 1.00 & 1.00 \\
\addlinespace[0pt]
PowerCons & 0.959 & 0.876 & 0.979 & 0.979 & 0.938 & 0.972 & 0.993 & 0.993 & 0.986 \\
\addlinespace[0pt]
RefrigerationDevices & 0.531 & 0.541 & 0.560 & 0.565 & 0.573 & 0.608 & 0.565 & 0.568 & 0.573 \\
\addlinespace[0pt]
ScreenType & 0.459 & 0.424 & 0.541 & 0.557 & 0.597 & 0.589 & 0.541 & 0.531 & 0.573 \\
\addlinespace[0pt]
SemgHandGenderCh2 & 0.864 & 0.664 & 0.873 & 0.891 & 0.871 & 0.925 & 0.932 & 0.884 & 0.912 \\
\addlinespace[0pt]
SemgHandMovementCh2 & 0.824 & 0.420 & 0.662 & 0.789 & 0.736 & 0.858 & 0.531 & 0.800 & 0.824 \\
\addlinespace[0pt]
SemgHandSubjectCh2 & 0.896 & 0.662 & 0.820 & 0.882 & 0.844 & 0.904 & 0.831 & 0.873 & 0.916 \\
\addlinespace[0pt]
ShapesAll & 0.762 & 0.752 & 0.760 & 0.765 & 0.602 & 0.788 & 0.148 & 0.762 & 0.808 \\
\addlinespace[0pt]
SmallKitchenAppliances & 0.725 & 0.731 & 0.784 & 0.773 & 0.731 & 0.795 & 0.792 & 0.765 & 0.813 \\
\addlinespace[0pt]
StarLightCurves & 0.961 & 0.928 & 0.955 & 0.962 & 0.945 & 0.958 & 0.958 & 0.960 & 0.961 \\
\addlinespace[0pt]
Strawberry & 0.968 & 0.841 & 0.937 & 0.971 & 0.950 & 0.953 & 0.980 & 0.966 & 0.964 \\
\addlinespace[0pt]
SwedishLeaf & 0.936 & 0.896 & 0.955 & 0.958 & 0.849 & 0.938 & 0.675 & 0.953 & 0.943 \\
\addlinespace[0pt]
Trace & 1.00 & 0.991 & 1.00 & 1.00 & 1.00 & 1.00 & 1.00 & 1.00 & 1.00 \\
\addlinespace[0pt]
TwoPatterns & 0.998 & 0.961 & 0.998 & 0.999 & 0.986 & 0.997 & 0.748 & 0.999 & 0.999 \\
\addlinespace[0pt]
UWaveGestureLibraryAll & 0.960 & 0.915 & 0.960 & 0.973 & 0.874 & 0.951 & 0.899 & 0.958 & 0.964 \\
\addlinespace[0pt]
UWaveGestureLibraryX & 0.779 & 0.711 & 0.779 & 0.811 & 0.743 & 0.806 & 0.728 & 0.783 & 0.802 \\
\addlinespace[0pt]
UWaveGestureLibraryY & 0.747 & 0.645 & 0.721 & 0.768 & 0.662 & 0.733 & 0.596 & 0.736 & 0.743 \\
\addlinespace[0pt]
UWaveGestureLibraryZ & 0.748 & 0.662 & 0.748 & 0.765 & 0.696 & 0.753 & 0.678 & 0.761 & 0.754 \\
\addlinespace[0pt]
Wafer & 1.00 & 0.992 & 1.00 & 1.00 & 0.995 & 1.00 & 0.995 & 1.00 & 1.00 \\
\addlinespace[0pt]
Worms & 0.631 & 0.522 & 0.667 & 0.679 & 0.604 & 0.673 & 0.637 & 0.698 & 0.693 \\
\addlinespace[0pt]
WormsTwoClass & 0.826 & 0.750 & 0.803 & 0.818 & 0.777 & 0.833 & 0.803 & 0.814 & 0.822 \\
\addlinespace[0pt]
Yoga & 0.908 & 0.894 & 0.897 & 0.906 & 0.876 & 0.897 & 0.903 & 0.907 & 0.913 \\
\bottomrule
\end{tabularx}
}
\end{table}

\clearpage
\begingroup
\setlength{\tabcolsep}{3pt}
\footnotesize{
\begin{longtable}[c]{@{}Z{0.22\textwidth}*{9}{c}@{}}
\caption{Mean and standard deviation of balanced accuracy for TSFresh-40 across 10 random seeds.\label{results-mean-acc-tsfresh40}}\\
\toprule
& KNN & NB & LR & SVM & DT & RF & AB & MLP & DLN \\ \midrule
\endfirsthead
\multicolumn{10}{c}{\tablename\ \thetable\ (continued)}\\
\toprule
& KNN & NB & LR & SVM & DT & RF & AB & MLP & DLN \\ \midrule
\endhead
\midrule
\multicolumn{10}{r}{continued on next page}\\
\endfoot
\bottomrule
\endlastfoot
ACSF1 & \meanstd{0.733}{0.023} & \meanstd{0.737}{6.7e-03} & \meanstd{0.685}{0.037} & \meanstd{0.742}{0.021} & \meanstd{0.660}{0.033} & \meanstd{0.848}{0.019} & \meanstd{0.521}{0.048} & \meanstd{0.726}{0.033} & \meanstd{0.761}{0.020} \\
\addlinespace[0pt]
ChlorineConcentration & \meanstd{0.682}{3.3e-05} & \meanstd{0.408}{0.0} & \meanstd{0.566}{0.010} & \meanstd{0.776}{0.010} & \meanstd{0.525}{0.031} & \meanstd{0.556}{0.026} & \meanstd{0.474}{0.015} & \meanstd{0.705}{0.018} & \meanstd{0.646}{0.014} \\
\addlinespace[0pt]
Computers & \meanstd{0.672}{0.013} & \meanstd{0.584}{0.0} & \meanstd{0.667}{7.8e-03} & \meanstd{0.674}{0.013} & \meanstd{0.620}{0.032} & \meanstd{0.679}{9.8e-03} & \meanstd{0.672}{0.012} & \meanstd{0.656}{0.014} & \meanstd{0.659}{0.035} \\
\addlinespace[0pt]
CricketX & \meanstd{0.636}{5.3e-03} & \meanstd{0.555}{0.0} & \meanstd{0.607}{0.017} & \meanstd{0.644}{9.1e-03} & \meanstd{0.452}{0.015} & \meanstd{0.597}{0.017} & \meanstd{0.408}{0.028} & \meanstd{0.603}{0.014} & \meanstd{0.602}{0.022} \\
\addlinespace[0pt]
CricketY & \meanstd{0.705}{0.013} & \meanstd{0.544}{0.0} & \meanstd{0.641}{0.018} & \meanstd{0.740}{0.025} & \meanstd{0.504}{0.016} & \meanstd{0.646}{0.012} & \meanstd{0.446}{0.019} & \meanstd{0.678}{0.018} & \meanstd{0.644}{0.024} \\
\addlinespace[0pt]
CricketZ & \meanstd{0.666}{0.010} & \meanstd{0.615}{0.0} & \meanstd{0.589}{0.028} & \meanstd{0.613}{0.022} & \meanstd{0.471}{0.029} & \meanstd{0.596}{0.022} & \meanstd{0.404}{0.020} & \meanstd{0.590}{0.026} & \meanstd{0.611}{0.020} \\
\addlinespace[0pt]
EOGHorizontalSignal & \meanstd{0.434}{7.9e-03} & \meanstd{0.492}{0.0} & \meanstd{0.514}{0.018} & \meanstd{0.520}{0.012} & \meanstd{0.520}{0.019} & \meanstd{0.545}{0.013} & \meanstd{0.295}{0.044} & \meanstd{0.505}{0.045} & \meanstd{0.503}{0.036} \\
\addlinespace[0pt]
EOGVerticalSignal & \meanstd{0.466}{4.1e-03} & \meanstd{0.375}{0.0} & \meanstd{0.442}{0.011} & \meanstd{0.448}{0.017} & \meanstd{0.425}{0.027} & \meanstd{0.489}{6.6e-03} & \meanstd{0.414}{0.029} & \meanstd{0.450}{0.018} & \meanstd{0.454}{0.022} \\
\addlinespace[0pt]
Earthquakes & \meanstd{0.539}{9.6e-03} & \meanstd{0.659}{0.0} & \meanstd{0.667}{0.013} & \meanstd{0.675}{0.017} & \meanstd{0.632}{0.055} & \meanstd{0.655}{0.016} & \meanstd{0.540}{0.016} & \meanstd{0.565}{0.025} & \meanstd{0.643}{0.043} \\
\addlinespace[0pt]
EthanolLevel & \meanstd{0.330}{4.8e-03} & \meanstd{0.431}{0.0} & \meanstd{0.454}{0.012} & \meanstd{0.410}{0.018} & \meanstd{0.348}{0.014} & \meanstd{0.377}{8.0e-03} & \meanstd{0.340}{0.012} & \meanstd{0.431}{0.021} & \meanstd{0.403}{0.016} \\
\addlinespace[0pt]
FaceAll & \meanstd{0.831}{4.5e-03} & \meanstd{0.824}{0.0} & \meanstd{0.816}{0.014} & \meanstd{0.850}{4.7e-03} & \meanstd{0.665}{6.9e-03} & \meanstd{0.808}{6.1e-03} & \meanstd{0.576}{0.037} & \meanstd{0.829}{8.9e-03} & \meanstd{0.827}{0.013} \\
\addlinespace[0pt]
Fish & \meanstd{0.851}{0.014} & \meanstd{0.788}{3.1e-03} & \meanstd{0.859}{5.9e-03} & \meanstd{0.854}{7.5e-03} & \meanstd{0.678}{0.019} & \meanstd{0.864}{0.022} & \meanstd{0.676}{0.029} & \meanstd{0.865}{0.017} & \meanstd{0.870}{0.017} \\
\addlinespace[0pt]
FordA & \meanstd{1.00}{0.0} & \meanstd{0.999}{0.0} & \meanstd{1.00}{0.0} & \meanstd{1.00}{0.0} & \meanstd{1.00}{0.0} & \meanstd{1.00}{0.0} & \meanstd{1.00}{0.0} & \meanstd{1.00}{0.0} & \meanstd{1.00}{0.0} \\
\addlinespace[0pt]
FordB & \meanstd{0.800}{3.6e-03} & \meanstd{0.621}{0.0} & \meanstd{0.800}{8.3e-04} & \meanstd{0.831}{5.8e-03} & \meanstd{0.739}{0.018} & \meanstd{0.803}{7.9e-03} & \meanstd{0.771}{9.1e-03} & \meanstd{0.807}{0.011} & \meanstd{0.805}{0.010} \\
\addlinespace[0pt]
FreezerRegularTrain & \meanstd{0.859}{7.0e-03} & \meanstd{0.861}{0.0} & \meanstd{0.942}{3.6e-03} & \meanstd{0.944}{4.6e-03} & \meanstd{0.924}{0.011} & \meanstd{0.952}{0.017} & \meanstd{0.963}{0.014} & \meanstd{0.937}{7.2e-03} & \meanstd{0.945}{3.6e-03} \\
\addlinespace[0pt]
GunPointAgeSpan & \meanstd{0.983}{3.2e-03} & \meanstd{0.914}{0.0} & \meanstd{0.938}{0.017} & \meanstd{0.935}{0.012} & \meanstd{0.940}{0.0} & \meanstd{0.963}{3.7e-03} & \meanstd{0.964}{9.6e-03} & \meanstd{0.968}{0.011} & \meanstd{0.962}{0.011} \\
\addlinespace[0pt]
GunPointMaleVsFemale & \meanstd{0.987}{6.6e-03} & \meanstd{0.967}{0.0} & \meanstd{0.993}{1.6e-03} & \meanstd{0.990}{9.3e-03} & \meanstd{0.965}{0.0} & \meanstd{0.969}{4.3e-03} & \meanstd{0.972}{0.0} & \meanstd{0.992}{4.1e-03} & \meanstd{0.993}{5.8e-03} \\
\addlinespace[0pt]
GunPointOldVsYoung & \meanstd{1.00}{0.0} & \meanstd{1.00}{0.0} & \meanstd{1.00}{0.0} & \meanstd{1.00}{0.0} & \meanstd{1.00}{1.1e-03} & \meanstd{1.00}{0.0} & \meanstd{0.999}{1.6e-03} & \meanstd{1.00}{0.0} & \meanstd{1.00}{0.0} \\
\addlinespace[0pt]
Ham & \meanstd{0.617}{0.031} & \meanstd{0.733}{0.0} & \meanstd{0.653}{0.060} & \meanstd{0.611}{0.022} & \meanstd{0.623}{0.052} & \meanstd{0.691}{0.029} & \meanstd{0.674}{0.026} & \meanstd{0.599}{0.033} & \meanstd{0.663}{0.023} \\
\addlinespace[0pt]
HandOutlines & \meanstd{0.851}{5.7e-04} & \meanstd{0.725}{0.0} & \meanstd{0.865}{3.5e-03} & \meanstd{0.886}{0.012} & \meanstd{0.869}{8.7e-03} & \meanstd{0.893}{1.9e-03} & \meanstd{0.845}{8.4e-03} & \meanstd{0.885}{0.016} & \meanstd{0.880}{0.011} \\
\addlinespace[0pt]
Haptics & \meanstd{0.398}{0.027} & \meanstd{0.361}{0.0} & \meanstd{0.434}{0.037} & \meanstd{0.462}{0.038} & \meanstd{0.394}{0.036} & \meanstd{0.469}{0.019} & \meanstd{0.430}{0.017} & \meanstd{0.439}{0.023} & \meanstd{0.467}{0.016} \\
\addlinespace[0pt]
InlineSkate & \meanstd{0.426}{0.028} & \meanstd{0.328}{0.0} & \meanstd{0.338}{0.025} & \meanstd{0.384}{0.011} & \meanstd{0.368}{0.026} & \meanstd{0.440}{0.039} & \meanstd{0.313}{0.026} & \meanstd{0.366}{0.023} & \meanstd{0.409}{0.029} \\
\addlinespace[0pt]
InsectWingbeatSound & \meanstd{0.630}{4.2e-03} & \meanstd{0.604}{0.0} & \meanstd{0.589}{0.042} & \meanstd{0.601}{0.023} & \meanstd{0.478}{0.017} & \meanstd{0.644}{9.9e-03} & \meanstd{0.413}{0.045} & \meanstd{0.574}{0.017} & \meanstd{0.609}{0.016} \\
\addlinespace[0pt]
LargeKitchenAppliances & \meanstd{0.816}{9.0e-03} & \meanstd{0.683}{0.0} & \meanstd{0.799}{0.010} & \meanstd{0.848}{0.010} & \meanstd{0.791}{0.045} & \meanstd{0.877}{8.3e-03} & \meanstd{0.850}{0.011} & \meanstd{0.821}{7.8e-03} & \meanstd{0.853}{0.018} \\
\addlinespace[0pt]
MixedShapesRegTrain & \meanstd{0.926}{7.8e-03} & \meanstd{0.888}{0.0} & \meanstd{0.927}{6.2e-03} & \meanstd{0.936}{5.7e-03} & \meanstd{0.854}{0.013} & \meanstd{0.922}{4.5e-03} & \meanstd{0.884}{7.2e-03} & \meanstd{0.928}{2.9e-03} & \meanstd{0.931}{8.1e-03} \\
\addlinespace[0pt]
MixedShapesSmallTrain & \meanstd{0.865}{4.1e-03} & \meanstd{0.877}{0.0} & \meanstd{0.884}{4.4e-03} & \meanstd{0.895}{3.2e-03} & \meanstd{0.774}{0.024} & \meanstd{0.880}{8.1e-03} & \meanstd{0.797}{0.021} & \meanstd{0.882}{0.014} & \meanstd{0.875}{0.010} \\
\addlinespace[0pt]
NonInvasiveFetalECG1 & \meanstd{0.858}{2.6e-03} & \meanstd{0.860}{0.0} & \meanstd{0.876}{0.014} & \meanstd{0.897}{2.7e-03} & \meanstd{0.744}{0.016} & \meanstd{0.852}{0.017} & \meanstd{0.183}{0.029} & \meanstd{0.883}{7.2e-03} & \meanstd{0.869}{8.3e-03} \\
\addlinespace[0pt]
NonInvasiveFetalECG2 & \meanstd{0.889}{2.9e-03} & \meanstd{0.895}{0.0} & \meanstd{0.903}{6.4e-03} & \meanstd{0.918}{2.5e-03} & \meanstd{0.786}{9.0e-03} & \meanstd{0.863}{0.029} & \meanstd{0.246}{0.043} & \meanstd{0.900}{2.8e-03} & \meanstd{0.898}{9.6e-03} \\
\addlinespace[0pt]
OSULeaf & \meanstd{0.829}{0.017} & \meanstd{0.852}{0.0} & \meanstd{0.883}{7.1e-03} & \meanstd{0.882}{0.010} & \meanstd{0.807}{0.013} & \meanstd{0.882}{8.5e-03} & \meanstd{0.758}{0.013} & \meanstd{0.843}{0.022} & \meanstd{0.861}{0.015} \\
\addlinespace[0pt]
Plane & \meanstd{0.981}{0.014} & \meanstd{1.00}{0.0} & \meanstd{0.986}{8.2e-03} & \meanstd{0.994}{7.4e-03} & \meanstd{0.963}{0.016} & \meanstd{0.991}{0.013} & \meanstd{0.837}{0.051} & \meanstd{0.989}{0.013} & \meanstd{0.988}{0.014} \\
\addlinespace[0pt]
PowerCons & \meanstd{0.953}{0.013} & \meanstd{0.876}{0.0} & \meanstd{0.972}{0.011} & \meanstd{0.959}{0.013} & \meanstd{0.898}{0.028} & \meanstd{0.944}{0.030} & \meanstd{0.977}{0.012} & \meanstd{0.968}{0.017} & \meanstd{0.973}{0.016} \\
\addlinespace[0pt]
RefrigerationDevices & \meanstd{0.515}{0.014} & \meanstd{0.541}{0.0} & \meanstd{0.533}{0.013} & \meanstd{0.549}{0.016} & \meanstd{0.547}{0.022} & \meanstd{0.586}{0.012} & \meanstd{0.530}{0.015} & \meanstd{0.527}{0.023} & \meanstd{0.537}{0.019} \\
\addlinespace[0pt]
ScreenType & \meanstd{0.448}{9.8e-03} & \meanstd{0.424}{0.0} & \meanstd{0.528}{6.9e-03} & \meanstd{0.524}{0.025} & \meanstd{0.504}{0.049} & \meanstd{0.559}{0.018} & \meanstd{0.524}{0.016} & \meanstd{0.489}{0.026} & \meanstd{0.521}{0.025} \\
\addlinespace[0pt]
SemgHandGenderCh2 & \meanstd{0.852}{0.013} & \meanstd{0.664}{0.0} & \meanstd{0.852}{0.033} & \meanstd{0.877}{8.0e-03} & \meanstd{0.856}{0.016} & \meanstd{0.914}{0.011} & \meanstd{0.919}{6.2e-03} & \meanstd{0.865}{0.014} & \meanstd{0.905}{7.8e-03} \\
\addlinespace[0pt]
SemgHandMovementCh2 & \meanstd{0.820}{5.5e-03} & \meanstd{0.420}{0.0} & \meanstd{0.648}{7.9e-03} & \meanstd{0.759}{0.031} & \meanstd{0.705}{0.030} & \meanstd{0.832}{0.018} & \meanstd{0.502}{0.018} & \meanstd{0.770}{0.022} & \meanstd{0.790}{0.031} \\
\addlinespace[0pt]
SemgHandSubjectCh2 & \meanstd{0.889}{6.1e-03} & \meanstd{0.662}{0.0} & \meanstd{0.805}{0.010} & \meanstd{0.868}{9.9e-03} & \meanstd{0.820}{0.021} & \meanstd{0.893}{0.014} & \meanstd{0.803}{0.013} & \meanstd{0.859}{8.8e-03} & \meanstd{0.901}{0.011} \\
\addlinespace[0pt]
ShapesAll & \meanstd{0.762}{0.0} & \meanstd{0.750}{2.8e-03} & \meanstd{0.738}{0.020} & \meanstd{0.756}{4.7e-03} & \meanstd{0.575}{0.012} & \meanstd{0.745}{0.028} & \meanstd{0.107}{0.018} & \meanstd{0.743}{0.010} & \meanstd{0.783}{0.020} \\
\addlinespace[0pt]
SmallKitchenAppliances & \meanstd{0.709}{0.016} & \meanstd{0.731}{0.0} & \meanstd{0.773}{5.5e-03} & \meanstd{0.763}{0.015} & \meanstd{0.711}{0.019} & \meanstd{0.785}{8.9e-03} & \meanstd{0.783}{5.5e-03} & \meanstd{0.743}{0.015} & \meanstd{0.774}{0.021} \\
\addlinespace[0pt]
StarLightCurves & \meanstd{0.960}{2.7e-04} & \meanstd{0.928}{0.0} & \meanstd{0.953}{1.6e-03} & \meanstd{0.960}{1.7e-03} & \meanstd{0.944}{6.1e-04} & \meanstd{0.956}{1.6e-03} & \meanstd{0.953}{2.5e-03} & \meanstd{0.958}{1.5e-03} & \meanstd{0.957}{2.8e-03} \\
\addlinespace[0pt]
Strawberry & \meanstd{0.960}{6.1e-03} & \meanstd{0.841}{0.0} & \meanstd{0.928}{5.2e-03} & \meanstd{0.955}{0.013} & \meanstd{0.931}{0.014} & \meanstd{0.941}{5.7e-03} & \meanstd{0.966}{0.010} & \meanstd{0.955}{6.9e-03} & \meanstd{0.952}{9.8e-03} \\
\addlinespace[0pt]
SwedishLeaf & \meanstd{0.933}{2.6e-03} & \meanstd{0.893}{4.4e-03} & \meanstd{0.946}{6.7e-03} & \meanstd{0.954}{2.3e-03} & \meanstd{0.839}{7.1e-03} & \meanstd{0.930}{4.8e-03} & \meanstd{0.603}{0.038} & \meanstd{0.936}{9.6e-03} & \meanstd{0.933}{5.5e-03} \\
\addlinespace[0pt]
Trace & \meanstd{0.997}{6.0e-03} & \meanstd{0.991}{0.0} & \meanstd{1.00}{0.0} & \meanstd{1.00}{0.0} & \meanstd{0.981}{0.018} & \meanstd{1.00}{0.0} & \meanstd{0.852}{0.13} & \meanstd{1.00}{0.0} & \meanstd{0.995}{0.012} \\
\addlinespace[0pt]
TwoPatterns & \meanstd{0.998}{2.1e-04} & \meanstd{0.961}{0.0} & \meanstd{0.993}{6.3e-03} & \meanstd{0.999}{1.8e-04} & \meanstd{0.980}{4.0e-03} & \meanstd{0.996}{2.3e-03} & \meanstd{0.746}{1.1e-03} & \meanstd{0.997}{1.5e-03} & \meanstd{0.996}{1.5e-03} \\
\addlinespace[0pt]
UWaveGestureLibraryAll & \meanstd{0.952}{3.7e-03} & \meanstd{0.915}{0.0} & \meanstd{0.951}{0.012} & \meanstd{0.970}{4.2e-03} & \meanstd{0.865}{6.6e-03} & \meanstd{0.946}{5.1e-03} & \meanstd{0.886}{9.0e-03} & \meanstd{0.955}{2.1e-03} & \meanstd{0.958}{3.9e-03} \\
\addlinespace[0pt]
UWaveGestureLibraryX & \meanstd{0.773}{3.4e-03} & \meanstd{0.711}{0.0} & \meanstd{0.770}{6.5e-03} & \meanstd{0.798}{0.012} & \meanstd{0.715}{0.016} & \meanstd{0.798}{7.9e-03} & \meanstd{0.718}{7.5e-03} & \meanstd{0.773}{7.4e-03} & \meanstd{0.787}{0.010} \\
\addlinespace[0pt]
UWaveGestureLibraryY & \meanstd{0.738}{8.5e-03} & \meanstd{0.645}{0.0} & \meanstd{0.712}{7.1e-03} & \meanstd{0.760}{6.9e-03} & \meanstd{0.654}{7.1e-03} & \meanstd{0.727}{4.5e-03} & \meanstd{0.582}{0.012} & \meanstd{0.724}{0.010} & \meanstd{0.728}{0.012} \\
\addlinespace[0pt]
UWaveGestureLibraryZ & \meanstd{0.738}{7.5e-03} & \meanstd{0.662}{0.0} & \meanstd{0.739}{7.6e-03} & \meanstd{0.758}{6.3e-03} & \meanstd{0.675}{0.017} & \meanstd{0.739}{0.026} & \meanstd{0.646}{0.016} & \meanstd{0.749}{0.012} & \meanstd{0.741}{0.010} \\
\addlinespace[0pt]
Wafer & \meanstd{1.00}{0.0} & \meanstd{0.992}{0.0} & \meanstd{1.00}{0.0} & \meanstd{1.00}{0.0} & \meanstd{0.995}{1.3e-03} & \meanstd{0.999}{1.0e-03} & \meanstd{0.995}{1.5e-03} & \meanstd{1.00}{5.8e-05} & \meanstd{0.999}{1.6e-03} \\
\addlinespace[0pt]
Worms & \meanstd{0.592}{0.024} & \meanstd{0.519}{6.5e-03} & \meanstd{0.618}{0.055} & \meanstd{0.663}{0.017} & \meanstd{0.551}{0.025} & \meanstd{0.626}{0.036} & \meanstd{0.546}{0.053} & \meanstd{0.622}{0.053} & \meanstd{0.621}{0.049} \\
\addlinespace[0pt]
WormsTwoClass & \meanstd{0.787}{0.023} & \meanstd{0.750}{0.0} & \meanstd{0.758}{0.024} & \meanstd{0.792}{0.021} & \meanstd{0.733}{0.033} & \meanstd{0.806}{0.018} & \meanstd{0.780}{0.014} & \meanstd{0.763}{0.033} & \meanstd{0.775}{0.037} \\
\addlinespace[0pt]
Yoga & \meanstd{0.899}{6.5e-03} & \meanstd{0.894}{0.0} & \meanstd{0.893}{3.1e-03} & \meanstd{0.901}{3.3e-03} & \meanstd{0.852}{0.011} & \meanstd{0.892}{3.9e-03} & \meanstd{0.897}{3.1e-03} & \meanstd{0.897}{8.2e-03} & \meanstd{0.894}{8.2e-03} \\
\end{longtable}
}
\endgroup

\clearpage
\begin{table}[t]
\centering
\caption{Geometric mean of inference OPs for TSFresh-40 across 10 random seeds assuming FP16 for floating-point and INT16 for integer arithmetic.}
\label{results-ops-tsfresh40}
\footnotesize{
\begin{tabularx}{\linewidth}{l*{9}{Y}}
\toprule
& KNN & NB & LR & SVM & DT & RF & AB & MLP & DLN \\ 
\midrule
ACSF1 & 1.10M & 2.93M & 1.95M & 11.2M & 1.10K & 149K & 198K & 12.9M & 83.5K \\
\addlinespace[0pt]
ChlorineConcentration & 22.1M & 880K & 585K & 20.8M & 1.79K & 176K & 243K & 15.5M & 81.1K \\
\addlinespace[0pt]
Computers & 2.81M & 587K & 195K & 12.0M & 1.19K & 151K & 180K & 5.22M & 52.1K \\
\addlinespace[0pt]
CricketX & 4.43M & 3.52M & 2.34M & 69.5M & 1.65K & 160K & 255K & 13.7M & 120K \\
\addlinespace[0pt]
CricketY & 4.43M & 3.52M & 2.34M & 70.3M & 1.39K & 179K & 260K & 21.6M & 140K \\
\addlinespace[0pt]
CricketZ & 4.43M & 3.52M & 2.34M & 61.4M & 1.52K & 148K & 259K & 13.1M & 122K \\
\addlinespace[0pt]
Earthquakes & 7.60M & 572K & 193K & 17.8M & 293 & 33.0K & 234K & 6.84M & 25.2K \\
\addlinespace[0pt]
EOGHorizontalSignal & 4.11M & 3.52M & 2.34M & 49.6M & 1.21K & 209K & 202K & 11.5M & 94.6K \\
\addlinespace[0pt]
EOGVerticalSignal & 4.11M & 3.52M & 2.34M & 38.7M & 1.21K & 147K & 251K & 13.8M & 112K \\
\addlinespace[0pt]
EthanolLevel & 10.3M & 1.17M & 780K & 41.0M & 933 & 149K & 189K & 7.96M & 59.4K \\
\addlinespace[0pt]
FaceAll & 6.42M & 4.11M & 2.73M & 67.7M & 1.28K & 210K & 267K & 13.5M & 109K \\
\addlinespace[0pt]
Fish & 2.04M & 2.11M & 1.38M & 15.5M & 1.23K & 176K & 237K & 11.9M & 63.9K \\
\addlinespace[0pt]
FordA & 172M & 587K & 195K & 14.6M & 183 & 18.6K & 136K & 15.3M & 8.36K \\
\addlinespace[0pt]
FordB & 42.6M & 579K & 194K & 128M & 1.15K & 178K & 265K & 10.7M & 69.9K \\
\addlinespace[0pt]
FreezerRegularTrain & 4.36M & 587K & 195K & 4.00M & 238 & 35.0K & 167K & 13.5M & 87.6K \\
\addlinespace[0pt]
GunPointAgeSpan & 1.53M & 594K & 196K & 1.57M & 549 & 69.5K & 169K & 12.5M & 39.8K \\
\addlinespace[0pt]
GunPointMaleVersusFemale & 2.77M & 594K & 196K & 4.08M & 183 & 35.8K & 171K & 16.2M & 19.9K \\
\addlinespace[0pt]
GunPointOldVersusYoung & 2.22M & 602K & 198K & 5.60M & 183 & 24.3K & 136K & 17.0M & 14.5K \\
\addlinespace[0pt]
Ham & 2.18M & 587K & 195K & 3.20M & 769 & 108K & 155K & 8.99M & 38.7K \\
\addlinespace[0pt]
HandOutlines & 11.4M & 579K & 194K & 67.1M & 1.41K & 189K & 253K & 9.59M & 47.8K \\
\addlinespace[0pt]
Haptics & 1.72M & 1.47M & 975K & 12.5M & 1.06K & 116K & 203K & 15.3M & 62.5K \\
\addlinespace[0pt]
InlineSkate & 1.09M & 2.05M & 1.37M & 11.5M & 1.01K & 145K & 204K & 18.3M & 44.3K \\
\addlinespace[0pt]
InsectWingbeatSound & 2.47M & 3.23M & 2.15M & 28.8M & 1.04K & 154K & 242K & 12.0M & 106K \\
\addlinespace[0pt]
LargeKitchenAppliances & 4.26M & 880K & 585K & 27.5M & 1.12K & 174K & 234K & 12.8M & 75.6K \\
\addlinespace[0pt]
MixedShapesRegularTrain & 10.2M & 1.47M & 975K & 30.1M & 1.28K & 178K & 213K & 15.3M & 67.5K \\
\addlinespace[0pt]
MixedShapesSmallTrain & 4.69M & 1.47M & 975K & 5.83M & 659 & 90.0K & 185K & 11.0M & 47.3K \\
\addlinespace[0pt]
NonInvasiveFetalECGThorax1 & 21.2M & 12.3M & 8.19M & 206M & 1.78K & 181K & 245K & 18.3M & 299K \\
\addlinespace[0pt]
NonInvasiveFetalECGThorax2 & 21.2M & 12.3M & 8.19M & 146M & 1.92K & 147K & 272K & 20.4M & 246K \\
\addlinespace[0pt]
OSULeaf & 5.82M & 1.76M & 1.17M & 15.1M & 750 & 141K & 176K & 11.5M & 62.4K \\
\addlinespace[0pt]
Plane & 2.10M & 2.05M & 1.37M & 7.38M & 1.04K & 86.4K & 215K & 12.6M & 43.5K \\
\addlinespace[0pt]
PowerCons & 2.31M & 594K & 196K & 2.21M & 512 & 55.4K & 176K & 8.88M & 36.6K \\
\addlinespace[0pt]
RefrigerationDevices & 4.35M & 891K & 589K & 22.6M & 1.15K & 174K & 239K & 14.0M & 74.6K \\
\addlinespace[0pt]
ScreenType & 4.26M & 880K & 585K & 37.0M & 1.12K & 143K & 209K & 9.44M & 75.4K \\
\addlinespace[0pt]
SemgHandGenderCh2 & 3.39M & 587K & 195K & 23.9M & 1.02K & 151K & 196K & 12.4M & 58.8K \\
\addlinespace[0pt]
SemgHandMovementCh2 & 5.13M & 1.76M & 1.17M & 59.9M & 1.45K & 177K & 224K & 22.7M & 69.3K \\
\addlinespace[0pt]
SemgHandSubjectCh2 & 5.13M & 1.47M & 975K & 45.3M & 1.46K & 201K & 245K & 21.4M & 74.0K \\
\addlinespace[0pt]
ShapesAll & 7.04M & 17.8M & 11.8M & 82.4M & 1.61K & 191K & 243K & 19.5M & 392K \\
\addlinespace[0pt]
SmallKitchenAppliances & 4.26M & 880K & 585K & 22.6M & 787 & 142K & 184K & 6.33M & 99.5K \\
\addlinespace[0pt]
StarLightCurves & 11.6M & 880K & 585K & 33.8M & 439 & 153K & 197K & 10.2M & 38.3K \\
\addlinespace[0pt]
Strawberry & 12.5M & 587K & 195K & 19.0M & 1.23K & 136K & 258K & 13.8M & 54.2K \\
\addlinespace[0pt]
SwedishLeaf & 5.72M & 4.40M & 2.93M & 56.4M & 1.43K & 203K & 259K & 29.2M & 144K \\
\addlinespace[0pt]
Trace & 3.18M & 1.14M & 770K & 5.54M & 439 & 38.9K & 145K & 14.0M & 30.2K \\
\addlinespace[0pt]
TwoPatterns & 11.6M & 1.17M & 780K & 56.3M & 1.39K & 154K & 195K & 11.6M & 51.6K \\
\addlinespace[0pt]
UWaveGestureLibraryAll & 34.5M & 2.35M & 1.56M & 81.2M & 1.37K & 167K & 264K & 17.7M & 74.9K \\
\addlinespace[0pt]
UWaveGestureLibraryX & 10.4M & 2.35M & 1.56M & 88.2M & 1.61K & 156K & 241K & 20.0M & 98.2K \\
\addlinespace[0pt]
UWaveGestureLibraryY & 14.4M & 2.35M & 1.56M & 119M & 1.39K & 142K & 217K & 24.1M & 124K \\
\addlinespace[0pt]
UWaveGestureLibraryZ & 14.4M & 2.35M & 1.56M & 64.5M & 1.59K & 167K & 216K & 22.7M & 106K \\
\addlinespace[0pt]
Wafer & 29.5M & 587K & 195K & 12.4M & 183 & 23.4K & 136K & 15.3M & 6.70K \\
\addlinespace[0pt]
Worms & 1.76M & 1.47M & 975K & 13.6M & 714 & 83.8K & 175K & 9.58M & 31.9K \\
\addlinespace[0pt]
WormsTwoClass & 2.41M & 579K & 194K & 20.7M & 769 & 112K & 172K & 14.1M & 69.4K \\
\addlinespace[0pt]
Yoga & 3.46M & 594K & 196K & 18.5M & 805 & 139K & 214K & 8.99M & 49.9K \\
\bottomrule
\end{tabularx}
}
\end{table}

\end{document}